\documentclass{article}
\usepackage[algo2e,ruled,vlined,linesnumbered]{algorithm2e}

\PassOptionsToPackage{sort,numbers, compress}{natbib}

\usepackage[final,main]{neurips_2025}

\usepackage[utf8]{inputenc} 
\usepackage[T1]{fontenc}    
\usepackage{hyperref}       
\usepackage{url}            
\usepackage{booktabs}       
\usepackage{amsfonts}       
\usepackage{nicefrac}       
\usepackage{microtype}      
\usepackage{xcolor}         
\usepackage{subcaption}
\usepackage{graphicx}

\usepackage{amsmath,nccmath}
\usepackage{amssymb}
\usepackage{mathtools}
\usepackage{amsthm}
\usepackage{wrapfig}
\usepackage{makecell}
\usepackage{commath}

\usepackage{algpseudocode}
\usepackage{float}
\usepackage{xfrac}
\usepackage[capitalize,noabbrev]{cleveref}
\usepackage{colortbl}
\usepackage{multirow}
\usepackage{titletoc}
\usepackage{listings}

\usepackage{xfrac}
\usepackage{tikz}
\def\checkmark{\tikz\fill[scale=0.4](0,.35) -- (.25,0) -- (1,.7) -- (.25,.15) -- cycle;}

\makeatletter
\renewcommand{\SetKwInOut}[2]{%
  \sbox\algocf@inoutbox{\KwSty{#2}\algocf@typo:}%
  \expandafter\ifx\csname InOutSizeDefined\endcsname\relax
    \newcommand\InOutSizeDefined{}\setlength{\inoutsize}{\wd\algocf@inoutbox}%
    \sbox\algocf@inoutbox{\parbox[t]{\inoutsize}{\KwSty{#2}\algocf@typo:\hfill}~}\setlength{\inoutindent}{\wd\algocf@inoutbox}%
  \else
    \ifdim\wd\algocf@inoutbox>\inoutsize%
    \setlength{\inoutsize}{\wd\algocf@inoutbox}%
    \sbox\algocf@inoutbox{\parbox[t]{\inoutsize}{\KwSty{#2}\algocf@typo:\hfill}~}\setlength{\inoutindent}{\wd\algocf@inoutbox}%
    \fi%
  \fi
  \algocf@newcommand{#1}[1]{%
    \ifthenelse{\boolean{algocf@inoutnumbered}}{\relax}{\everypar={\relax}}%
    {\let\\\algocf@newinout\hangindent=\inoutindent\hangafter=1\parbox[t]{\inoutsize}{\KwSty{#2}\algocf@typo:\hfill}~##1\par}%
    \algocf@linesnumbered
  }}%
\makeatother

\SetKwInput{KwInput}{Input}
\SetKwInput{KwOutput}{Output}

\newcommand{\Identity}{{\rm I\kern-.2em l}}
\newcommand{\Expect}{\mathbb{E}}
\newcommand{\Expectbracket}[1]{\mathbb{E}\left[ #1 \right]}
\newcommand{\Expectsubbracket}[2]{\mathbb{E}_{#1}\left[ #2 \right]}
\newcommand{\Expectcond}[2]{\mathbb{E}\left[\left. #1 \right| #2 \right]}

\newcommand{\normsq}[1]{\left\Vert #1 \right\Vert^2}
\newcommand{\innerprod}[1]{\left\langle #1 \right\rangle}

\newcommand{\camera}[1]{{\leavevmode\color{black}#1}} 

\definecolor{darkgreen}{rgb}{0.0, 0.5, 0.0}

\newcommand{\algname}[0]{MESS+\xspace}

\allowdisplaybreaks

\usepackage{thmtools}

\theoremstyle{plain}
\newtheorem{theorem}{Theorem} 
\newtheorem{lemma}{Lemma} 
\newtheorem{corollary}{Corollary} 
\theoremstyle{definition}
\newtheorem{assumption}{Assumption} 
\theoremstyle{remark}

\crefname{theorem}{Theorem}{Theorems}
\crefname{definition}{Definition}{Definitions}
\crefname{assumption}{Assumption}{Assumptions}

\title{MESS+: Dynamically Learned Inference-Time LLM Routing in Model Zoos with Service Level Guarantees}

\author{%
    Herbert Woisetschl\"ager \\
    Technical University of Munich\\
    \texttt{h.woisetschlaeger@tum.de} 
    \And
    Ryan Zhang \\
    Horace Greeley High School\\
    \texttt{ryzhangofficial@gmail.com} 
    \AND
    \hspace{1.1em}Shiqiang Wang \\
    \hspace{1.1em}IBM Research\\
    \hspace{1.1em}\texttt{shiqiang.wang@ieee.org} 
    \And
    \hspace{0.5em}Hans-Arno Jacobsen \\
    \hspace{0.5em}University of Toronto\\
    \hspace{0.5em}\texttt{jacobsen@eecg.toronto.edu} 
}

\begin{document}

\maketitle

\begin{abstract}
    Open-weight large language model (LLM) zoos provide access to numerous high-quality models, but selecting the appropriate model for specific tasks remains challenging and requires technical expertise. Most users simply want factually correct, safe, and satisfying responses without concerning themselves with model technicalities, while inference service providers prioritize minimizing operating costs. These competing interests are typically mediated through service level agreements (SLAs) that guarantee minimum service quality.
    We introduce \algname, a stochastic optimization algorithm for cost-optimal LLM request routing while providing rigorous SLA compliance guarantees. \algname learns request satisfaction probabilities of LLMs in real-time as users interact with the system, based on which model selection decisions are made by solving a per-request optimization problem. Our algorithm includes a novel combination of virtual queues and request satisfaction prediction, along with a theoretical analysis of cost optimality and constraint satisfaction.
    Across a wide range of state-of-the-art LLM benchmarks, \algname achieves an average of $2\times$ cost savings compared to existing LLM routing techniques.

\end{abstract}

\section{Introduction}
\label{sec:intro}
\begin{wrapfigure}[19]{R}{0.4\linewidth}
    \vspace{-1em}
    \centering\includegraphics[width=\linewidth]{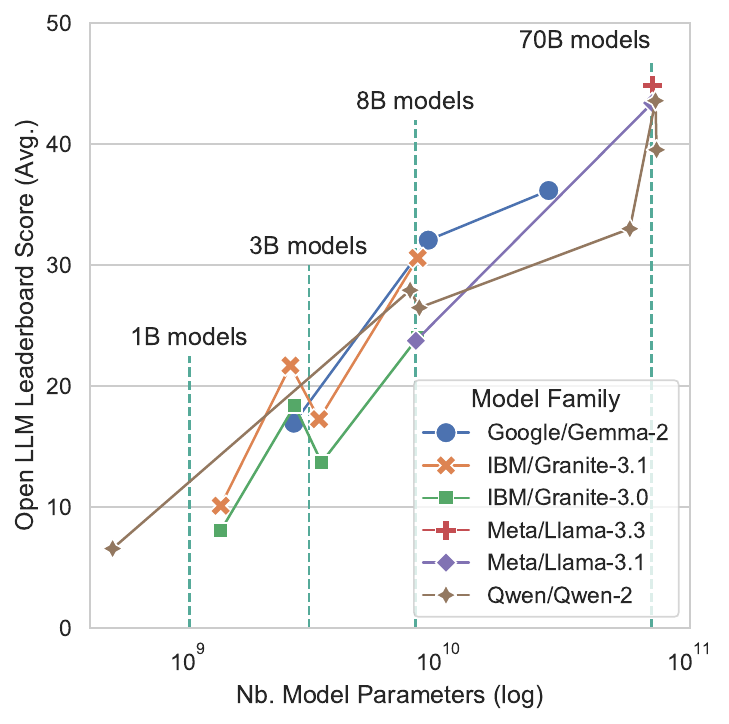}
    \caption{OpenLLM-Leaderboard performance comparison of popular LLM families. Each family typically consists of a minimum of three models with distinct capabilities and cost characteristics.}\label{fig:open_llm_leaderboard_stats}%
\end{wrapfigure}%

As the number of open-weight large language models (LLMs), such as Llama~\citep{llama3_paper}, Granite~\citep{granitegranite} or Qwen~\citep{qwen2.5_paper}, increases rapidly, deep learning infrastructure providers and end users are confronted with an abundance of models (model zoo) for their language processing tasks. 
Typically, each LLM family comes with at least three models, each with different capabilities and resource requirements (\Cref{fig:open_llm_leaderboard_stats}).
Sometimes, there is an update to the model weights that is released as a minor checkpoint (e.g., Llama 3.1 70B and Llama 3.3 70B).
This leaves many users questioning what is the best model to use and whether common benchmark results apply to their specific needs~\citep{Humza2023}.
Currently, the best way to approach model selection is educated guessing, using LLM benchmarks as a proxy to estimate model performance, or spending significant efforts to curate human-preference datasets for request routing ~\citep{ong2024routellm}. 
Since working with LLMs can be expensive~\citep{Samsi2023}, minimizing costs is an equally high priority for end users and inference endpoint operators. This leaves us with the following tri-fold problem.

\textbf{End-users primarily care about a factually correct and safe model output.}
When inquiring about text information, e.g., by asking questions or requesting language translation, end users are mostly interested in obtaining factually correct and sufficiently clear language output~\citep{pmlr-v235-wettig24a}.
Additionally, many users are unfamiliar with the technical details of LLMs, making it challenging for them to select the right model for the job, i.e., their primary references are domain-specific benchmark rankings~\citep{open-llm-leaderboard-v2, eval-harness}. 
However, there is no intuitive method to compare the complexity of individual requests with benchmark tasks. 
Thus, we require a method that learns over time whether a model can satisfy an incoming request as users interact with LLMs.

\textbf{Inference endpoint providers prioritize low operating costs, and emerging AI legislation requires sustainable computing best practices.}
Operating infrastructure that can run state-of-the-art LLMs can be costly. 
Microsoft has announced it will acquire a stake in the Three Mile Island nuclear power plant in the United States to satisfy the energy demand of its planned data center in Pennsylvania, which has two reasons: consistent energy delivery and low energy cost~\citep{reuters2024}.
At the same time, globally emerging AI regulation (e.g., EU AI Act Article 95)~\citep{eu_ai_act} emphasizes energy monitoring and compliance with sustainable computing best practices.
Taken together, tighter regulation and ever increasing cost underpin the necessity of energy-optimal service operations. 
The current operating model of many inference service providers where only a single model is served per dedicated endpoint does not offer the necessary flexibility to keep cost in check.
When aiming to provide cost optimal inference endpoints, we need a method that can choose the most efficient model that is likely to satisfy an incoming request.

\textbf{Enterprise use-cases require a consistently high-quality model inference output while keeping costs in check.}
This unites the requirement for high-quality model outputs and price sensitivity.
Thus, commercial players typically rely on service-level agreements (SLAs) when sourcing services for their own products. An example of an SLA is one that specifies a percentage of requests to be automated by LLM-powered AI agents, while the remaining requests need to be handled by human experts. 
Typically, different SLA levels offer different quality standards at different prices. 
However, to date, it is usually up to end users to decide which models to use for their requests without any lower bounds on request satisfaction guarantees.
As such, this is neither cost effective for the inference endpoint provider nor for the end user as both tend to overpay on operating costs and inference response costs, respectively. 
Thus, we require a method that minimizes operating costs while guaranteeing SLA compliance, i.e., a minimum request satisfaction rate over time.

In summary, we ask the fundamental question: 
\begin{quote}
    \textit{%
        How can we design an algorithm for selecting LLMs from a model zoo that minimizes the operating cost while providing rigorous SLA compliance guarantees?
        }
\end{quote}
This problem is challenging in several aspects. 
First, we need to find a way to learn whether an LLM can satisfy an incoming request as users interact with an inference endpoint over time. 
Second, we require a method to guarantee SLA compliance over time, while allowing the inference service provider to minimize their operating costs. 
Taken together, we have to guarantee that our per-request routing decisions rigorously satisfy an SLA requirement while ensuring cost optimality.

\textbf{Contributions.} 
Our paper introduces \textbf{\algname} (\textbf{M}odel S\textbf{E}lection with Cost-optimal \textbf{S}ervice-level Guarantee\textbf{S}), a new stochastic optimization framework for minimizing operating costs and rigorously guaranteeing SLA compliance. 
Compared to existing LLM request routing techniques, we offer a solution to \emph{cost optimal request routing} and provide a \emph{lower bound request satisfaction guarantee}.
Our approach dynamically learns request satisfaction probabilities in an online fashion as users interact with the system and makes routing decisions based on operating cost and request satisfaction over time. 
We provide \emph{theoretical guarantees for cost-optimal SLA compliance with model zoos}.

\begin{table}[t]
    \centering
    \caption{Related work overview. We are the first to introduce a cost-optimal stochastic optimization framework for LLM request routing with SLA guarantees. 
    }
    \label{tab:related_work}

    \resizebox{\linewidth}{!}{
        \begin{tabular}{@{}llrccl@{}}
        \toprule
        \thead{Approach}   & \thead{Technique}       & \thead{\# of LLMs \\ in Zoo}     & \thead{Cost-\\aware}    & \thead{Service-Level \\ Guarantee} & \thead{Source}     \\ \midrule
        LLM-Blender        & LLM Ensemble            &  $>2$                        & --                      & --                                  & \citet{llm_blender} \\
        AutoMix            & Self Verification       &  $2$                           & --                      & --                                  & \citet{aggarwal2024automix} \\
        Hybrid-LLM         & Preference Data         &  $2$                           & --                      & --                                  & \citet{ding2024hybrid} \\
        Zooter             & Reward-Model Labels     &  $>2$                        & --                      & --                                  & \citet{zooter} \\
        RouterDC           & Contrastive Learning    &  $>2$                        & --                      & --                                  & \citet{chen2024routerdc} \\
        TensorOpera Router & BertSim Scores               &  $>2$                        & \checkmark              & --                                  & \citet{stripelis-etal-2024-tensoropera} \\
        RouteLLM           & Preference Data         &  $2$                           & \checkmark              & --                                  & \citet{ong2024routellm} \\ 
        \midrule
        \textbf{MESS+ (ours)}       & \camera{Satisfaction Scores \& Online Optimization} &  $>2$                    & \checkmark              & \checkmark                          & This paper \\ \bottomrule
        \end{tabular}
    }
\end{table}

\textbf{Related Work.}
While prior works in LLM request routing have made significant contributions in various directions (\Cref{tab:all_benchmark_evals}), \algname distinguishes itself through its formal optimization approach to cost efficiency with rigorous SLA guarantees. 
Specifically, RouteLLM~\cite{ong2024routellm}, Zooter~\cite{zooter}, and RouterDC~\cite{chen2024routerdc} focus on optimizing routing decisions based on model capabilities and query characteristics without formal SLA guarantees. 
LLM-Blender~\cite{llm_blender} and AutoMix~\cite{aggarwal2024automix} emphasize quality improvement through ensembling and self-verification approaches, respectively, without providing guarantees for a lower bound request satisfaction rate over time. 
Similarly, Hybrid-LLM~\cite{ding2024hybrid} introduces quality-aware routing between models of different sizes, and TensorOpera Router~\cite{stripelis-etal-2024-tensoropera} balances performance metrics empirically. 
In contrast, \algname uniquely formulates request routing as a stochastic optimization problem that minimizes operating costs while providing a minimum request satisfaction guarantee over time, adapting dynamically through online learning as users interact with the system. 
This optimization-driven approach with provable guarantees positions \algname as the first framework to deliver cost-optimal SLA compliance in LLM zoos.

\section{MESS+: Model Selection with Cost-Optimal Service Level Guarantees}
\label{sec:methodology}

The overall goal of \algname is to find the most suitable LLM for each inference request $t \in \{1,2,\ldots,T\}$ to minimize the operating cost $E_{m,t}$, while conforming to model performance constraints defined by an SLA over time to ensure contractual compliance.

\subsection{Problem Formulation}
Consider a language model zoo with $M$ different LLMs. 
For every request $t$, each model is associated with a user satisfaction $s_{m,t} \in \{0,1\}$ indicating whether model $m$ can satisfy the $t$-th request, where $m\in\{1,2,\ldots,M\}$. The value of $s_{m,t}$ is \textit{unknown} before request $t$ arrives.

\textbf{Inference Cost} (Objective). 
Each model in a zoo incurs a certain amount of cost $E_{m,t}$ (e.g., cost for an API call, energy consumption of an inference request), which can vary greatly based on the model and inference request size.
For instance, a zoo can include models with 1B, 8B, and 70B parameters.  
These differences in size make costs volatile. 
In a scenario where users can choose the model and are unsure which model fits their request, they are likely to always choose the largest model, making model serving expensive, rendering costs even more volatile and overall hard to predict.

\textbf{Service-Level Agreement} (Constraint).
Typically, contracts related to a service contain a list of requirements, including an SLA defining a minimum service quality. 
The SLA functions as a measurable control system with defined input variables, output metrics, and acceptable tolerance ranges.
We define $\alpha$ to be the target request satisfaction rate over time\footnote{In practice, $\alpha$ should be chosen with a certain safety margin from the SLA requirement such that we do not violate the SLA even if the average request satisfaction is slightly below $\alpha$.}, i.e., the relative share of requests that need to be served with a satisfactory LLM response over time. 

\textbf{Control Problem.}
Taking the objective and constraint together, we can formalize the following problem of minimizing the average operating cost per request under a minimum performance requirement defined via an SLA over time: 
\begin{subequations}
\label{eq:original_problem}
\begin{align}
    \textstyle
    \min_{\{y_{m,t}:\forall t, m\}} \quad & \textstyle \frac{1}{T} \sum_{t=1}^T \sum_{m=1}^M \Expectbracket{y_{m,t} E_{m,t}}, \label{eq:energy_minimization} \\
    \text{s.t.} \quad & \textstyle \frac{1}{T} \sum_{t=1}^T \sum_{m=1}^M \Expectbracket{y_{m,t} s_{m,t}} \geq \alpha, \label{eq:accuracy_constraint} \\
    \textstyle
     & \textstyle \sum_{m=1}^M y_{m,t} = 1,\quad \forall t \in \{1, \dots, T\}, \label{eq:one_model_constraint} \\
     & \textstyle y_{m,t} \in \{0, 1\},\quad \forall t  \in \{1, \dots, T\}, m  \in \{1, \dots, M\}, \label{eq:binary_constraint}
\end{align}
\end{subequations}
where $y_{m,t}=1$ if model $m$ is chosen and $y_{m,t}=0$ otherwise.

\textbf{Challenges.} 
Our optimization problem involves an inherent trade-off between request satisfaction and operating cost, since larger LLMs, and thus more capable ones, typically yield higher satisfaction rates while consuming more resources at the same time.
As such, we see a \textit{correlation over time between the objective and constraints}.    
Further, optimizing the operating cost involves a time average that is hard to predict a priori as the properties of future requests are generally \textit{unknown} and \textit{heterogeneous}.

\begin{wrapfigure}[21]{R}{0.5\textwidth}
    \vspace{-3.1em}
    \begin{algorithm2e}[H]
        \caption{\textbf{M}odel S\textbf{E}lection with Cost-optimal \textbf{S}ervice-level Guarantee\textbf{S} (\algname)}
        \label{algo:messplus}
        \scriptsize
        \KwInput{
            $T$, 
            $V$,
            $\alpha$,
            $c$,
            learning rate $\eta_t>0, \forall t$
        }
        
        \KwOutput{$\{y_{m,t}\!:\!\forall m,\! t\}$, outputs of chosen models for all~$t$}
        
        Initialize $Q_1 \leftarrow 0$, random vector $\mathbf{z}_1$; 
        
        \For{$t \leftarrow 1$ \KwTo $T$}{
            Compute $p_t \leftarrow \min\left(1, \sfrac{c}{\sqrt[4]{t}}\right)$\; 
            
            Sample $X_t \sim \text{Bernoulli}(p_t)$\;
            
            \eIf{$X_t = 1$ or $t=1$}{
                \tcp{Explore model zoo}
                $y_{m,t}\leftarrow 1, \forall m$; \tcp{all models queried}\label{alg-line:select_all_models}
                \ForEach{$m \in \{1, 2, ..., M\}$}{
                    Obtain true request satisfaction $s_m(t)$\;
                }

                $\mathbf{z}_{t+1} \leftarrow \mathbf{z}_{t} - \eta_t\nabla F(\mathbf{z}_{t}, \mathbf{a}_{t})$; \ \tcp{SGD using request $t$'s content $\mathbf{a}_t$}\label{alg-line:sgd}
                $m^{*}\leftarrow\arg\max_m s_{m,t} $\;                
            }{
                \tcp{Online decision making with user satisfaction predictor}
                Predict request satisfaction probability $\hat{s}_{m\!,t}\!, \forall m$\;
                $m^* \!\leftarrow\!\arg\min_m V  E_{m,t} + Q_t (\alpha - \hat{s}_{m,t})$;                 \tcp{Solve Problem \eqref{eq:per-request-optimization}}\!
                $y_{m^*,t}\leftarrow 1$, $y_{m',t}\leftarrow 0, \forall m'\neq m^*$\;

                $\mathbf{z}_{t+1} \leftarrow \mathbf{z}_{t}$\;
            }

            Get output from model $m^{*}$ and its accuracy $s_{m^{\!*}, t}$\;
            
            $Q_{t+1}\leftarrow \max\{0, Q_t + \alpha - s_{m^{\!*}, t}\}$; 
            \tcp{Virtual queue update} 
        }
    \end{algorithm2e}
\end{wrapfigure}

\subsection{Online Decision Making}

We introduce an online decision making process that addresses the aforementioned challenges. 
The quantities $E_{m,t}$ and $s_{m,t}$ are captured for every request without knowledge of future statistics. The full procedure of \algname is described in Algorithm~\ref{algo:messplus}.

\textbf{Methodology.} Our approach is inspired by the Lyapunov drift-plus-penalty framework \cite{neely2022stochastic}, with \textit{novel extensions} to support a request satisfaction predictor that is learned in an online manner and used in per-request model selection decisions.

\textbf{Request Satisfaction Prediction.} 
Since $s_{m,t}$ is only known after invoking model $m$, we need a mechanism that predicts whether a model $m$ will meet the request satisfaction requirements, so that we can select an appropriate LLM for every incoming request \textit{before} the request is sent to any LLM.
We learn a predictor $\hat{s}_{m,t}\in[0,1]$ online that predicts the \textit{probability} that model $m$ can satisfy an incoming request $t$ for all request-model combinations. The predictor takes the request as input and extracts useful information of the request using a lightweight model to make the prediction. For different requests, the prediction $\hat{s}_{m,t}$ is usually different.
The detailed procedure of this prediction is described in \Cref{sec:user-preference-estimation}. 

\textbf{Virtual Queues.}
We capture SLA violations, i.e., accumulated undershooting of $\alpha$, in a \textit{virtual queue}~$Q$ with the following update procedure after receiving and processing request $t$:
\begin{align}
    \textstyle
    Q_{t+1} = \max \left\{0, Q_t + \alpha - \sum_{m=1}^M y_{m,t}s_{m,t}\right\},
    \label{eq:queue-update}
\end{align}
where for $t=1$, we initialize $Q_1=0$.
Intuitively, this captures the cumulative constraint violations, which can be seen by comparing \eqref{eq:accuracy_constraint} and \eqref{eq:queue-update}. 
Hence, we aim to collectively minimize our objective \eqref{eq:energy_minimization} and the queue length\footnote{We note that the average satisfaction rate over requests needs to be greater than or equal to $\alpha$ in the constraint, so the direction of inequalities in the constraint is opposite to \cite{neely2022stochastic}, thus our queue update equation in \eqref{eq:queue-update} is slightly different from that in \cite{neely2022stochastic}.}.

\textbf{Decision Problem for Each Request.} 
We aim to minimize the operating cost of every inference request $t$ while complying with our SLA requirement $\alpha$. This trade-off is formulated as follows: 
\begin{subequations}
\begin{align}
    \textstyle
    \min_{\{y_{m,t}:\forall m\}} \quad V\cdot \sum_{m=1}^M & \textstyle y_{m,t} E_{m,t} + Q_t \left(\alpha - \sum_{m=1}^M y_{m,t}\hat{s}_{m,t}\right)\mathrm{,} \label{eq:per-request_obj} \\
    \textrm{s.t.}\quad  & \textrm{Constraints \eqref{eq:one_model_constraint}, \eqref{eq:binary_constraint}}\,.
\end{align}%
\label{eq:per-request-optimization}%
\end{subequations}
As SLAs come in various configurations, we introduce the parameter $V > 0$ in \eqref{eq:per-request-optimization} that controls the speed at which we reach $\alpha$ and the trade-off between operating cost and time at which we can guarantee constraint (SLA) satisfaction.
The effect of $V$ will be further discussed in both theory and experiments in later sections. In particular, constraint satisfaction is theoretically guaranteed for large enough $T$ (see \Cref{sec:performance_analysis}). 
\camera{When $V$ is small, the constraint violation decreases faster in $T$, i.e., we achieve the SLA requirement $\alpha$ more quickly, but the cost is higher, and vice versa. In practice, we see that a fixed $V$ obtained from coarse tuning works reasonably well across a wide range of benchmarks (see Section~\ref{sec:experiments}).}

Note that we assume that the user satisfaction is captured immediately after the LLM response is generated. In practice, users would be shown a feedback prompt to assess the response quality or request human assistance if the response is unsatisfactory. Therefore, in \eqref{eq:queue-update}, we use $s_{m,t}$ to update the virtual queue length to capture the actual constraint satisfaction, but we use $\hat{s}_{m,t}, \forall m$, to solve the per-request optimization problem \eqref{eq:per-request-optimization}.
We consider the operating cost per request $E_{m,t}$, $\forall m,t$, to be known\footnote{Practically, this can be done by profiling the cost of inference calls before adding a model to the zoo.} once we receive the request and obtain some of its basic information, e.g., number of tokens in the prompt.

\subsection{Request Satisfaction Prediction}
\label{sec:user-preference-estimation}

We now describe how to obtain the predicted request satisfaction probability $\hat{s}_{m,t}$, $\forall m, t$, which is required to solve \eqref{eq:per-request-optimization}. To facilitate the description, let $\mathbf{z}_{t}$ denote the parameter vector of the \textit{lightweight} request satisfaction predictor used for request $t$, and $\mathbf{a}_t \sim \mathcal{A}$ denote the $t$-th input request sampled from some distribution $\mathcal{A}$. The predictor provides an $M$-dimensional output $\hat{s}(\mathbf{z}_{t}, \mathbf{a}_{t})$, which includes $\{\hat{s}_{m,t}, \forall m\}$.  We write the $m$-th component of the output vector as $\hat{s}(\mathbf{z}_{t}, \mathbf{a}_{t})_m = \hat{s}_{m,t}$.
We omit the subscript $t$ in $\mathbf{z}_{t}$ in the following when it is unnecessary to specify the request index $t$.

We define the regularized cross entropy objective of request satisfaction predictions for over all possible incoming requests and for all models: 
\begin{align}
\label{eq:cross-entropy}
\textstyle
        F(\mathbf{z}) := -\Expectsubbracket{\mathbf{a}_{t}\sim\mathcal{A}}{\frac{1}{M}\sum_{m=1}^M \left( s_{m,t}\log \hat{s}(\mathbf{z}, \mathbf{a}_{t})_m  + (1\!-\! s_{m,t})  \log (1\!-\!\hat{s}(\mathbf{z}, \mathbf{a}_{t}))_m \right) }\! +\! \frac{\mu}{2}\normsq{\mathbf{z}},
\end{align}
where the last term is a regularization term when $\mu>0$. 

We learn $\mathbf{z}$ through a probabilistic exploration and update procedure. 
To \textit{explore} the model zoo, we query all models\footnote{In Line~\ref{alg-line:select_all_models} of \Cref{algo:messplus}, we set $y_{m,t}=1$ for all $m$ to reflect that all the models are queried during exploration. This is with a slight abuse of notation, since this choice of $\{y_{m,t}\}$ does not satisfy \eqref{eq:one_model_constraint}. However, we use this configuration to indicate that cost has been incurred for querying all the LLMs.} with the same input request to obtain their actual request satisfaction $s_{m,t}$, allowing us to learn $\mathbf{z}$.
More specifically, as shown in \Cref{algo:messplus}, we sample from a distribution $X_t \sim \mathrm{Bernoulli}(p_t)$, where the exploration probability $p_t = \min(1, \frac{c}{\sqrt[4]{t}})$ and $c > 0$ is a parameter that adjusts this probability. 
The probability $p_t$ decays over time as the estimation $\hat{s}_{m,t}$ improves with each exploration and update step.
The larger $c$, the more likely it is to perform an exploration step over time.
When exploring, we always use the output from the largest model as the final model output as we have already incurred operating cost to query the largest model, i.e., we \textit{do not} use the solution from \eqref{eq:per-request-optimization} in this case and return the output of the largest model.
Especially for the first few arriving requests, when we do not know how to choose the optimal LLM for request $t$, we must explore each $m$ in the model zoo for the best-performing model for each request. 
In this way, we capture the actual user preference $s_{m,t}$ of each $m$, which we use to learn~$\mathbf{z}$. 

We note that the predictor we train here only predicts whether each model $m$ can produce a satisfactory response to an incoming request. Different from related works such as~\cite{ong2024routellm,chen2024routerdc}, we do not use this trained predictor as a router directly. Instead, the output of the predictor is fed into the per-request optimization problem \eqref{eq:per-request-optimization}, and the model selection decision is obtained by solving \eqref{eq:per-request-optimization}.

\textbf{Key Insight.}
The key idea behind exploration and predictor update with a probability $p_t$ that decreases in $t$ is to strike a balance between obtaining an accurate predictor and limiting the additional cost incurred due to exploration. If $p_t$ decreases too fast in $t$, it will take a long time to obtain an accurate predictor although the cost overhead due to exploration is low. In contrast, if $p_t$ decreases too slowly in $t$, we obtain an accurate predictor quickly, but the cost overhead is also high since we do many more exploration steps than it is actually needed. By choosing $p_t = \min(1, \frac{c}{\sqrt[4]{t}})$, we have a good balance between the predictor accuracy and exploration cost overhead, as also confirmed by our theoretical analysis in the next section.

\section{Performance Analysis}
\label{sec:performance_analysis}

In this section, we provide a theoretical performance analysis of our proposed \algname algorithm.
We focus on showing constraint satisfaction and optimality of our solution.
Since SLAs are often volume-based, service quality guarantees are typically given for a specific number of requests or transactions. 
Thus, we focus on obtaining results that hold for any $T\geq 1$ in our analysis, instead of only considering $T\rightarrow\infty$ as in~\cite{neely2022stochastic}. As commonly done in the literature, our analysis relies on some reasonable assumptions to make the problem mathematically tractable.
The full proofs for all the following theorems are provided in the appendix.

\subsection{Constraint Satisfaction}

\begin{assumption}
\label{assumption:queue_length}
Let $\tilde{m} := \arg\max_m \hat{s}_{m,t}$ and $\psi \geq 0$ be some constant independent of the total number of requests $T$ and $\psi \ll T$.
We assume that, when $t \geq \psi$, there exist $\beta>0$ and $q \in (0, 1]$, such that for any $m$ (inclusive of $\tilde{m}$) with $\hat{s}_{\tilde{m},t} - \hat{s}_{m,t} \leq \beta$, we have $\Pr\{s_{m,t} = 1\}\geq q > \alpha$.
\end{assumption}
This assumption states that, after a finite number of requests $\psi$, our request satisfaction predictor becomes accurate enough, so that models within a certain range $\beta$ of the maximum predicted satisfaction guarantees that the minimum probability of actual satisfaction is at least $q>\alpha$.

\begin{theorem}
\label{theorem:max_queue_length}
    For any $t\geq 1$, we have the following upper bounds on the virtual queue length: 
    \begin{align}
    \textstyle
    \Expectbracket{Q_t} \leq \max\left\{\alpha \psi; \frac{V\Delta_E}{\beta}\right\} + \sqrt{t} \,\,\textrm{ and }\,\, \frac{1}{t}\sum_{\tau=1}^t\Expectbracket{Q_t} \leq \max\left\{\alpha \psi; \frac{V\Delta_E}{\beta}\right\} + \frac{1}{2(q-\alpha)},
    \end{align}
    where $\Delta_E := E_{\max} - E_{\min}$, in which $E_{\mathrm{max}}$ and $E_{\mathrm{min}}$ are the maximum and minimum operating costs of any model, respectively.
\end{theorem}

The proof for \Cref{theorem:max_queue_length} is based on a unique observation that, when $Q_t > \max\left\{\alpha \psi; \sfrac{V\Delta_E}{\beta}\right\}$, the solution to \eqref{eq:per-request-optimization} is guaranteed to satisfy $\hat{s}_{\tilde{m},t} - \hat{s}_{m^*,t} \leq \beta$, where $m^*$ denotes the solution to \eqref{eq:per-request-optimization} such that $y_{m^*,t}=1$. The final result is then obtained by bounding the Lyapunov drift of an auxiliary queue capturing how much $Q_t$ exceeds $\max\left\{\alpha \psi; \sfrac{V\Delta_E}{\beta}\right\}$.
Based on the queue update in~\eqref{eq:queue-update}, it is easy to obtain the following corollary.

\begin{corollary}\label{corollary:constraint_satisfaction_at_t}
    We have the following upper bound of constraint violation (averaged over time):
    \begin{align}
        \textstyle
        \alpha\! -\! \frac{1}{T}\sum_{t=1}^T\sum_{m=1}^M \!\Expectbracket{y_{m,t}s_{m,t}} = \frac{\Expectbracket{Q_T}}{T} \leq \max\left\{\frac{\alpha \psi}{T}; \frac{V\Delta_E}{\beta T}\right\} \!+\! \frac{1}{\sqrt{T}} = \mathcal{O}\left(\frac{V}{T}\! + \!\frac{1}{\sqrt{T}}\right) .  
    \end{align}
\end{corollary}
We observe that the constraint violation is guaranteed to be arbitrarily small when $T$ is sufficiently large. If full SLA compliance (denoted by $\alpha'$) needs to be guaranteed after a finite number $T_0$ of requests, we can choose $\alpha$ to be slightly larger than $\alpha'$, so that $\max\left\{\frac{\alpha \psi}{T_0}; \frac{V\Delta_E}{\beta T_0}\right\} + \frac{1}{\sqrt{T_0}} \leq \alpha-\alpha'$, which holds for $\alpha=\alpha'  +\frac{V\Delta_E}{\beta T_0}+\frac{1}{\sqrt{T_0}}$ when $\frac{\alpha \psi}{T_0} \leq \frac{V\Delta_E}{\beta T_0}$. In practice, as we see in our experiments in \Cref{sec:experiments}, our \algname algorithm satisfies the constraint empirically after a relatively small number (e.g., slightly more than a thousand) of requests even if we use $\alpha$ as the SLA requirement directly.

\subsection{Cost Optimality}

\begin{assumption}
    \label{assumption:iid_arrival}
    We assume that both the input request content $\mathbf{a}_t$ and cost $E_{m,t}, \forall m$ are  i.i.d. across $t$, while for the same $t$, $E_{m,t}$ may be dependent on $\mathbf{a}_t$.
\end{assumption}

\begin{assumption}
\label{assumption:predictor_loss}
The predictor training loss $F(\mathbf{z})$ is $L$-smooth. It also satisfies the Polyak–Łojasiewicz (PL) condition with parameter $\mu>0$ (and $\mu\leq L$), i.e., $\tfrac12\|\nabla F(\mathbf{z})\|^2 \;\ge\; \mu\bigl(F(\mathbf{z})-F_\mathrm{min}\bigr), \forall \mathbf{z}$, where $F_\mathrm{min}:=\min_\mathbf{z}F(\mathbf{z})$. 
Its stochastic gradient is unbiased and has a bounded variance of $\sigma^2$, i.e., $\Expect\big[\nabla F(\mathbf{z},\mathbf{a})\big| \mathbf{z}\big]=\nabla F(\mathbf{z})$ and $\Expect\big[\normsq{\nabla F(\mathbf{z},\mathbf{a}) - \nabla F(\mathbf{z})}\big|\mathbf{z}\big]\leq\sigma^2$, $\forall \mathbf{z}$, where $\nabla F(\mathbf{z},\mathbf{a})$ denotes the stochastic gradient of $F(\mathbf{z})$ on sample $\mathbf{a}\sim\mathcal{A}$.
\end{assumption}
\Cref{assumption:iid_arrival} is commonly used in the Lyapunov drift-plus-penalty framework~\cite{neely2022stochastic} and \Cref{assumption:predictor_loss} is common in stochastic gradient descent (SGD) convergence analysis. In our experiments, we empirically show that \algname also works well in non-i.i.d. settings (\Cref{app:non_stationary_benchmark}).

\begin{theorem}
    \label{theorem:optimality_bound}
    For $\{y_{m,t}:\forall m, t\}$ obtained from the MESS+ algorithm (\Cref{algo:messplus}), there exists a learning rate schedule $\{\eta_t:\forall t\}$, such that we have
    \begin{align}
        \textstyle
        \Expectbracket{\frac{1}{T}\sum_{t=1}^T \sum_{m=1}^M y_{m,t} E_{m,t} }   
        \leq E^\textnormal{OPT} +\mathcal{O}\left(\frac{M}{\sqrt[4]{T}} + \frac{1}{V} + M F_\mathrm{min}  \right) , 
    \end{align}    
    where $E^\textnormal{OPT}$ is the optimal solution to \eqref{eq:original_problem} that is obtained from an idealized stationary policy which assumes full statistical knowledge of requests in $t\in\{1,\ldots,T\}$.
\end{theorem}

The proof of \Cref{theorem:optimality_bound} includes \textit{key novel steps} to capture the joint effect of per-request optimization in \eqref{eq:per-request-optimization}, the error of the request satisfaction predictor, and the additional cost incurred due to exploration for training the predictor. The latter two aspects do not exist in the framework in~\cite{neely2022stochastic}. 
In particular, we first bound the drift-plus-penalty expression and obtain a term $\Expectbracket{Q_t}\cdot \Expect\big[ \max_m\{\,\abs{\hat{s}_{m,t} - s_{m,t}}\}\big]$ in the bound. Then, we further bound the prediction error $\Expect\big[ \max_m\{\,\abs{\hat{s}_{m,t} - s_{m,t}}\}\big]$ using SGD convergence analysis while considering properties of the cross-entropy loss \eqref{eq:cross-entropy}, where the number of SGD steps for predictor training (Line~\ref{alg-line:sgd} of \Cref{algo:messplus}) is related to the exploration probability $p_t=\mathcal{O}(\sfrac{1}{\sqrt[4]{t}})$. We also incorporate the cost overhead for exploration that is $\mathcal{O}(\sfrac{M}{\sqrt[4]{T}})$. Combining these and using the average queue length bound in \Cref{theorem:max_queue_length}, we obtain the result.

We have several important observations from \Cref{theorem:optimality_bound}. First, combining with \Cref{theorem:max_queue_length}, we can confirm that $V$ controls the trade-off between cost optimality and constraint satisfaction, where a larger $V$ boosts cost optimality but slows down constraint satisfaction, and vice versa. Second, the cost optimality gap depends on $M F_\mathrm{min}$, where we recall that $M$ is the number of LLMs in the zoo and $F_\mathrm{min}$ is the minimum loss that can be obtained for training the request satisfaction predictor. When the predictor is capable, $F_\mathrm{min}$ is small, which is what we also observe in the experiments in \Cref{sec:experiments}. Finally, when we choose $V=\sqrt{T}$, we can observe from Theorems~\ref{theorem:max_queue_length} and \ref{theorem:optimality_bound} that, as $T\rightarrow\infty$, MESS+ guarantees full SLA satisfaction and approximate cost optimality up to $M F_\mathrm{min}$, i.e., the minimum loss of the predictor times the number of LLMs in the zoo.
\vspace{-0.4em}


\section{Experiments}
\label{sec:experiments}
We demonstrate the effectiveness of \algname across a set of experiments with state-of-the-art LLM benchmarks.
Our code is publicly available\footnote{\algname code repository: https://github.com/laminair/mess-plus} and we provide full experimental details in the appendix. 
\vspace{-0.4em}

\subsection{Setup}

\textbf{Language Model Zoo \& Benchmarks.}
\camera{Our model zoo is comprised of three models: Llama 3.2 1B (L1B), Llama 3.1 8B (L8B), and Llama 3.3 70B model (L70B). 
} 
We use the LM-Eval Harness~\cite{eval-harness} and deploy three reasoning benchmarks (ARC Easy, ARC Challenge, and Winogrande)~\cite{allenai:arc,Sakaguchi2020} as well as five Q\&A benchmarks (BoolQ, LogiQA, PiQA, SciQ, and SocialIQA)~\cite{clark2019boolq,Liu2020,Bisk2020,SciQ,sap2019}.
All benchmarks are evaluated zero-shot.  
On a per-sample basis, the benchmarks generate binary feedback signals that indicate whether a request has been satisfied. 
These binary signals are used as labels for training our request satisfaction predictor.

\textbf{Request Satisfaction Predictor.}
In line with related work~\cite{ong2024routellm}, we choose ModernBERT~\cite{modernbert} as a transformer backbone and implement a multi-label classifier on top of it, where each label corresponds to whether a model can satisfy a user request.
We freeze the transformer parameters and only train the classifier with SGD.
The classifier is trained online while running each benchmark as described in our algorithm. 
We set $c = 0.1$, if not specified otherwise.

\textbf{User \& Service Provider Requirements.}
For each benchmark, we consider a pre-defined $\alpha$ stating the minimum request satisfaction rate over time. In practice, this is specified in the SLA between the user and service provider. 
The value of $\alpha$ in our experiments is set individually for every benchmark, based on the capabilities of the models in our zoo. 
To complete the SLA, a user and service provider need to agree on the cost of an inference service, i.e., they need to negotiate how many SLA violations are acceptable in the beginning. 
The service provider sets $V$ accordingly, to minimize operating costs over time. 
Regardless of $V$, \algname will eventually converge towards $\alpha$, as our theory has shown; $V$ only defines how long it may take.
We set $V = 0.0001$ by default.
\textit{Since SLA metrics are usually measured over a period of time instead of instantaneously, it is sufficient to satisfy SLA requirements after a pre-defined number of requests.}

\textbf{Objective.}
We  measure the effectiveness of \algname by its ability to meet $\alpha$ at minimal operating cost. 
In our experiments, we use the per-request energy consumption when querying an LLM (measured in megajoule, MJ) as the cost metric. 
\camera{We also present the model call ratio, i.e., the share of benchmark requests routed to each model.}

\textbf{Baselines.}
We first look at each individual LLM with regard to their operating cost and request satisfaction capabilities. 
Then, we compare \algname with three adaptive routing baselines, namely \textit{RouteLLM}~\cite{ong2024routellm}, \textit{RouterDC}~\cite{chen2024routerdc}, and \textit{``educated guessing''}. 
We configure RouteLLM with their BERT-based router model configuration. 
Note that RouteLLM only supports routing between two models, so
we set the small and large model to L1B and L70B, respectively.
RouterDC supports routing between multiple LLMs, i.e., can route our entire model zoo.
We adopt the configuration from the RouterDC paper~\cite{chen2024routerdc}.
Additionally, we employ an ``educated guessing'' baseline that randomly chooses an LLM by assuming the availability of prior knowledge on the probability that each LLM satisfies the request, while conforming to our SLA requirements over time.
\camera{For all the baselines, we tune available hyperparameters so that the baselines satisfy the SLA requirement while being the most cost efficient.}
We provide further details on all the baselines in the appendix.
\vspace{-0.4em}

\begin{table}
\setlength{\tabcolsep}{1.5pt}
\centering
\caption{Main results. Performance across three reasoning and five Q\&A benchmarks. \textcolor{darkgreen}{Green} highlights all methods that satisfy the service level requirement $\alpha$ and \textcolor{red}{red} all violations.
The most cost efficient single model satisfying $\alpha$ is \underline{underlined} and the most efficient adaptive method is highlighted in \textbf{bold}.
We report operating costs in megajoule (MJ) energy consumption.
For a full overview with more $\alpha$ variations please see the appendix.}
\vspace{8pt}
\label{tab:all_benchmark_evals}
\resizebox{1.0\textwidth}{!}{
    \begin{tabular}{lccrccrccr}
\cmidrule(lr){1-1}\cmidrule(lr){2-4}\cmidrule(lr){5-7}\cmidrule(lr){8-10}
Benchmark & \multicolumn{3}{c}{ARC Challenge ($\alpha = 50\%$)} & \multicolumn{3}{c}{ARC Easy ($\alpha = 75\%$)} & \multicolumn{3}{c}{BoolQ  ($\alpha = 80\%$)} \\
Method & \thead{Operating \\ Cost (in MJ)} & \thead{Request. \\ Satisfaction (in \%)} & \thead{Model Call Ratio \\ (L70B/L8B/L1B)} & \thead{Operating \\ Cost (in MJ)} & \thead{Request. \\ Satisfaction (in \%)} & \thead{Model Call Ratio \\ (L70B/L8B/L1B)} & \thead{Operating \\ Cost (in MJ)} & \thead{Request. \\ Satisfaction (in \%)} & \thead{Model Call Ratio \\ (L70B/L8B/L1B)} \\
\cmidrule(lr){1-1}\cmidrule(lr){2-4}\cmidrule(lr){5-7}\cmidrule(lr){8-10}
Llama 3.2 1B only & 0.09$\scriptscriptstyle\pm0.00$ & \textcolor{red}{37.88$\scriptscriptstyle\pm5.39$} & 0\% / 0\% / 100\% & 0.20$\scriptscriptstyle\pm0.00$ & \textcolor{red}{62.76$\scriptscriptstyle\pm5.37$} & 0\% / 0\% / 100\% & 0.14$\scriptscriptstyle\pm0.00$ & \textcolor{red}{69.17$\scriptscriptstyle\pm5.13$} & 0\% / 0\% / 100\% \\
Llama 3.1 8B only & \underline{0.46$\scriptscriptstyle\pm0.00$} & \textcolor{darkgreen}{54.44$\scriptscriptstyle\pm5.54$} & 0\% / 100\% / 0\% & \underline{0.97$\scriptscriptstyle\pm0.00$} & \textcolor{darkgreen}{79.72$\scriptscriptstyle\pm4.47$} & 0\% / 100\% / 0\% & \underline{0.43$\scriptscriptstyle\pm0.00$} & \textcolor{darkgreen}{84.16$\scriptscriptstyle\pm4.06$} & 0\% / 100\% / 0\% \\
Llama 3.3 70B only & 2.35$\scriptscriptstyle\pm0.01$ & \textcolor{darkgreen}{60.84$\scriptscriptstyle\pm5.43$} & 100\% / 0\% / 0\% & 4.05$\scriptscriptstyle\pm0.01$ & \textcolor{darkgreen}{83.12$\scriptscriptstyle\pm4.16$} & 100\% / 0\% / 0\% & 3.40$\scriptscriptstyle\pm0.00$ & \textcolor{darkgreen}{88.78$\scriptscriptstyle\pm3.51$} & 100\% / 0\% / 0\% \\
\cmidrule(lr){1-1}\cmidrule(lr){2-4}\cmidrule(lr){5-7}\cmidrule(lr){8-10}
Educated Guessing & 1.00$\scriptscriptstyle\pm0.09$ & \textcolor{darkgreen}{51.65$\scriptscriptstyle\pm2.98$} & 35\% / 31\% / 34\% & 2.00$\scriptscriptstyle\pm0.08$ & \textcolor{red}{74.00$\scriptscriptstyle\pm4.39$} & 31\% / 32\% / 36\% & 1.31$\scriptscriptstyle\pm0.04$ & \textcolor{darkgreen}{80.47$\scriptscriptstyle\pm1.08$} & 33\% / 34\% / 33\% \\
RouteLLM~\cite{ong2024routellm} & 1.24$\scriptscriptstyle\pm0.10$ & \textcolor{darkgreen}{51.17$\scriptscriptstyle\pm2.93$} & 50\% / 0\% / 50\% & 4.05$\scriptscriptstyle\pm0.01$ & \textcolor{darkgreen}{82.54$\scriptscriptstyle\pm2.12$} & 100\% / 0\% / 0\% & 2.96$\scriptscriptstyle\pm0.04$ & \textcolor{darkgreen}{86.83$\scriptscriptstyle\pm1.27$} & 87\% / 0\% / 13\% \\
RouterDC~\cite{chen2024routerdc} & 2.09$\scriptscriptstyle\pm0.06$ & \textcolor{darkgreen}{60.94$\scriptscriptstyle\pm2.92$} & 88\% / 12\% / 0\% & 3.61$\scriptscriptstyle\pm0.06$ & \textcolor{darkgreen}{82.30$\scriptscriptstyle\pm2.60$} & 85\% / 15\% / 0\% & 2.14$\scriptscriptstyle\pm0.05$ & \textcolor{darkgreen}{87.06$\scriptscriptstyle\pm2.70$} & 58\% / 42\% / 0\% \\
\textbf{MESS+ (ours)} & \textbf{0.83$\scriptscriptstyle\pm0.07$} & \textcolor{darkgreen}{53.64$\scriptscriptstyle\pm3.13$} & 41\% / 41\% / 18\% & \textbf{1.74$\scriptscriptstyle\pm0.06$} & \textcolor{darkgreen}{77.06$\scriptscriptstyle\pm1.76$} & 22\% / 61\% / 18\% & \textbf{0.90$\scriptscriptstyle\pm0.04$} & \textcolor{darkgreen}{82.16$\scriptscriptstyle\pm1.68$} & 31\% / 45\% / 24\% \\
\cmidrule(lr){1-1}\cmidrule(lr){2-4}\cmidrule(lr){5-7}\cmidrule(lr){8-10}
\end{tabular}
    }
    \vspace{0.5em}

\centering
    \resizebox{1.0\textwidth}{!}{
        \begin{tabular}{lccrccrccr}
\cmidrule(lr){1-1}\cmidrule(lr){2-4}\cmidrule(lr){5-7}\cmidrule(lr){8-10}
Benchmark & \multicolumn{3}{c}{LogiQA ($\alpha = 40\%$)} & \multicolumn{3}{c}{PiQA ($\alpha = 78\%$)} & \multicolumn{3}{c}{SciQ ($\alpha = 96\%$)} \\
Method & \thead{Operating \\ Cost (in MJ)} & \thead{Request. \\ Satisfaction (in \%)} & \thead{Model Call Ratio \\ (L70B/L8B/L1B)} & \thead{Operating \\ Cost (in MJ)} & \thead{Request. \\ Satisfaction (in \%)} & \thead{Model Call Ratio \\ (L70B/L8B/L1B)} & \thead{Operating \\ Cost (in MJ)} & \thead{Request. \\ Satisfaction (in \%)} & \thead{Model Call Ratio \\ (L70B/L8B/L1B)} \\
\cmidrule(lr){1-1}\cmidrule(lr){2-4}\cmidrule(lr){5-7}\cmidrule(lr){8-10}
Llama 3.2 1B only & 0.17$\scriptscriptstyle\pm0.00$ & \textcolor{red}{27.19$\scriptscriptstyle\pm4.94$} & 0\% / 0\% / 100\% & 0.07$\scriptscriptstyle\pm0.00$ & \textcolor{red}{74.05$\scriptscriptstyle\pm4.87$} & 0\% / 0\% / 100\% & 0.10$\scriptscriptstyle\pm0.00$ & \textcolor{red}{93.80$\scriptscriptstyle\pm2.68$} & 0\% / 0\% / 100\% \\
Llama 3.1 8B only & 0.81$\scriptscriptstyle\pm0.00$ & \textcolor{red}{29.03$\scriptscriptstyle\pm5.04$} & 0\% / 100\% / 0\% & \underline{0.36$\scriptscriptstyle\pm0.00$} & \textcolor{darkgreen}{79.33$\scriptscriptstyle\pm4.50$} & 0\% / 100\% / 0\% & \underline{0.44$\scriptscriptstyle\pm0.00$} & \textcolor{darkgreen}{97.00$\scriptscriptstyle\pm1.90$} & 0\% / 100\% / 0\% \\
Llama 3.3 70B only & \underline{4.11$\scriptscriptstyle\pm0.02$} & \textcolor{darkgreen}{49.31$\scriptscriptstyle\pm5.56$} & 100\% / 0\% / 0\% & 1.84$\scriptscriptstyle\pm0.01$ & \textcolor{darkgreen}{82.70$\scriptscriptstyle\pm4.20$} & 100\% / 0\% / 0\% & 2.23$\scriptscriptstyle\pm0.02$ & \textcolor{darkgreen}{97.10$\scriptscriptstyle\pm1.87$} & 100\% / 0\% / 0\% \\
\cmidrule(lr){1-1}\cmidrule(lr){2-4}\cmidrule(lr){5-7}\cmidrule(lr){8-10}
Educated Guessing & 2.51$\scriptscriptstyle\pm0.09$ & \textcolor{red}{39.88$\scriptscriptstyle\pm4.57$} & 56\% / 21\% / 22\% & 0.76$\scriptscriptstyle\pm0.04$ & \textcolor{darkgreen}{78.89$\scriptscriptstyle\pm1.52$} & 34\% / 32\% / 34\% & 0.92$\scriptscriptstyle\pm0.09$ & \textcolor{darkgreen}{96.51$\scriptscriptstyle\pm1.49$} & 31\% / 36\% / 32\% \\
RouteLLM~\cite{ong2024routellm} & 3.97$\scriptscriptstyle\pm0.04$ & \textcolor{darkgreen}{47.71$\scriptscriptstyle\pm3.38$} & 98\% / 0\% / 2\% & 1.25$\scriptscriptstyle\pm0.05$ & \textcolor{darkgreen}{78.35$\scriptscriptstyle\pm1.42$} & 66\% / 0\% / 34\% & 2.16$\scriptscriptstyle\pm0.04$ & \textcolor{darkgreen}{97.76$\scriptscriptstyle\pm0.73$} & 95\% / 0\% / 5\% \\
RouterDC~\cite{chen2024routerdc} & 2.67$\scriptscriptstyle\pm0.08$ & \textcolor{darkgreen}{47.13$\scriptscriptstyle\pm3.08$} & 70\% / 29\% / 2\% & 1.85$\scriptscriptstyle\pm0.01$ & \textcolor{darkgreen}{82.34$\scriptscriptstyle\pm1.33$} & 100\% / 0\% / 0\% & 1.90$\scriptscriptstyle\pm0.07$ & \textcolor{darkgreen}{97.95$\scriptscriptstyle\pm0.81$} & 82\% / 18\% / 0\% \\
\textbf{MESS+ (ours)} & \textbf{2.50$\scriptscriptstyle\pm0.09$} & \textcolor{darkgreen}{41.02$\scriptscriptstyle\pm3.59$} & 59\% / 17\% / 23\% & \textbf{0.67$\scriptscriptstyle\pm0.04$} & \textcolor{darkgreen}{79.20$\scriptscriptstyle\pm2.58$} & 35\% / 45\% / 19\% & \textbf{0.83$\scriptscriptstyle\pm0.04$} & \textcolor{darkgreen}{96.01$\scriptscriptstyle\pm2.05$} & 27\% / 39\% / 34\% \\
\cmidrule(lr){1-1}\cmidrule(lr){2-4}\cmidrule(lr){5-7}\cmidrule(lr){8-10}
\end{tabular}
    }
    \vspace{0.5em}

    \centering
    \resizebox{1.0\textwidth}{!}{
        \begin{tabular}{lccrccrccr}
\cmidrule(lr){1-1}\cmidrule(lr){2-4}\cmidrule(lr){5-7}\cmidrule(lr){8-10}
Benchmark & \multicolumn{3}{c}{SocialIQA ($\alpha = 44\%$)} & \multicolumn{3}{c}{Winogrande ($\alpha = 70\%$)} & \multicolumn{3}{c}{\textit{Avg. across all Benchmarks ($\alpha = 66\%$)}} \\
Method & \thead{Operating \\ Cost (in MJ)} & \thead{Request. \\ Satisfaction (in \%)} & \thead{Model Call Ratio \\ (L70B/L8B/L1B)} & \thead{Operating \\ Cost (in MJ)} & \thead{Request. \\ Satisfaction (in \%)} & \thead{Model Call Ratio \\ (L70B/L8B/L1B)} & \thead{Operating \\ Cost (in MJ)} & \thead{Request. \\ Satisfaction (in \%)} & \thead{Model Call Ratio \\ (L70B/L8B/L1B)} \\
\cmidrule(lr){1-1}\cmidrule(lr){2-4}\cmidrule(lr){5-7}\cmidrule(lr){8-10}
Llama 3.2 1B only & 0.13$\scriptscriptstyle\pm0.00$ & \textcolor{red}{41.71$\scriptscriptstyle\pm5.48$} & 0\% / 0\% / 100\% & 0.06$\scriptscriptstyle\pm0.00$ & \textcolor{red}{59.67$\scriptscriptstyle\pm5.45$} & 0\% / 0\% / 100\% & 0.12$\scriptscriptstyle\pm0.00$ & \textcolor{red}{58.28$\scriptscriptstyle\pm4.92$} & 0\% / 0\% / 100\% \\
Llama 3.1 8B only & \underline{0.59$\scriptscriptstyle\pm0.00$} & \textcolor{darkgreen}{48.31$\scriptscriptstyle\pm5.55$} & 0\% / 100\% / 0\% & \underline{0.25$\scriptscriptstyle\pm0.00$} & \textcolor{darkgreen}{73.64$\scriptscriptstyle\pm4.90$} & 0\% / 100\% / 0\% & \underline{0.54$\scriptscriptstyle\pm0.00$} & \textcolor{darkgreen}{68.20$\scriptscriptstyle\pm4.49$} & 0\% / 100\% / 0\% \\
Llama 3.3 70B only & 3.00$\scriptscriptstyle\pm0.00$ & \textcolor{darkgreen}{48.67$\scriptscriptstyle\pm5.56$} & 100\% / 0\% / 0\% & 1.29$\scriptscriptstyle\pm0.00$ & \textcolor{darkgreen}{79.08$\scriptscriptstyle\pm4.52$} & 100\% / 0\% / 0\% & 2.91$\scriptscriptstyle\pm0.01$ & \textcolor{darkgreen}{73.70$\scriptscriptstyle\pm4.35$} & 100\% / 0\% / 0\% \\
\cmidrule(lr){1-1}\cmidrule(lr){2-4}\cmidrule(lr){5-7}\cmidrule(lr){8-10}
Educated Guessing & 1.22$\scriptscriptstyle\pm0.06$ & \textcolor{darkgreen}{47.71$\scriptscriptstyle\pm2.50$} & 33\% / 32\% / 35\% & 0.54$\scriptscriptstyle\pm0.04$ & \textcolor{darkgreen}{70.67$\scriptscriptstyle\pm3.35$} & 35\% / 30\% / 35\% & 1.28$\scriptscriptstyle\pm0.07$ & \textcolor{darkgreen}{67.47$\scriptscriptstyle\pm2.73$} & 36\% / 31\% / 33\% \\
RouteLLM~\cite{ong2024routellm} & 2.02$\scriptscriptstyle\pm0.07$ & \textcolor{darkgreen}{44.32$\scriptscriptstyle\pm2.40$} & 65\% / 0\% / 35\% & 1.27$\scriptscriptstyle\pm0.02$ & \textcolor{darkgreen}{80.82$\scriptscriptstyle\pm2.53$} & 97\% / 0\% / 3\% & 2.04$\scriptscriptstyle\pm0.05$ & \textcolor{darkgreen}{71.19$\scriptscriptstyle\pm2.10$} & 82\% / 0\% / 18\% \\
RouterDC~\cite{chen2024routerdc} & 2.89$\scriptscriptstyle\pm0.03$ & \textcolor{darkgreen}{46.76$\scriptscriptstyle\pm2.62$} & 95\% / 5\% / 0\% & 1.30$\scriptscriptstyle\pm0.00$ & \textcolor{darkgreen}{80.86$\scriptscriptstyle\pm2.49$} & 100\% / 0\% / 0\% & 2.11$\scriptscriptstyle\pm0.04$ & \textcolor{darkgreen}{73.17$\scriptscriptstyle\pm2.32$} & 85\% / 15\% / 0\% \\
\textbf{MESS+ (ours)} & \textbf{0.67$\scriptscriptstyle\pm0.04$} & \textcolor{darkgreen}{45.88$\scriptscriptstyle\pm3.14$} & 22\% / 38\% / 41\% & \textbf{0.52$\scriptscriptstyle\pm0.04$} & \textcolor{darkgreen}{73.57$\scriptscriptstyle\pm3.14$} & 43\% / 40\% / 17\% & \textbf{1.08$\scriptscriptstyle\pm0.05$} & \textcolor{darkgreen}{68.44$\scriptscriptstyle\pm2.63$} & 34\% / 40\% / 26\% \\
\cmidrule(lr){1-1}\cmidrule(lr){2-4}\cmidrule(lr){5-7}\cmidrule(lr){8-10}
\end{tabular}
    }
\end{table}

\subsection{Results}
We divide our evaluations into four main segments and an addendum focused on practical aspects.
First, we evaluate at the overall cost optimality objective. 
Second, we look at the request satisfaction rate.
Third, we explore the control dynamics of $V$. 
Fourth, we explore the routing overhead and how to train the predictor.
We then evaluate the characteristics of \algname by looking into sparse user feedback and increasing the number of models in a zoo.

\textbf{Cost Optimality.}
The main results are shown in \Cref{tab:all_benchmark_evals} where we set individual SLA constraints $\alpha$ per benchmark. 
Our key observation is that, among all the adaptive methods, \textit{\algname is consistently the cost optimal solution for achieving a target request satisfaction rate}, which is due to its effective model choice. 
While the baselines prefer choosing larger and more expensive LLMs from our zoo, our approach tends to rely more on cost effective and smaller models, while providing satisfactory responses at the rate specified by $\alpha$. 
By selecting larger models, the baselines overshoot our SLA requirement at the expense of a notable cost overhead.
Overall, \algname is about $2\times$ more cost efficient than existing model routing techniques and $20\%$ more efficient than our random baseline that knows average benchmark statistics when routing a request. 

\textbf{Request Satisfaction.}
For users, it is key that their requests are getting responses with a guaranteed minimum satisfaction rate, so that they can reliably offload tasks to an AI co-pilot. 
When looking at how precisely an adaptive routing technique approaches the SLA requirement $\alpha$, \textit{\algname usually shows the smallest margin, i.e., with \algname routing the LLM zoo provides responses that are closely matching the SLA requirement and therefore cost optimal}.
All other dynamic routing techniques tend to overshoot $\alpha$ and prefer the strongest model in the zoo to satisfy responses.
Our ``educated guessing'' baseline also undershoots $\alpha$ in some cases, which renders it impractical for minimum service level guarantees, in addition to its requirement of prior statistical knowledge that is usually impractical to obtain.
In general, overshooting $\alpha$ naturally means higher per-request operating cost.
Practically, \algname needs a pre-defined number of steps to converge towards $\alpha$, which is acceptable from an SLA perspective as discussed earlier. 

\begin{figure}
    \vspace{-12pt}
    \centering
    \includegraphics[width=0.9\linewidth]{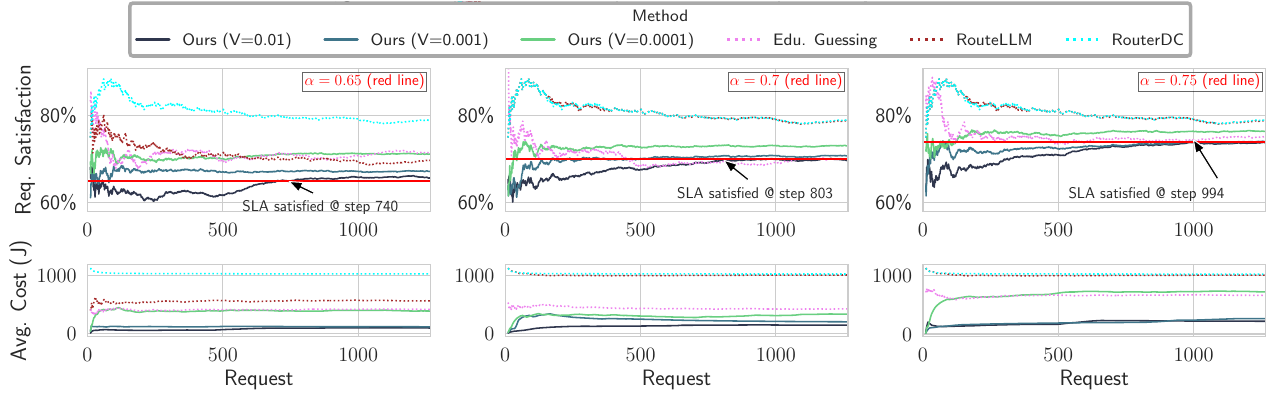} 
    \caption{We run several experiments on the Winogrande benchmark with varying $\alpha$ and $V$ configurations to show the request satisfaction and cost dynamics over time. With \algname, the average request satisfaction rate always converges toward $\alpha$. We further report the first step at which the highest $V$ value satisfies our SLA requirement. Other benchmarks are in the appendix.}
    \label{fig:alpha_v_interplay}
\end{figure}

\begin{wrapfigure}[22]{R}{0.32\linewidth}
    \vspace{-16pt}
    \centering
    \includegraphics[width=0.95\linewidth]{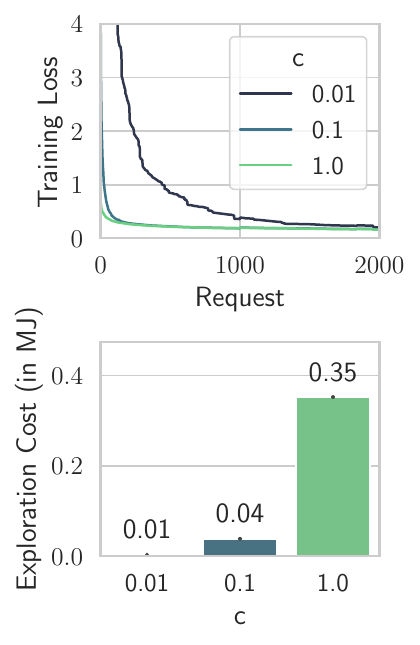}
    \caption{Predictor training performance, averaged across all 8 benchmarks. We control the  exploration probability of \algname with $c$. Our predictor learns effectively with a small $c$ already.}
    \label{fig:c_ablation_study}
\end{wrapfigure}

\textbf{Effect of $V$.}
As discussed in previous sections, $V$ controls the trade-off between the speed of convergence to constraint satisfaction and cost efficiency.
\Cref{fig:alpha_v_interplay} shows that choosing a large $V$ leads to longer convergence times for constraint satisfaction, regardless of the value of $\alpha$. 
Conversely, a small value of $V$ yields constraint satisfaction within a short amount of time at the expense of increased operating costs.
This manifests in the varying operating costs per request over time. 
When comparing to our baselines, we observe that \textit{\algname offers the lowest per-request operating cost among the dynamic routing approaches} by a large margin.

\textbf{Routing Overhead.}
Adaptive routing incurs a cost overhead for routing an incoming request to an LLM within a zoo. 
We measure the cost to make this decision with \algname across benchmarks (\Cref{tab:routing_cost_overhead}). 
For an incoming request, obtaining satisfaction probabilities from \textit{our predictor incurs a minimal cost overhead} of $4.65\%$ on average, compared to the cost for making an inference call to an LLM.
Overall, the cost for making a request satisfaction prediction depends on the length of the incoming request.

\begin{table}
    \centering
    \caption{Routing overhead compared to the average per inference call cost per benchmark. The routing overhead remains well below $10\%$ of the average inference call cost and depend largely on the sequence length of incoming requests. The average inference call costs per benchmark are computed across all LLMs in a zoo.}
    \label{tab:routing_cost_overhead}
    \resizebox{1\textwidth}{!}{
        \begin{tabular}{lccccccccc}
        \toprule
        Metric & \multicolumn{1}{c}{ARC Challenge} & \multicolumn{1}{c}{ARC Easy} & \multicolumn{1}{c}{BoolQ} & \multicolumn{1}{c}{LogiQA} & \multicolumn{1}{c}{PiQA} & \multicolumn{1}{c}{SciQ} & \multicolumn{1}{c}{SocialIQA} & \multicolumn{1}{c}{Winogrande} & \multicolumn{1}{c}{Avg.} \\
        \midrule
        Avg. Predictor Cost (J) & 5.75$\scriptscriptstyle\pm0.96$ & 5.69$\scriptscriptstyle\pm0.56$ & 16.80$\scriptscriptstyle\pm0.62$ & 41.13$\scriptscriptstyle\pm0.88$ & 15.38$\scriptscriptstyle\pm0.70$ & 15.12$\scriptscriptstyle\pm0.85$ & 26.67$\scriptscriptstyle\pm0.60$ & 4.89$\scriptscriptstyle\pm0.72$ & 16.43$\scriptscriptstyle\pm0.74$ \\
        Avg. LLM Call Cost (J) & 589.97$\scriptscriptstyle\pm100.18$ & 516.07$\scriptscriptstyle\pm95.29$ & 236.17$\scriptscriptstyle\pm39.93$ & 833.21$\scriptscriptstyle\pm144.34$ & 273.44$\scriptscriptstyle\pm44.30$ & 263.64$\scriptscriptstyle\pm48.86$ & 304.45$\scriptscriptstyle\pm53.03$ & 300.56$\scriptscriptstyle\pm55.07$ & 414.69$\scriptscriptstyle\pm72.63$ \\
        \midrule
        Prediction Overhead (\%) & 1.01$\scriptscriptstyle\pm0.28$ & 1.15$\scriptscriptstyle\pm0.28$ & 7.35$\scriptscriptstyle\pm1.56$ & 5.12$\scriptscriptstyle\pm1.19$ & 5.82$\scriptscriptstyle\pm1.67$ & 5.95$\scriptscriptstyle\pm1.26$ & 9.08$\scriptscriptstyle\pm1.96$ & 1.69$\scriptscriptstyle\pm0.47$ & 4.65$\scriptscriptstyle\pm1.08$ \\
        \bottomrule
        \end{tabular}
    }
\end{table}

\textbf{Predictor Training.}
As \algname requires online learning of a predictor that predicts the request satisfaction for every LLM in the zoo, we study how the rate of exploration that is parameterized with $c$ affects both the predictor training convergence and cost overhead incurred by exploration and training. 
Our study shows that \textit{a relatively small value of $c$ already leads to a strong predictor within a short amount of time} (\Cref{fig:c_ablation_study}), as shown by the training loss. 
The lower we choose $c$ the faster we move from exploring the model zoo in the beginning towards using the \algname objective \eqref{eq:per-request-optimization} for model selection decision making. 

\begin{table}[t]
    \centering
    \caption{\algname provides strong performance even when users provide sparse feedback (i.e., only when a user sends a feedback signal).}
    \label{tab:sparse_feedback}
    \resizebox{\textwidth}{!}{
        \begin{tabular}{lcccccccc}
        \cmidrule(lr){1-1}\cmidrule(lr){2-3}\cmidrule(lr){4-5}\cmidrule(lr){6-7}\cmidrule(lr){8-9}
         Benchmark & \multicolumn{2}{c}{ARC C. ($\alpha = 50\%$)} & \multicolumn{2}{c}{ARC E. ($\alpha = 75\%$)} & \multicolumn{2}{c}{BoolQ ($\alpha = 20\%$)} & \multicolumn{2}{c}{LogiQA ($\alpha = 40\%$)} \\
         \textit{Feedback Density} & $20\%$ & $100\%$ & $20\%$ & $100\%$ & $20\%$ & $100\%$ & $20\%$ & $100\%$ \\\cmidrule(lr){1-1}\cmidrule(lr){2-3}\cmidrule(lr){4-5}\cmidrule(lr){6-7}\cmidrule(lr){8-9}
        Request Satisfaction (in \%) & 50.20$\scriptscriptstyle\pm5.39$ & 50.64$\scriptscriptstyle\pm2.28$ & 75.10$\scriptscriptstyle\pm1.12$ & 75.66$\scriptscriptstyle\pm0.34$ & 20.66$\scriptscriptstyle\pm3.17$ & 20.34$\scriptscriptstyle\pm1.78$ & 40.22$\scriptscriptstyle\pm0.86$ & 40.46$\scriptscriptstyle\pm0.09$ \\
        Cost (in MJ) & 0.79$\scriptscriptstyle\pm0.01$ & 0.83$\scriptscriptstyle\pm0.01$ & 1.69$\scriptscriptstyle\pm0.01$ & 1.74$\scriptscriptstyle\pm0.01$ & 0.87$\scriptscriptstyle\pm0.01$ & 0.90$\scriptscriptstyle\pm0.01$ & 2.47$\scriptscriptstyle\pm0.02$ & 2.50$\scriptscriptstyle\pm0.01$ \\\cmidrule(lr){1-1}\cmidrule(lr){2-3}\cmidrule(lr){4-5}\cmidrule(lr){6-7}\cmidrule(lr){8-9}
        \end{tabular}
    }

    \resizebox{\textwidth}{!}{
        \begin{tabular}{lcccccccc}
            \cmidrule(lr){1-1}\cmidrule(lr){2-3}\cmidrule(lr){4-5}\cmidrule(lr){6-7}\cmidrule(lr){8-9}
            Benchmark & \multicolumn{2}{c}{PiQA ($\alpha = 78\%$)} & \multicolumn{2}{c}{SciQ ($\alpha = 96\%$)} & \multicolumn{2}{c}{SocialIQA ($\alpha = 44\%$)} & \multicolumn{2}{c}{Winogrande ($\alpha = 70\%$)} \\
            \textit{Feedback Density} & $20\%$ & $100\%$ & $20\%$ & $100\%$ & $20\%$ & $100\%$ & $20\%$ & $100\%$ \\ \cmidrule(lr){1-1}\cmidrule(lr){2-3}\cmidrule(lr){4-5}\cmidrule(lr){6-7}\cmidrule(lr){8-9}
            Request Satisfaction (in \%) & 78.48$\scriptscriptstyle\pm1.18$ & 78.91$\scriptscriptstyle\pm0.83$ & 96.08$\scriptscriptstyle\pm1.13$ & 96.11$\scriptscriptstyle\pm0.62$ & 44.24$\scriptscriptstyle\pm1.42$ & 44.53$\scriptscriptstyle\pm0.91$ & 70.97$\scriptscriptstyle\pm1.71$ & 70.36$\scriptscriptstyle\pm1.03$ \\
            Cost (in MJ) & 0.64$\scriptscriptstyle\pm0.01$ & 0.67$\scriptscriptstyle\pm0.01$ & 0.81$\scriptscriptstyle\pm0.01$ & 0.83$\scriptscriptstyle\pm0.01$ & 0.66$\scriptscriptstyle\pm0.01$ & 0.67$\scriptscriptstyle\pm0.01$ & 0.50$\scriptscriptstyle\pm0.01$ & 0.52$\scriptscriptstyle\pm0.01$ \\ \cmidrule(lr){1-1}\cmidrule(lr){2-3}\cmidrule(lr){4-5}\cmidrule(lr){6-7}\cmidrule(lr){8-9}
            
        \end{tabular}
    }
    
\end{table}

\textbf{Sparse User Feedback.} 
\algname relies on online user feedback. We evaluate how the performance of \algname varies when only a fraction of users actually provides feedback for requests that would normally be used to train our predictor. 
We compare perfect conditions (full feedback) to sparse feedback with $20\%$ of requests receiving user feedback. 
Our approach maintains strong performance when introducing sparsity (\Cref{tab:sparse_feedback}).
\begin{wraptable}[14]{r}{0.5\textwidth}
    \caption{
        Main results with a larger model zoo containing models from the Qwen 2/2.5 family. 
        \algname scales well as the number of models grows and shows effective routing capabilities.
        Detailed results can be found in the appendix.
    }
    \label{tab:qwen2_scaled_zoo}
    \resizebox{0.5\textwidth}{!}{

        \begin{tabular}{lccccccccc}
        \cmidrule(lr){1-1}\cmidrule(lr){2-4}
        Category & \multicolumn{3}{c}{Mean ($\alpha = 67\%$)} \\
        Subcategory & \thead{Operating \\ Cost} & \thead{Request. \\ Satisfaction} & \thead{Model Call Ratio \\ (Q32B/Q7B/Q1.5B/Q0.5B)} \\
        \cmidrule(lr){1-1}\cmidrule(lr){2-4}
        Qwen2 0.5B & 0.12$\scriptscriptstyle\pm0.00$ & \textcolor{red}{54.12$\scriptscriptstyle\pm45.15$} & 0\% / 0\% / 0\% / 100\% \\
        Qwen2 1.5B & 0.16$\scriptscriptstyle\pm0.00$ & \textcolor{red}{61.13$\scriptscriptstyle\pm4.79$} & 0\% / 0\% / 100\% / 0\% \\
        Qwen2 7B & \underline{0.40$\scriptscriptstyle\pm0.00$} & \textcolor{darkgreen}{67.07$\scriptscriptstyle\pm4.61$} & 0\% / 100\% / 0\% / 0\% \\
        Qwen2.5 32B & 1.60$\scriptscriptstyle\pm0.00$ & \textcolor{darkgreen}{70.91$\scriptscriptstyle\pm4.48$} & 100\% / 0\% / 0\% / 0\% \\
        Educated Guessing & 0.99$\scriptscriptstyle\pm0.00$ & \textcolor{darkgreen}{67.02$\scriptscriptstyle\pm2.32$} & 53\% / 26\% / 11\% / 10\% \\
        
        RouteLLM & 1.37$\scriptscriptstyle\pm0.00$ & \textcolor{darkgreen}{69.01$\scriptscriptstyle\pm2.31$} & 83\% / 0\% / 0\% / 17\% \\
        
        RouterDC & 1.13$\scriptscriptstyle\pm0.00$ & \textcolor{darkgreen}{69.17$\scriptscriptstyle\pm2.47$} & 63\% / 17\% / 9\% / 11\% \\
        
        \textbf{MESS+ (ours)} & \textbf{0.84$\scriptscriptstyle\pm0.00$} & \textcolor{darkgreen}{67.55$\scriptscriptstyle\pm3.23$} & 48\% / 26\% / 10\% / 16\% \\
        \cmidrule(lr){1-1}\cmidrule(lr){2-4}
        \end{tabular}
    }
    \vspace{-12pt}
\end{wraptable}
Over time, both the dense feedback and sparse scenario provide strict SLA compliance and incur similar operating cost. 
The operating cost for the sparse feedback are slightly lower than with dense feedback. 
This is because of $Q$ receiving fewer updates and a prolonged time to stabilize, which leads to a higher favorability of smaller and cheaper models in the zoo in the beginning (avg. $1.05~\textrm{MJ}$ vs. $1.08~\textrm{MJ}$ across 8 benchmarks).

\textbf{Larger Model Zoo.} 
To demonstrate the scalability of \algname, we deploy a larger model zoo containing four models: Qwen 2.5 32B (Q32B), Qwen 2 7B (Q7B), Qwen 2 1.5B (Q1.5B), and Qwen 2 0.5B (Q0.5B).
The zoo not only is larger but also contains models that have more similar cost characteristics than our LLama3 model zoo. 
This makes routing in our case more difficult.
Our approach exhibits strong performance and strictly maintains SLA compliance under these more challenging conditions (\Cref{tab:qwen2_scaled_zoo}).
Specifically, our routing approach outperforms existing baselines in terms of cost efficiency by a factor of up to $1.6\times$, demonstrating the suitability of \algname for providing the most appropriate models for any incoming user request even in larger model zoos.

\vspace{-6pt}
\section{Conclusion}
\vspace{-6pt}
\label{sec:conclusions}
We have presented \algname, which is a novel theoretically grounded method for automatic model selection in language model zoos. 
On average, our approach reduces operating costs by $2\times$ compared to well-established adaptive routing baselines in the literature, while still satisfying SLA requirements.
Overall, \algname shows strong generalization across various state-of-the-art benchmarks and is the first approach to optimizing operating costs while providing rigorous service level guarantees, when serving user requests with a model zoo.
Further, our online-learning based approach to model selection removes the need for curating routing preference datasets.
Our technique can be particularly useful for cost-aware systems that serve quality-sensitive workloads and require a lower-bound on user satisfaction rate.

While our work shows strong performance across a wide range of tasks, we observe some practical limitations that can be addressed in future work. 
Our approach expects readily available user satisfaction labels for any request. 
In reality, user feedback may be only sparingly available. 
This would require a modified strategy for virtual queue update and zoo exploration based on feedback availability. 
In addition, our predictor training approach expects complete request satisfaction labels for each model during exploration. 
Practically, this can be challenging as showing multiple outputs to users and requesting feedback may be confusing.
Nevertheless, it is worth pointing out that our work serves as an important foundation for these extensions in the future.

\section*{Acknowledgements}
\vspace{-8pt}
This work is supported by the Bavarian Ministry of Economic Affairs, Regional Development and Energy (Grant: DIK0446/01).

\bibliography{main}

\begin{thebibliography}{26}
\providecommand{\natexlab}[1]{#1}
\providecommand{\url}[1]{\texttt{#1}}
\expandafter\ifx\csname urlstyle\endcsname\relax
  \providecommand{\doi}[1]{doi: #1}\else
  \providecommand{\doi}{doi: \begingroup \urlstyle{rm}\Url}\fi

\bibitem[Aggarwal et~al.(2024)Aggarwal, Madaan, Anand, Potharaju, Mishra, Zhou, Gupta, Rajagopal, Kappaganthu, Yang, Upadhyay, Faruqui, and .]{aggarwal2024automix}
P.~Aggarwal, A.~Madaan, A.~Anand, S.~P. Potharaju, S.~Mishra, P.~Zhou, A.~Gupta, D.~Rajagopal, K.~Kappaganthu, Y.~Yang, S.~Upadhyay, M.~Faruqui, and M.~.
\newblock Automix: Automatically mixing language models.
\newblock In \emph{The Thirty-eighth Annual Conference on Neural Information Processing Systems}, 2024.
\newblock URL \url{https://openreview.net/forum?id=e6WrwIvgzX}.

\bibitem[Bisk et~al.(2020)Bisk, Zellers, Le~bras, Gao, and Choi]{Bisk2020}
Y.~Bisk, R.~Zellers, R.~Le~bras, J.~Gao, and Y.~Choi.
\newblock Piqa: Reasoning about physical commonsense in natural language.
\newblock \emph{Proceedings of the AAAI Conference on Artificial Intelligence}, 34\penalty0 (05):\penalty0 7432–7439, Apr. 2020.
\newblock ISSN 2159-5399.
\newblock \doi{10.1609/aaai.v34i05.6239}.
\newblock URL \url{http://dx.doi.org/10.1609/aaai.v34i05.6239}.

\bibitem[Chen et~al.(2024)Chen, Jiang, Lin, Kwok, and Zhang]{chen2024routerdc}
S.~Chen, W.~Jiang, B.~Lin, J.~Kwok, and Y.~Zhang.
\newblock Router{DC}: Query-based router by dual contrastive learning for assembling large language models.
\newblock In \emph{The Thirty-eighth Annual Conference on Neural Information Processing Systems}, 2024.
\newblock URL \url{https://openreview.net/forum?id=7RQvjayHrM}.

\bibitem[Clark et~al.(2019)Clark, Lee, Chang, Kwiatkowski, Collins, and Toutanova]{clark2019boolq}
C.~Clark, K.~Lee, M.-W. Chang, T.~Kwiatkowski, M.~Collins, and K.~Toutanova.
\newblock Boolq: Exploring the surprising difficulty of natural yes/no questions.
\newblock In \emph{NAACL}, 2019.

\bibitem[Clark et~al.(2018)Clark, Cowhey, Etzioni, Khot, Sabharwal, Schoenick, and Tafjord]{allenai:arc}
P.~Clark, I.~Cowhey, O.~Etzioni, T.~Khot, A.~Sabharwal, C.~Schoenick, and O.~Tafjord.
\newblock Think you have solved question answering? try arc, the ai2 reasoning challenge.
\newblock \emph{arXiv:1803.05457v1}, 2018.

\bibitem[{Council of the European Union}(2024)]{eu_ai_act}
{Council of the European Union}.
\newblock {Regulation (EU) 2024/1689 of the European Parliament and of the Council of 13 June 2024 laying down harmonised rules on artificial intelligence and amending Regulations (EC) No 300/2008, (EU) No 167/2013, (EU) No 168/2013, (EU) 2018/858, (EU) 2018/1139 and (EU) 2019/2144 and Directives 2014/90/EU, (EU) 2016/797 and (EU) 2020/1828 (Artificial Intelligence Act)Text with EEA relevance.}, jul 2024.
\newblock URL \url{https://eur-lex.europa.eu/eli/reg/2024/1689/oj}.
\newblock Document 32024R1689.

\bibitem[Ding et~al.(2024)Ding, Mallick, Wang, Sim, Mukherjee, R{\"u}hle, Lakshmanan, and Awadallah]{ding2024hybrid}
D.~Ding, A.~Mallick, C.~Wang, R.~Sim, S.~Mukherjee, V.~R{\"u}hle, L.~V.~S. Lakshmanan, and A.~H. Awadallah.
\newblock Hybrid {LLM}: Cost-efficient and quality-aware query routing.
\newblock In \emph{The Twelfth International Conference on Learning Representations}, 2024.
\newblock URL \url{https://openreview.net/forum?id=02f3mUtqnM}.

\bibitem[Dubey et~al.(2024)Dubey, Jauhri, et~al.]{llama3_paper}
A.~Dubey, A.~Jauhri, et~al.
\newblock The {Llama} 3 herd of models, 2024.
\newblock URL \url{https://arxiv.org/abs/2407.21783}.

\bibitem[Fourrier et~al.(2024)Fourrier, Habib, Lozovskaya, Szafer, and Wolf]{open-llm-leaderboard-v2}
C.~Fourrier, N.~Habib, A.~Lozovskaya, K.~Szafer, and T.~Wolf.
\newblock Open llm leaderboard v2.
\newblock \url{https://huggingface.co/spaces/open-llm-leaderboard/open_llm_leaderboard}, 2024.

\bibitem[Gao et~al.(2021)Gao, Tow, et~al.]{eval-harness}
L.~Gao, J.~Tow, et~al.
\newblock A framework for few-shot language model evaluation, Sept. 2021.
\newblock URL \url{https://doi.org/10.5281/zenodo.5371628}.

\bibitem[Granite~Team(2024)]{granitegranite}
I.~Granite~Team.
\newblock Granite 3.0 language models, Oct. 2024.
\newblock URL \url{https://github.com/ibm-granite/granite-3.0-language-models/blob/main/paper.pdf}.

\bibitem[Jiang et~al.(2023)Jiang, Ren, and Lin]{llm_blender}
D.~Jiang, X.~Ren, and B.~Y. Lin.
\newblock {LLM}-blender: Ensembling large language models with pairwise ranking and generative fusion.
\newblock In A.~Rogers, J.~Boyd-Graber, and N.~Okazaki, editors, \emph{Proceedings of the 61st Annual Meeting of the Association for Computational Linguistics (Volume 1: Long Papers)}, pages 14165--14178, Toronto, Canada, July 2023. Association for Computational Linguistics.
\newblock \doi{10.18653/v1/2023.acl-long.792}.
\newblock URL \url{https://aclanthology.org/2023.acl-long.792/}.

\bibitem[Johannes~Welbl(2017)]{SciQ}
M.~G. Johannes~Welbl, Nelson F.~Liu.
\newblock Crowdsourcing multiple choice science questions.
\newblock 2017.

\bibitem[Liu et~al.(2020)Liu, Cui, Liu, Huang, Wang, and Zhang]{Liu2020}
J.~Liu, L.~Cui, H.~Liu, D.~Huang, Y.~Wang, and Y.~Zhang.
\newblock Logiqa: A challenge dataset for machine reading comprehension with logical reasoning.
\newblock In \emph{Proceedings of the Twenty-Ninth International Joint Conference on Artificial Intelligence}, IJCAI-PRICAI-2020, page 3622–3628. International Joint Conferences on Artificial Intelligence Organization, July 2020.
\newblock \doi{10.24963/ijcai.2020/501}.
\newblock URL \url{http://dx.doi.org/10.24963/ijcai.2020/501}.

\bibitem[Lu et~al.(2024)Lu, Yuan, Lin, Lin, Yuan, Zhou, and Zhou]{zooter}
K.~Lu, H.~Yuan, R.~Lin, J.~Lin, Z.~Yuan, C.~Zhou, and J.~Zhou.
\newblock Routing to the expert: Efficient reward-guided ensemble of large language models.
\newblock In K.~Duh, H.~Gomez, and S.~Bethard, editors, \emph{Proceedings of the 2024 Conference of the North American Chapter of the Association for Computational Linguistics: Human Language Technologies (Volume 1: Long Papers)}, pages 1964--1974, Mexico City, Mexico, June 2024. Association for Computational Linguistics.
\newblock \doi{10.18653/v1/2024.naacl-long.109}.
\newblock URL \url{https://aclanthology.org/2024.naacl-long.109/}.

\bibitem[Naveed et~al.(2023)Naveed, Khan, Qiu, Saqib, Anwar, Usman, Akhtar, Barnes, and Mian]{Humza2023}
H.~Naveed, A.~U. Khan, S.~Qiu, M.~Saqib, S.~Anwar, M.~Usman, N.~Akhtar, N.~Barnes, and A.~Mian.
\newblock A comprehensive overview of large language models, 2023.
\newblock URL \url{https://arxiv.org/abs/2307.06435}.

\bibitem[Neely(2022)]{neely2022stochastic}
M.~Neely.
\newblock \emph{Stochastic network optimization with application to communication and queueing systems}.
\newblock Springer Nature, 2022.

\bibitem[Ong et~al.(2024)Ong, Almahairi, Wu, Chiang, Wu, Gonzalez, Kadous, and Stoica]{ong2024routellm}
I.~Ong, A.~Almahairi, V.~Wu, W.-L. Chiang, T.~Wu, J.~E. Gonzalez, M.~W. Kadous, and I.~Stoica.
\newblock Routellm: Learning to route llms with preference data.
\newblock \emph{arXiv preprint arXiv:2406.18665}, 2024.

\bibitem[{Qwen} et~al.(2024){Qwen}, Yang, Yang, Zhang, Hui, Zheng, Yu, Li, Liu, Huang, Wei, Lin, Yang, Tu, Zhang, Yang, Yang, Zhou, Lin, Dang, Lu, Bao, Yang, Yu, Li, Xue, Zhang, Zhu, Men, Lin, Li, Tang, Xia, Ren, Ren, Fan, Su, Zhang, Wan, Liu, Cui, Zhang, and Qiu]{qwen2.5_paper}
{Qwen}, A.~Yang, B.~Yang, B.~Zhang, B.~Hui, B.~Zheng, B.~Yu, C.~Li, D.~Liu, F.~Huang, H.~Wei, H.~Lin, J.~Yang, J.~Tu, J.~Zhang, J.~Yang, J.~Yang, J.~Zhou, J.~Lin, K.~Dang, K.~Lu, K.~Bao, K.~Yang, L.~Yu, M.~Li, M.~Xue, P.~Zhang, Q.~Zhu, R.~Men, R.~Lin, T.~Li, T.~Tang, T.~Xia, X.~Ren, X.~Ren, Y.~Fan, Y.~Su, Y.~Zhang, Y.~Wan, Y.~Liu, Z.~Cui, Z.~Zhang, and Z.~Qiu.
\newblock Qwen2.5 technical report, 2024.
\newblock URL \url{https://arxiv.org/abs/2412.15115}.

\bibitem[Reuters(2024)]{reuters2024}
Reuters.
\newblock Microsoft deal propels three mile island restart, with key permits still needed.
\newblock \url{https://www.reuters.com/markets/deals/constellation-inks-power-supply-deal-with-microsoft-2024-09-20/}, 2024.
\newblock [Accessed 24-09-2024].

\bibitem[Sakaguchi et~al.(2020)Sakaguchi, Le~Bras, Bhagavatula, and Choi]{Sakaguchi2020}
K.~Sakaguchi, R.~Le~Bras, C.~Bhagavatula, and Y.~Choi.
\newblock Winogrande: An adversarial winograd schema challenge at scale.
\newblock \emph{Proceedings of the AAAI Conference on Artificial Intelligence}, 34\penalty0 (05):\penalty0 8732–8740, Apr. 2020.
\newblock ISSN 2159-5399.
\newblock \doi{10.1609/aaai.v34i05.6399}.
\newblock URL \url{http://dx.doi.org/10.1609/aaai.v34i05.6399}.

\bibitem[Samsi et~al.(2023)Samsi, Zhao, McDonald, Li, Michaleas, Jones, Bergeron, Kepner, Tiwari, and Gadepally]{Samsi2023}
S.~Samsi, D.~Zhao, J.~McDonald, B.~Li, A.~Michaleas, M.~Jones, W.~Bergeron, J.~Kepner, D.~Tiwari, and V.~Gadepally.
\newblock From words to watts: Benchmarking the energy costs of large language model inference.
\newblock In \emph{2023 IEEE High Performance Extreme Computing Conference (HPEC)}, pages 1--9, 2023.
\newblock \doi{10.1109/HPEC58863.2023.10363447}.

\bibitem[Sap et~al.(2019)Sap, Rashkin, Chen, LeBras, and Choi]{sap2019}
M.~Sap, H.~Rashkin, D.~Chen, R.~LeBras, and Y.~Choi.
\newblock Socialiqa: Commonsense reasoning about social interactions, 2019.
\newblock URL \url{https://arxiv.org/abs/1904.09728}.

\bibitem[Stripelis et~al.(2024)Stripelis, Xu, Hu, Shah, Jin, Yao, Zhang, Zhang, Avestimehr, and He]{stripelis-etal-2024-tensoropera}
D.~Stripelis, Z.~Xu, Z.~Hu, A.~D. Shah, H.~Jin, Y.~Yao, J.~Zhang, T.~Zhang, S.~Avestimehr, and C.~He.
\newblock {T}ensor{O}pera router: A multi-model router for efficient {LLM} inference.
\newblock In F.~Dernoncourt, D.~Preo{\c{t}}iuc-Pietro, and A.~Shimorina, editors, \emph{Proceedings of the 2024 Conference on Empirical Methods in Natural Language Processing: Industry Track}, pages 452--462, Miami, Florida, US, Nov. 2024. Association for Computational Linguistics.
\newblock \doi{10.18653/v1/2024.emnlp-industry.34}.
\newblock URL \url{https://aclanthology.org/2024.emnlp-industry.34/}.

\bibitem[Warner et~al.(2024)Warner, Chaffin, Clavié, Weller, Hallström, Taghadouini, Gallagher, Biswas, Ladhak, Aarsen, Cooper, Adams, Howard, and Poli]{modernbert}
B.~Warner, A.~Chaffin, B.~Clavié, O.~Weller, O.~Hallström, S.~Taghadouini, A.~Gallagher, R.~Biswas, F.~Ladhak, T.~Aarsen, N.~Cooper, G.~Adams, J.~Howard, and I.~Poli.
\newblock Smarter, better, faster, longer: A modern bidirectional encoder for fast, memory efficient, and long context finetuning and inference, 2024.
\newblock URL \url{https://arxiv.org/abs/2412.13663}.

\bibitem[Wettig et~al.(2024)Wettig, Gupta, Malik, and Chen]{pmlr-v235-wettig24a}
A.~Wettig, A.~Gupta, S.~Malik, and D.~Chen.
\newblock {Q}u{R}ating: Selecting high-quality data for training language models.
\newblock In R.~Salakhutdinov, Z.~Kolter, K.~Heller, A.~Weller, N.~Oliver, J.~Scarlett, and F.~Berkenkamp, editors, \emph{Proceedings of the 41st International Conference on Machine Learning}, volume 235 of \emph{Proceedings of Machine Learning Research}, pages 52915--52971. PMLR, 21--27 Jul 2024.
\newblock URL \url{https://proceedings.mlr.press/v235/wettig24a.html}.

\end{thebibliography}
\bibliographystyle{abbrvnat}

\newpage
\section*{NeurIPS Paper Checklist}

\begin{enumerate}

\item {\bf Claims}
    \item[] Question: Do the main claims made in the abstract and introduction accurately reflect the paper's contributions and scope?
    \item[] Answer: \answerYes{} 
    \item[] Justification: We propose a new algorithm that introduces guarantees for minimum user satisfaction rates in language model zoos while optimizing for operating cost, which can be practical for inference endpoint services.

\item {\bf Limitations}
    \item[] Question: Does the paper discuss the limitations of the work performed by the authors?
    \item[] Answer: \answerYes{} 
    \item[] Justification: The limitations are discussed at the end of the conclusion section (\Cref{sec:conclusions}).

\item {\bf Theory assumptions and proofs}
    \item[] Question: For each theoretical result, does the paper provide the full set of assumptions and a complete (and correct) proof?
    \item[] Answer: \answerYes{} 
    \item[] Justification: We provide formal proofs for all our theorems in the supplementary material.

    \item {\bf Experimental result reproducibility}
    \item[] Question: Does the paper fully disclose all the information needed to reproduce the main experimental results of the paper to the extent that it affects the main claims and/or conclusions of the paper (regardless of whether the code and data are provided or not)?
    \item[] Answer: \answerYes{} 
    \item[] Justification: Our codebase will be made public (for now: anonymized for review) and is based on the widely used LM-Eval harness for better usability by others. We describe our algorithm in detail in the main paper and provide extensive details on the technical implementation in the appendix. 

\item {\bf Open access to data and code}
    \item[] Question: Does the paper provide open access to the data and code, with sufficient instructions to faithfully reproduce the main experimental results, as described in supplemental material?
    \item[] Answer: \answerYes{} 
    \item[] Justification: All code repositories and benchmarks we use are open-source. Our code (incl. that to reproduce our data) will be made public, if accepted. 

\item {\bf Experimental setting/details}
    \item[] Question: Does the paper specify all the training and test details (e.g., data splits, hyperparameters, how they were chosen, type of optimizer, etc.) necessary to understand the results?
    \item[] Answer: \answerYes{} 
    \item[] Justification: We provide a high-level overview of the experimental setup in the main paper and discuss all remaining details in the appendix.

\item {\bf Experiment statistical significance}
    \item[] Question: Does the paper report error bars suitably and correctly defined or other appropriate information about the statistical significance of the experiments?
    \item[] Answer: \answerYes{} 
    \item[] Justification: Our experiments have been repeated multiple times and all reported results are based on the mean observations.

\item {\bf Experiments compute resources}
    \item[] Question: For each experiment, does the paper provide sufficient information on the computer resources (type of compute workers, memory, time of execution) needed to reproduce the experiments?
    \item[] Answer: \answerYes{} 
    \item[] Justification: We provide full details of our hardware setup in the appendix.
    
\item {\bf Code of ethics}
    \item[] Question: Does the research conducted in the paper conform, in every respect, with the NeurIPS Code of Ethics \url{https://neurips.cc/public/EthicsGuidelines}?
    \item[] Answer: \answerYes{} 
    \item[] Justification: Our work adheres to the NeurIPS ethical guidelines.

\item {\bf Broader impacts}
    \item[] Question: Does the paper discuss both potential positive societal impacts and negative societal impacts of the work performed?
    \item[] Answer: \answerYes{} 
    \item[] Justification: Our work focuses on cost efficiency. In case of language models, this can have substantial positive effects on the energy consumption, which reduces the environmental burden of AI. In all other regards, our work bears the same risks and safeguards as the underlying models used in the language model zoo. 
    
\item {\bf Safeguards}
    \item[] Question: Does the paper describe safeguards that have been put in place for responsible release of data or models that have a high risk for misuse (e.g., pretrained language models, image generators, or scraped datasets)?
    \item[] Answer: \answerNA{} 
    \item[] Justification: Our work provides the same safeguards as those implemented in the models in our language model zoo.

\item {\bf Licenses for existing assets}
    \item[] Question: Are the creators or original owners of assets (e.g., code, data, models), used in the paper, properly credited and are the license and terms of use explicitly mentioned and properly respected?
    \item[] Answer: \answerYes{} 
    \item[] Justification: We have credited the authors of all code libraries, datasets, and models used in our paper.

\item {\bf New assets}
    \item[] Question: Are new assets introduced in the paper well documented and is the documentation provided alongside the assets?
    \item[] Answer: \answerYes{} 
    \item[] Justification: Our code repository will be made openly available once the paper is accepted.

\item {\bf Crowdsourcing and research with human subjects}
    \item[] Question: For crowdsourcing experiments and research with human subjects, does the paper include the full text of instructions given to participants and screenshots, if applicable, as well as details about compensation (if any)? 
    \item[] Answer: \answerNA{} 
    \item[] Justification: --

\item {\bf Institutional review board (IRB) approvals or equivalent for research with human subjects}
    \item[] Question: Does the paper describe potential risks incurred by study participants, whether such risks were disclosed to the subjects, and whether Institutional Review Board (IRB) approvals (or an equivalent approval/review based on the requirements of your country or institution) were obtained?
    \item[] Answer: \answerNA{} 
    \item[] Justification: --

\item {\bf Declaration of LLM usage}
    \item[] Question: Does the paper describe the usage of LLMs if it is an important, original, or non-standard component of the core methods in this research? Note that if the LLM is used only for writing, editing, or formatting purposes and does not impact the core methodology, scientific rigorousness, or originality of the research, declaration is not required.
    \item[] Answer: \answerNA{} 
    \item[] Justification: We have only used LLMs to improve writing on a sentence level.

\end{enumerate}

\newpage
\appendix

\begin{center}
    {\bf\Large Appendix}
\end{center}

\startcontents[sections]
\printcontents[sections]{l}{1}{\setcounter{tocdepth}{2}}
\newpage

\setcounter{section}{0}
\renewcommand\thesection{\Alph{section}}
\numberwithin{equation}{section}
\counterwithin{figure}{section}
\counterwithin{algocf}{section}
\counterwithin{theorem}{section}
\counterwithin{lemma}{section}
\counterwithin{remark}{section}

\section{Proofs}

\subsection{Proof of \texorpdfstring{\Cref{theorem:max_queue_length}}{Theorem \ref{theorem:max_queue_length}}}

Let $\gamma := \max\left\{\alpha \psi; \frac{V\Delta_E}{\beta}\right\}$. For any given $t$, let $m^*$ denote the selected model obtained from the solution to \eqref{eq:per-request-optimization}, i.e., $y_{m^*,t}=1$ and  $y_{m',t}=0$ for $m'\neq m^*$. 

We first prove that, when $Q_t > \gamma$, we have $\hat{s}_{\tilde{m},t} - \hat{s}_{m^*,t} \leq \beta$, where $\tilde{m} := \arg\max_m \hat{s}_{m,t}$ as defined in  \Cref{assumption:queue_length}. The proof is by contradiction. Suppose $\hat{s}_{\tilde{m},t} - \hat{s}_{m^*,t} > \beta$, then choosing $\tilde{m}$ instead of $m^*$, i.e., letting 
$y_{\tilde{m},t}=1$ and $y_{m^*,t}=0$, will reduce the second term of the objective \eqref{eq:per-request_obj}  by $Q_t \beta$. The increase in the first term of  \eqref{eq:per-request_obj}  due to this change is at most $V\Delta_E$. Since $Q_t > \gamma \geq \frac{V\Delta_E}{\beta}$, we know that this alternative model choice decreases the objective \eqref{eq:per-request_obj}, which contradicts that $m^*$ is the optimal model choice from \eqref{eq:per-request-optimization}.

From  \Cref{assumption:queue_length}, we then have $\Pr\{s_{m^*,t} = 1\}\geq q > \alpha$ when $Q_t > \gamma$. 
We also note that the queue arrival at step $t$ is $\alpha - s_{m^*,t} \in (0, 1)$.

Let $Z_t := \max\{0, Q_t - \gamma\}$. We have
\begin{align*}
    &\frac{1}{2}\Expectcond{Z^2_{t+1} - Z^2_t}{Z_t} \\
    & = \frac{1}{2} \Expectcond{\left(\max\{0, Z_t + \alpha - s_{m^*,t}\}\right)^2 - Z^2_t}{Z_t} \\
    & \leq \frac{1}{2} \Expectcond{\left(Z_t + \alpha - s_{m^*,t}\right)^2 - Z^2_t}{Z_t} \\
    & \leq \Expectcond{Z_t\left(\alpha - s_{m^*,t}\right) + \frac{1}{2}\left(\alpha - s_{m^*,t}\right)^2 }{Z_t} \\
    & \leq \Expectcond{Z_t\left(\alpha - s_{m^*,t}\right) }{Z_t} + \frac{1}{2} \\
    & \leq Z_t\left(\alpha - q\right) + \frac{1}{2}, 
\end{align*}
where the last inequality is because, as shown above, $\Pr\{s_{m^*,t} = 1\}\geq q$ when $Q_t > \gamma$ which is equivalent to $Z_t>0$.

Taking total expectation gives
\begin{align*}
    \Expectbracket{Z^2_{t+1} - Z^2_t}  \leq -2\Expectbracket{Z_t}\left(q-\alpha\right) + 1 \leq 1.
\end{align*}

Since $Z_1=0$, telescoping gives
\begin{align}
    \Expectbracket{Z^2_{t}}  \leq -2\sum_{\tau=1}^{t-1} \Expectbracket{Z_\tau}\left(q-\alpha\right) + (t-1), 
    \label{eq:proof_queue_length_0}
\end{align}
for $t>1$.

Then, noting that $q > \alpha$ and $Z_t\geq 0$, by Jensen's inequality, we have
\begin{align}
    \Expectbracket{Z_t} \leq \sqrt{\Expectbracket{Z^2_{t}} } \leq \sqrt{t-1} \leq \sqrt{t},
    \label{eq:proof_queue_length_1}
\end{align}
for $t\geq 1$.

In addition, from \eqref{eq:proof_queue_length_0}, we also have
\begin{align*}
    2\sum_{\tau=1}^{t-1} \Expectbracket{Z_\tau}\left(q-\alpha\right)  \leq (t-1) - \Expectbracket{Z^2_{t}} \leq t-1.
\end{align*}
Therefore, after replacing $t-1$ with $t$,
\begin{align}
    \frac{1}{t}\sum_{\tau=1}^{t} \Expectbracket{Z_\tau}  \leq \frac{1}{2(q-a)},
    \label{eq:proof_queue_bound_average}
\end{align}
for $t\geq 1$.

The final result then follows by combining $Q_t \leq \gamma + Z_t$ with \eqref{eq:proof_queue_length_1} and \eqref{eq:proof_queue_bound_average}, giving the two bounds.
\qed

\subsection{Proof of \texorpdfstring{\Cref{theorem:optimality_bound}}{Theorem \ref{theorem:optimality_bound}}}

We prove \Cref{theorem:optimality_bound} by first introducing a few lemmas.

\begin{lemma}
\label{lemma:sgd_bound}Choosing $\eta_k=\min\left\{\frac{2}{\mu (k+1)}, \frac{1}{L}\right\}$,
after $k\geq 1$ steps of predictor training using SGD with the loss function defined in \eqref{eq:cross-entropy}, we have
\begin{align}
    \Expectbracket{F(\mathbf{z}_k)}-F_\mathrm{min}\leq \mathcal{O}\left(\frac{1}{k}\right).
    \label{eq:proof_sgd_final}
\end{align}
\end{lemma}
\begin{proof}
Let $k$ denote the index of exploration steps. The SGD update of predictor training is
\begin{align}
\mathbf{z}_{k+1} \leftarrow \mathbf{z}_{k} - \eta_k\nabla F(\mathbf{z}_{k}, \mathbf{a}_k)
\end{align}
Let $\mathbf{g}_k:= \nabla F(\mathbf{z}_{k}, \mathbf{a}_k)$ for convenience.

By $L$-smoothness, we have
\begin{align}
\Expectcond{F(\mathbf{z}_{k+1})}{\mathbf{z}_k} & \leq F(\mathbf{z}_{k}) - \eta_k \innerprod{\nabla F(\mathbf{z}_{k}),\Expectcond{ \mathbf{g}_k}{\mathbf{z}_k} } + \frac{L\eta_k^2}{2}\Expectcond{\normsq{\mathbf{g}_k}}{\mathbf{z}_k}\nonumber \\
&\leq F(\mathbf{z}_{k}) - \eta_k \normsq{\nabla F(\mathbf{z}_{k})} + \frac{L\eta_k^2}{2}\left(\normsq{\nabla F(\mathbf{z}_{k})} + \sigma^2\right)\nonumber \\
&= F(\mathbf{z}_{k}) -\left( \eta_k - \frac{L\eta_k^2}{2} \right) \normsq{\nabla F(\mathbf{z}_{k})} + \frac{L\eta_k^2 \sigma^2}{2}. 
\label{eq:proof_sgd_1}
\end{align}

From PL condition, we have $\normsq{\nabla F(\mathbf{z})} \geq 2\mu\left(F(\mathbf{z})-F_\mathrm{min}\right), \forall \mathbf{z}$. When $\eta_k < \frac{2}{L}$, subtracting $F_\mathrm{min}$ on both sides of \eqref{eq:proof_sgd_1} and plugging in the PL inequality gives
\begin{align}
\Expectcond{F(\mathbf{z}_{k+1}) - F_\mathrm{min}}{\mathbf{z}_k} & \leq  (F(\mathbf{z}_{k}) - F_\mathrm{min}) -2\mu\left( \eta_k - \frac{L\eta_k^2}{2} \right) \left(F(\mathbf{z}_k)-F_\mathrm{min}\right) + \frac{L\eta_k^2 \sigma^2}{2} \nonumber \\
 & =  \left(1 -2\mu \eta_k \left(1  - \frac{L\eta_k}{2} \right)\right) \left(F(\mathbf{z}_k)-F_\mathrm{min}\right) + \frac{L\eta_k^2 \sigma^2}{2}. 
\label{eq:proof_sgd_2}
\end{align}

Let $\eta_k \leq \frac{1}{L}$. We have
\begin{align}
\Expectcond{F(\mathbf{z}_{k+1}) - F_\mathrm{min}}{\mathbf{z}_k} & \leq  
 \left(1 -\mu \eta_k \right) \left(F(\mathbf{z}_k)-F_\mathrm{min}\right) + \frac{L\eta_k^2 \sigma^2}{2}. 
\label{eq:proof_sgd_3}
\end{align}

Taking total expectation gives
\begin{align}
    \Expectbracket{F(\mathbf{z}_{k+1})} - F_\mathrm{min} & \leq  
 \left(1 -\mu \eta_k \right) \left(\Expectbracket{F(\mathbf{z}_k)}-F_\mathrm{min}\right) + \frac{L\eta_k^2 \sigma^2}{2}.     
\end{align}

Let $\mathcal{F}_k := \Expectbracket{F(\mathbf{z}_k)}-F_\mathrm{min}$. We have
\begin{align*}
    \mathcal{F}_2 &\leq (1-\mu\eta_1) \mathcal{F}_1 + \frac{L\eta_1^2 \sigma^2}{2}\\
    \mathcal{F}_3 &\leq (1-\mu\eta_1)(1-\mu\eta_2) \mathcal{F}_1 + \frac{L\sigma^2 ((1-\mu\eta_2)\eta_1^2 + \eta_2^2) }{2} \\
    \mathcal{F}_4 &\leq (1-\mu\eta_1)(1-\mu\eta_2)(1-\mu\eta_3) \mathcal{F}_1 + \frac{L\sigma^2 ((1-\mu\eta_3)(1-\mu\eta_2)\eta_1^2 + (1-\mu\eta_3)\eta_2^2 + \eta_3^2) }{2} \\
    \ldots
\end{align*}
Therefore,
\begin{align}
    \mathcal{F}_k &\leq \mathcal{F}_1 \prod_{\kappa=1}^{k-1}(1-\mu\eta_\kappa)  + \frac{L\sigma^2\sum_{\kappa=1}^{k-1}\eta_\kappa^2\prod_{\kappa'=\kappa+1}^{k-1}(1-\mu\eta_{\kappa'})}{2}.    \label{eq:proof_sgd_product_form}
\end{align}

Recall that $\eta_k=\min\left\{\frac{2}{\mu (k+1)}, \frac{1}{L}\right\}$. We note that, due to $L$-smoothness, we have $\mu\leq L$, because otherwise the $L$-smoothness contradicts with the PL condition. Therefore, $\frac{2L}{\mu} \geq 2$. We have $\eta_k=\frac{1}{L}$ when $k\leq \tilde{k} := \left\lfloor \frac{2L}{\mu}-1\right\rfloor$. For $k>\tilde{k}$, we have $\eta_k=\frac{2}{\mu (k+1)}$, and in this case, $1-\mu\eta_k = \frac{k-1}{k+1}$.

We first consider the second term of \eqref{eq:proof_sgd_product_form}.
For $k\leq\tilde{k}+1$,
\begin{align*}
    G(k) := \sum_{\kappa=1}^{k-1}\eta_\kappa^2\prod_{\kappa'=\kappa+1}^{k-1}(1-\mu\eta_{\kappa'}) = \sum_{\kappa=1}^{k-1}\frac{1}{L^2}\left(1-\frac{\mu}{L}\right)^{\kappa-1}=\frac{1}{L^2}\cdot\frac{1-\left(1-\frac{\mu}{L}\right)^{k-2}}{\frac{\mu}{L}}\leq \frac{1}{L\mu}.
\end{align*}
Let $k_0 :=\tilde{k}+1$. From the above, we have $G(k_0)\leq \frac{1}{L\mu}$. 
For $k>k_0$, we note that
\begin{align*}
    G(k+1)&=(1-\mu\eta_k)G(k) + \eta_k^2 \\
    &= \left(1-\frac{2}{k+1}\right)G(k) + \frac{4}{\mu^2 (k+1)^2}\\
    &= \frac{k-1}{k+1}\cdot G(k) + \frac{4}{\mu^2 (k+1)^2}.
\end{align*}
Thus,
\begin{align*}
    G(k_0+1)&\leq \frac{\tilde{k}}{L\mu (k_0+1)} + \frac{4}{\mu^2 (k_0+1)^2} \\
    G(k_0+2)&\leq \frac{\tilde{k}k_0}{L\mu (k_0+1)(k_0+2)} + \frac{4k_0}{\mu^2 (k_0+1)^2(k_0+2)}  + \frac{4}{\mu^2 (k_0+2)^2} \\
    G(k_0+3)&\leq \frac{\tilde{k}k_0}{L\mu (k_0+2)(k_0+3)} + \frac{4k_0}{\mu^2 (k_0+1)(k_0+2)(k_0+3)}  + \frac{4(k_0+1)}{\mu^2 (k_0+2)^2(k_0+3)} \\ &\quad + \frac{4}{\mu^2 (k_0+3)^2} \\
    G(k_0+4)&\leq \frac{\tilde{k}k_0}{L\mu (k_0+3)(k_0+4)} + \frac{4k_0}{\mu^2 (k_0+1)(k_0+3)(k_0+4)}  + \frac{4(k_0+1)}{\mu^2 (k_0+2)(k_0+3)(k_0+4)} \\ &\quad+ \frac{4(k_0+2)}{\mu^2 (k_0+3)^2(k_0+4)} + \frac{4}{\mu^2 (k_0+4)^2} \\
    \ldots
\end{align*}
For general $k>k_0$, after upper bounding some terms, we have
\begin{align}
    G(k)&\leq \frac{\tilde{k}k_0}{L\mu (k-1)k} + \frac{1}{\mu^2}  \sum_{\kappa=k_0}^{k-2}\frac{4\kappa}{(\kappa+1) (k-1)k} + \frac{4}{\mu^2 k^2} \nonumber \\
    &\leq \frac{\tilde{k}k_0}{L\mu (k-1)k} + \frac{1}{\mu^2}  \sum_{\kappa=k_0}^{k-2}\frac{4}{(k-1)k} +  \frac{4}{\mu^2 k^2} \nonumber\\
    &\leq \frac{\tilde{k}k_0}{L\mu (k-1)k} +  \frac{4}{\mu^2 k}\nonumber\\
    &=\mathcal{O}\left(\frac{1}{k}\right),
    \label{eq:proof_sgd_second_term}
\end{align}
where the last equality is because $\tilde{k}$ and $k_0$ are constants as they only depend on $L$ and $\mu$.

Next, consider the first term of \eqref{eq:proof_sgd_product_form}. When $k\leq k_0$,
\begin{align*}
    H(k):= \prod_{\kappa=1}^{k-1}(1-\mu\eta_\kappa) = \left(1-\frac{\mu}{L}\right)^{k-1}.
    \label{eq:proof_sgd_first_term}
\end{align*}

For $k>k_0$, we have
\begin{align*}
    H(k_0+1) &= H(k_0)\cdot\frac{\tilde{k}}{k_0+1}\\
    H(k_0+2) &= H(k_0)\cdot\frac{\tilde{k}k_0}{(k_0+1)(k_0+2)}\\
    H(k_0+3) &= H(k_0)\cdot\frac{\tilde{k}k_0}{(k_0+2)(k_0+3)}\\
    H(k_0+4) &= H(k_0)\cdot\frac{\tilde{k}k_0}{(k_0+3)(k_0+4)}\\
    \ldots
\end{align*}
For general $k\geq k_0$,
\begin{align}
    H(k) = \left(1-\frac{\mu}{L}\right)^{\tilde{k}} \cdot\frac{\tilde{k}k_0}{(k-1)k} =\mathcal{O}\left(\frac{1}{k^2}\right),
\end{align}
because $\tilde{k}$ and $k_0$ are constants as they only depend on $L$ and $\mu$.

Combining \eqref{eq:proof_sgd_second_term} and \eqref{eq:proof_sgd_first_term} with \eqref{eq:proof_sgd_product_form}, we obtain
\begin{align}
    \mathcal{F}_k = \Expectbracket{F(\mathbf{z}_k)}-F_\mathrm{min}\leq \mathcal{O}\left(\frac{1}{k}\right).
\end{align}
\end{proof}

\begin{lemma}
\label{lemma:absolute_difference_bound}
Choosing $\eta_k=\min\left\{\frac{2}{\mu (k+1)}, \frac{1}{L}\right\}$, for any $t\geq 1$, assume that after $t$ requests, $k$ steps of predictor training has occurred. We have
\begin{align}
    \Expectbracket{\max_m|\hat{s}_{m,t} - s_{m,t}|} \leq  M F_\mathrm{min} + \mathcal{O}\left(\frac{M}{k}\right).
\end{align}
\end{lemma}

\begin{proof}
For the ease of discussion, let $\epsilon$ denote the upper bound of $\Expectbracket{F(\mathbf{z}_k)}-F_\mathrm{min}$ so that $\epsilon=\mathcal{O}\left(\frac{1}{k}\right)$ according to \Cref{lemma:sgd_bound}.

When $\Expectbracket{F(\mathbf{z}_k)}-F_\mathrm{min}\leq \epsilon$, from \eqref{eq:cross-entropy} and noting that the cross entropy is non-negative, we have
\begin{align}
    &\Expectbracket{\max_m \left\{ -s_{m,t}\log  \hat{s}_{m,t} - (1- s_{m,t})  \log (1- \hat{s}_{m,t} )\right\} }\nonumber \\
    &\leq \Expectbracket{\sum_{m=1}^M \left( -s_{m,t}\log  \hat{s}_{m,t} - (1- s_{m,t})  \log (1- \hat{s}_{m,t} ) \right) }\nonumber \\
    &\leq M \Expectbracket{F(\mathbf{z}_k)}\nonumber \\
    &\leq M F_\mathrm{min} + \epsilon M. \label{proof_lemma:absolute_difference_bound_1}
\end{align}

For arbitrary sample $\mathbf{a}_t$ and model $m$, let
\begin{align*}
    \Gamma := -  s_{m,t}\log \hat{s}_{m,t}  - (1- s_{m,t})  \log (1-\hat{s}_{m,t}).
\end{align*}
We consider two cases as follows.

When $s_{m,t}=0$, we have
\begin{align*}
    &\Gamma = -\log (1-\hat{s}_{m,t}) \\
    &\Rightarrow \hat{s}_{m,t} = 1-e^{-\Gamma}\\
    &\Rightarrow |\hat{s}_{m,t} - s_{m,t}| = 1-e^{-\Gamma}.
\end{align*}
When $s_{m,t}=1$, we have
\begin{align*}
    &\Gamma = -\log \hat{s}_{m,t} \\
    &\Rightarrow 1-\hat{s}_{m,t} = 1-e^{-\Gamma}\\
    &\Rightarrow |\hat{s}_{m,t} - s_{m,t}| = 1-e^{-\Gamma}.
\end{align*}

Noting the elementary inequality $e^{-\Gamma}\geq 1-\Gamma$, we obtain
\begin{align*}
    |\hat{s}_{m,t} - s_{m,t}| \leq \Gamma.
\end{align*}

Because this relation holds for any sample and the corresponding $\Gamma$ defined on the sample, the expectation of the cross-entropy loss cannot be smaller than the expectation of the absolute difference. Combining with \eqref{proof_lemma:absolute_difference_bound_1}, we have
\begin{align}
    \Expectbracket{\max_m|\hat{s}_{m,t} - s_{m,t}|} \leq M F_\mathrm{min} + \epsilon M = M F_\mathrm{min} + \mathcal{O}\left(\frac{M}{k}\right).
\end{align}
\end{proof}

\begin{lemma}
\label{lemma:k_bound}
Let $k$ denote the random variable of the number of predictor training steps after processing $t$ requests. We have
\begin{align}
    \Expectbracket{k} \leq \mathcal{O}\left(t^{\frac{3}{4}}\right) \textnormal{ and }
    \Expectbracket{\frac{1}{k}} \leq \mathcal{O}\left(\frac{1}{t^{\frac{3}{4}}}\right).
\end{align}    
\end{lemma}

\begin{proof}
Recall that $X_t \sim \mathrm{Bernoulli}(p_t)$ is an indicator denoting whether an SGD step for predictor training occurs when processing request $t$. 
In the following, we assume that $k$ is the total number of SGD steps after processing $t$ requests. 
We have
\begin{align}
    \lambda := \Expectbracket{k} = \Expectbracket{\sum_{\tau=1}^t X_\tau} = \sum_{\tau=1}^t p_\tau = \sum_{\tau=1}^t \min\left(1, \frac{c}{\sqrt[4]{\tau}}\right) = \Theta\left(t^{\frac{3}{4}}\right).
    \label{eq:proof_k_expectation}
\end{align}
This proves the first result.

Considering $\frac{1}{k}$, we note that
\begin{align}
    \Expectbracket{\frac{1}{k}} = \Expectbracket{\frac{1}{k}\cdot\Identity_{k\geq\lambda/2}} + \Expectbracket{\frac{1}{k}\cdot\Identity_{k<\lambda/2}} \leq \frac{2}{\lambda} + \Expectbracket{\frac{1}{k}\cdot\Identity_{k<\lambda/2}},
    \label{eq:proof_k_decompose}
\end{align}
where $\Identity_\mathcal{C}$ is an indicator function of whether the condition $\mathcal{C}$ holds.

We now consider the last term in \eqref{eq:proof_k_decompose}.
The multiplicative Chernoff bound shows that
\begin{align*}
    \Pr\{k\leq(1-\delta)\lambda\}\leq e^{-\frac{\delta^2\lambda}{2}},
\end{align*}
for $0<\delta<1$. Choosing $\delta=\frac{1}{2}$ gives
\begin{align*}
    \Pr\left\{k\leq\frac{\lambda}{2}\right\}\leq e^{-\frac{\lambda}{8}}.
\end{align*}
Because $k\geq 1$,
\begin{align}
    \Expectbracket{\frac{1}{k}\cdot\Identity_{k<\lambda/2}} \leq \Expectbracket{\Identity_{k<\lambda/2}} \leq e^{-\frac{\lambda}{8}} = \frac{1}{e^{\frac{\lambda}{8}}} \leq \frac{1}{1+\frac{\lambda}{8}},
    \label{eq:proof_k_decompose_second_term}
\end{align}
where the last inequality is due to the elementary relation that $e^x\geq 1+x$ for any $x$.

Combining \eqref{eq:proof_k_expectation}, \eqref{eq:proof_k_decompose}, and \eqref{eq:proof_k_decompose_second_term}, we obtain
\begin{align}
    \Expectbracket{\frac{1}{k}} \leq \mathcal{O}\left(\frac{1}{t^{\frac{3}{4}}}\right).
\end{align}
\end{proof}

Based on these lemmas, we are now ready to prove \Cref{theorem:optimality_bound}.

\begin{proof}[Proof of \Cref{theorem:optimality_bound}]
We first consider any $t$ such that $X_t = 0$, i.e., no exploration or predictor training.
For the ease of presentation, let $m^*$ denote the optimal solution to \eqref{eq:per-request-optimization} for some given $t$, i.e., $y_{m^*,t}=1$ and $y_{m',t}=0$ for $m'\neq m^*$, where $t$ is inferred from the context. 
We consider the following Lyapunov drift of the queue length: 
\begin{align}
    \frac{1}{2}\Expectcond{Q^2_{t+1} - Q^2_t}{Q_t} 
    & = \frac{1}{2} \Expectcond{\left(\max\{0, Q_t + \alpha - s_{m^*,t}\}\right)^2 - Q^2_t}{Q_t} \nonumber\\
    & \leq \frac{1}{2} \Expectcond{\left(Q_t + \alpha - s_{m^*,t}\right)^2 - Q^2_t}{Q_t} \nonumber\\
    & \leq \frac{1}{2} \Expectcond{2Q_t (\alpha - s_{m^*,t}) + 1}{Q_t} \nonumber\\    
    & =  \Expectcond{Q_t (\alpha - s_{m^*,t} + \hat{s}_{m^*,t} - \hat{s}_{m^*,t})}{Q_t} + \frac{1}{2} \nonumber\\  
    & = \Expectcond{Q_t (\alpha - \hat{s}_{m^*,t})}{Q_t} +  \Expectcond{Q_t (\hat{s}_{m^*,t} - s_{m^*,t} )}{Q_t} + \frac{1}{2} \nonumber\\  
    & \leq \Expectcond{Q_t (\alpha - \hat{s}_{m^*,t})}{Q_t} +  Q_t \Expectbracket{\max_m|\hat{s}_{m,t} - s_{m,t}|} + \frac{1}{2},
\label{eq:proof_optimality_drift}
\end{align}
where the second inequality follows from expanding the square and $(\alpha - s_{m^*,t})^2 \leq 1$.

Let $\{y_{m,t}^{\textnormal{OPT}},\forall m\}$ denote the result of an optimal stationary policy to \eqref{eq:original_problem}.
For our online decision-making problem \eqref{eq:per-request-optimization}, we get
\begin{align}
    & \Expectcond{V E_{m^*,t} + Q_t\left(\alpha - \hat{s}_{m^*,t}\right)}{Q_t} \nonumber\\
    & \leq \Expectcond{V\cdot \sum_{m=1}^M y_{m,t}^{\textnormal{OPT}} E_{m,t} + Q_t \sum_{m=1}^M y_{m,t}^{\textnormal{OPT}}\left(\alpha - \hat{s}_{m,t} \right)}{Q_t} \nonumber\\
    & = \Expectcond{V\cdot \sum_{m=1}^M y_{m,t}^{\textnormal{OPT}} E_{m,t} + Q_t \left(\sum_{m=1}^M y_{m,t}^{\textnormal{OPT}}(\alpha - \hat{s}_{m,t} + s_{m,t} - s_{m,t} )\right)}{Q_t} \nonumber\\
    & = V \Expectbracket{\sum_{m=1}^M y_{m,t}^{\textnormal{OPT}} E_{m,t}} + Q_t \Expectbracket{\sum_{m=1}^M y_{m,t}^{\textnormal{OPT}}(s_{m,t} - \hat{s}_{m,t} ) } + Q_t \Expectbracket{\sum_{m=1}^M y_{m,t}^{\textnormal{OPT}}(\alpha - s_{m,t} ) } \nonumber\\
    & \leq V E^\textrm{OPT} + Q_t \Expectbracket{ \max_m\{\,\abs{\hat{s}_{m,t} - s_{m,t}}\}},
    \label{eq:proof_drift_plus_penalty}
\end{align}
where the last inequality is because the optimal stationary policy satisfies the constraint \eqref{eq:accuracy_constraint} and the cost is minimized when the constraint holds with equality, thus $\Expectbracket{\sum_{m=1}^M y_{m,t}^{\textnormal{OPT}}(\alpha - s_{m,t} ) }=0$.

Combining \eqref{eq:proof_optimality_drift} and \eqref{eq:proof_drift_plus_penalty}, we obtain
\begin{align*}  
    & \frac{1}{2}\Expectcond{Q^2_{t+1} - Q^2_t}{Q_t} + V\Expectcond{ \sum_{m=1}^M y_{m,t} E_{m,t} }{Q_t} \\
    &\leq \Expectcond{Q_t (\alpha - \hat{s}_{m^*,t})}{Q_t} +  Q_t \Expectbracket{\max_m|\hat{s}_{m,t} - s_{m,t}|} + \frac{1}{2} + V\mathbb{E}\left[ \sum_{m=1}^M y_{m,t} E_{m,t} \Big| Q_t \right] \\
    &= \Expectcond{V E_{m^*,t} + Q_t\left(\alpha - \hat{s}_{m^*,t}\right)}{Q_t} +  Q_t \Expectbracket{\max_m|\hat{s}_{m,t} - s_{m,t}|} + \frac{1}{2} \\
    &\leq V E^\textrm{OPT} + Q_t \Expectbracket{ \max_m\{\,\abs{\hat{s}_{m,t} - s_{m,t}}\}} +  Q_t \Expectbracket{\max_m|\hat{s}_{m,t} - s_{m,t}|} + \frac{1}{2} \\
    &= V E^\textrm{OPT} + 2Q_t \Expectbracket{ \max_m\{\,\abs{\hat{s}_{m,t} - s_{m,t}}\}}  + \frac{1}{2}. 
\end{align*}

Taking total expectation, we obtain
\begin{align*}  
    & \frac{1}{2}\Expectbracket{Q^2_{t+1} - Q^2_t} + V\Expectbracket{ \sum_{m=1}^M y_{m,t} E_{m,t} } \\
    &\leq V E^\textrm{OPT} + 2\Expectbracket{Q_t}\cdot \Expectbracket{ \max_m\{\,\abs{\hat{s}_{m,t} - s_{m,t}}\}}  + \frac{1}{2}. 
\end{align*}

Therefore,
\begin{align*}  
    &\Expectbracket{ \sum_{m=1}^M y_{m,t} E_{m,t} } \\
    &\leq E^\textrm{OPT} + \frac{2}{V}\Expectbracket{Q_t}\cdot \Expectbracket{ \max_m\{\,\abs{\hat{s}_{m,t} - s_{m,t}}\}}  + \frac{1}{2V} +\frac{1}{2V}\Expectbracket{Q^2_t - Q^2_{t+1}}\\
    &\leq E^\textrm{OPT} + \frac{2}{V}\Expectbracket{Q_t}\cdot\left(\Expectbracket{\mathcal{O}\left(\frac{M}{k}\right)} + M F_\mathrm{min}\right)  + \frac{1}{2V} +\frac{1}{2V}\Expectbracket{Q^2_t - Q^2_{t+1}}\\
    &\leq E^\textrm{OPT} + \frac{2}{V}\left((\gamma+\sqrt{t})\mathcal{O}\left(\frac{M}{t^{\frac{3}{4}}}\right) + M F_\mathrm{min}\Expectbracket{Q_t}\right)  + \frac{1}{2V} +\frac{1}{2V}\Expectbracket{Q^2_t - Q^2_{t+1}},
\end{align*}
where the second inequality uses \Cref{lemma:absolute_difference_bound} and the last inequality uses \Cref{theorem:max_queue_length} and $\gamma := \max\left\{\alpha \psi; \frac{V\Delta_E}{\beta}\right\}$.

For any $t$ with $X_t=1$, we note that the maximum cost due to exploration when processing such requests is $M E_{\mathrm{max}}$.

We now include all $t$ where either $X_t=0$ or $X_t=1$. 
Then,
\begin{align*}
    &\Expectbracket{\frac{1}{T}\sum_{t=1}^T \sum_{m=1}^M y_{m,t} E_{m,t} }   \\  
    &\leq E^\textrm{OPT} + \frac{2}{V}\left(\frac{1}{T}\sum_{t=1}^T(\gamma+\sqrt{t})\mathcal{O}\left(\frac{M}{t^{\frac{3}{4}}}\right) + \frac{M F_\mathrm{min}}{T}\sum_{t=1}^T\Expectbracket{Q_t}\right)  + \frac{1}{2V} +\frac{1}{2VT}\Expectbracket{Q^2_1 - Q^2_{t+1}} \\&\quad + \frac{\Expectbracket{K} ME_\mathrm{max}}{T}\\
    &\leq E^\textrm{OPT} +\mathcal{O}\left(\frac{1}{T}\sum_{t=1}^T\frac{M}{t^{\frac{3}{4}-\frac{1}{2}}} + M F_\mathrm{min}  + \frac{1}{V}+ \frac{M}{T^\frac{1}{4}}\right) \\
    &= E^\textrm{OPT} +\mathcal{O}\left(\frac{M}{\sqrt[4]{T}} + M F_{\mathrm{min}}  +  \frac{1}{V}\right),  
\end{align*}
where the second inequality uses \Cref{theorem:max_queue_length} and \Cref{lemma:k_bound}.
\end{proof} 

\clearpage

\section{Experimental Details}
\label{app:experimental_design}

In this section we provide full details for our experimental design. 

\textbf{Datasets \& Benchmarks.}
We use the LM-Eval Harness as a basis for all of our evaluations. 
We use a zero-shot setting for each benchmark. 
Our evaluations are based on the default configurations provided by LM Eval.
We use vLLM v0.8.4 as the inference backend for LM Eval.

\textbf{LLM Zoo.}
Our zoo coniststs of three Llama models, namely Llama 3.2 1B, Llama 3.1 8B, and Llama 3.3 70B. 
We use publicly available model checkpoints available on the HuggingFace Hub, specifically, \texttt{meta-llama/Llama-3.2-1B-Instruct}, \texttt{unsloth/Meta-Llama-3.1-8B-Instruct-bnb-4bit}, \newline and \texttt{unsloth/Llama-3.3-70B-Instruct-bnb-4bit}.

\textbf{Request Satisfaction Predictor.}
We design a predictor model based on the ModernBert transformer and implement a classification head on top of it. 
Before we pass the final-BERT-layer outputs into the classification layer, we pool the outputs and use a dropout (= 0.1) for better training effectiveness. 
We first have a linear layer, followed by a layer norm operation, followed by a ReLU activation, another dropout (= 0.1), and then a final linear layer. 
We use a sigmoid function to compute the classifier logits. 
For each benchmark in our main paper, we ran a hyperparameter sweep to identify the most effective hyperparameter combinations. 
They are listed in \Cref{tab:hp_configuration}.

\begin{table}[H]
    \caption{Predictor model hyperparameter configurations across benchmarks.}
    \label{tab:hp_configuration}
    \centering
    \resizebox{0.8\textwidth}{!}{
        \begin{tabular}{lrrrrr}
        \toprule
        Benchmark     & Learning Rate & Weight Decay & Momentum & Max. Seq. Len. & Dropout \\
        \midrule
        ARC Challenge & 0.0606        & 0.01         & 0.90     & 256            & 0.1     \\
        ARC Easy      & 0.0826        & 0.01         & 0.90     & 256            & 0.1     \\
        BoolQ         & 0.0767        & 0.01         & 0.95     & 128            & 0.1     \\
        LogiQA        & 0.0272        & 0.01         & 0.90     & 256            & 0.1     \\
        PiQA          & 0.0367        & 0.01         & 0.90     & 64             & 0.1     \\
        SciQ          & 0.0596        & 0.01         & 0.95     & 64             & 0.1     \\
        SocialIQA     & 0.0542        & 0.01         & 0.95     & 64             & 0.1     \\
        Winogrande    & 0.0660        & 0.01         & 0.90     & 64             & 0.1   \\
        \bottomrule
        \end{tabular}
    }
\end{table}

\textbf{Random Baseline with Constraint Satisfaction.}
We implement a random baseline that follows constraint satisfaction, or in other words, SLA compliance. 
We use the following implementation to facilitate the baseline. 

\begin{lstlisting}[language=Python]
def calculate_probabilities(model_accuracies: list, alpha: float):
    """
    Function to compute a priori probabilities for a baseline 
    model selection process that provides SLA compliance over time. 
    """
    accuracies = np.array(model_accuracies)
    n = len(accuracies)

    # Check if possible
    if alpha > max(accuracies):
        raise ValueError("Alpha too high")

    p = np.ones(n) / n

    for _ in range(5000):
        current_acc = np.dot(p, accuracies)

        if current_acc >= alpha - 1e-6:
            return p

        # Simple update
        for i in range(n):
            if accuracies[i] > current_acc:
                p[i] *= 1.01  # Increase good models
            else:
                p[i] *= 0.99  # Decrease bad models

        # Normalize
        p = p / np.sum(p)

   
    idx_sorted = np.argsort(accuracies)[::-1]

    # Calculate minimum probability for best model
    best_acc = accuracies[idx_sorted[0]]
    worst_acc = accuracies[idx_sorted[-1]]

    # Start with minimum probabilities for all
    min_prob = 1e-10
    p = np.full(n, min_prob)
    remaining = 1.0 - n * min_prob

    # Distribute remaining probability
    for i in range(n):
        idx = idx_sorted[i]

        if i == n - 1:
            p[idx] += remaining
        else:
            # Give more to better models
            weight = (accuracies[idx] - worst_acc) / (best_acc - worst_acc)
            allocation = remaining * weight * 0.8
            p[idx] += allocation
            remaining -= allocation

    p = p / np.sum(p)

    return p
\end{lstlisting}

\textbf{RouteLLM.}
To reproduce the RouteLLM results, we integrated the RouteLLM controller into our existing evaluation pipeline. RouteLLM supports only two models: a weak and a strong model. We configured these as the Llama 3.2 1B (weak) and 70B (strong) models, respectively. 
We chose to use the BERT-based router of RouteLLM.
We swept the routing threshold from 0.1 to 0.9 in increments of 0.1. As shown in Table \ref{tab:alpha_threshold_mapping}, RouteLLM requires careful tuning of the routing threshold to achieve desired performance, which lacks a direct mapping to user-specified service level requirements.

\begin{table}[ht]
\centering
\caption{Mapping from \algname $\alpha$ to RouteLLM decision threshold.}
\label{tab:alpha_threshold_mapping}
\resizebox{0.5\textwidth}{!}{
    \begin{tabular}{lcccccc}
\toprule
Dataset        & $\alpha_1$ & Thresh$_1$ & $\alpha_2$ & Thresh$_2$ & $\alpha_3$ & Thresh$_3$ \\
\midrule
ARC Challenge & 0.7        & 0.5        & 0.6        & 0.6        & 0.7        & 0.4        \\
ARC Easy      & 0.6        & 0.8        & 0.7        & 0.65       & 0.6        & 0.75       \\
BoolQ        & 0.5        & 0.8        & 0.5        & 0.85       & 0.6        & 0.7        \\
LogiQA         & 0.7        & 0.3        & 0.5        & 0.45       & 0.5        & 0.4        \\
PiQA           & 0.6        & 0.75       & 0.5        & 0.81       & 0.6        & 0.78       \\
SciQ           & 0.5        & 0.97       & 0.6        & 0.95       & 0.5        & 0.96       \\
SocialIQA    & 0.7        & 0.44       & 0.7        & 0.46       & 0.7        & 0.42       \\
Winogrande     & 0.5        & 0.75       & 0.6        & 0.65       & 0.5        & 0.7        \\
\bottomrule
\end{tabular}

}
\end{table}

\textbf{RouterDC.}
To reproduce RouterDC, we trained its routing module on our benchmark tasks. The router encodes each query using the pretrained encoder \texttt{microsoft/deberta-v3-base} and compares its representation to a set of trainable expert embeddings, one for each model in the ensemble. Cosine similarity is used to produce a logit vector over the experts, and the router is optimized to prefer more accurate models via contrastive losses. In our setup, the candidate experts were Llama 3 Instruct models with 1B, 8B, and 70B parameters. Training was performed for 1,000 steps using the AdamW optimizer, with a batch size of 64 and a learning rate of $5 \times 10^{-5}$. The training set consisted of 100 queries from each of ten benchmarks, totaling 1,000 samples. For evaluation, we load the trained checkpoint and integrate the router into our existing evaluation pipeline.

\textbf{Code.}
Our code base is made fully public on GitHub. 
It can be found here: \textit{https://github.com/laminair/mess-plus}

\textbf{Evaluation Setup.}
While our experiments can be run on a single GPU (with ~80GB VRAM), we conducted our experiments using 2 H100 GPUs.
We distribute the LLMs and the predictor as follows: 
The small and medium sized LLMs along with the predictor are located on 1 GPU and the large model is placed on the other GPU.
We repeat each experiment with three different random seeds ($[42, 43, 44]$).
Since we query the LLMs sequentially, we can capture their individual energy consumption. 
When doing parallel calls, it is necessary to place each LLM on a separate GPU and configure the Zeus monitor to properly return the energy statistics for each model. 

\section{Additional Evaluations}
\label{app:addl_evaluations}

\subsection{Additional Results Related to Our Main Findings}
Our experimental results demonstrate that the parameter V, which controls the priority given to cost efficiency in the \algname routing algorithm, exhibits significant influence on both energy consumption and performance metrics across multiple benchmarks (\Cref{tab:results_part12,tab:results_part13}). 
In the main results with the standard V value, \algname achieves remarkable energy efficiency with an average operating cost of 1.08 MJ while maintaining satisfactory performance (68.44\% request satisfaction) across all benchmarks. 
When reducing V to 0.0001, thereby decreasing the emphasis on energy efficiency, we observe a 65.7\% increase in operating costs to 1.79 MJ with only a marginal improvement in performance to 69.16\%. 
Further reducing V to 0.00001 yields an additional cost increase to 1.88 MJ (74.1\% higher than the standard configuration) while performance improves only slightly to 69.41\%. 
These diminishing returns highlight the effectiveness of our approach in balancing the performance-efficiency trade-off. 
Notably, the distribution of model calls shifts substantially as V decreases—the utilization of the 70B model increases from 34\% with standard V to 54\% with V=0.00001, while mid-sized 8B model usage decreases from 40\% to 23\%, indicating a clear preference for higher-capacity models when efficiency constraints are relaxed. 
Individual benchmarks exhibit varying sensitivities to the V parameter; LogiQA shows the most substantial performance gain (41.02\% to 43.89\%) with decreased V values, while SciQ maintains relatively stable performance ($\approx 96\%$) despite significant variations in model call distribution. 
The ARC Easy benchmark demonstrates one of the most dramatic cost increases, from 1.74 MJ to 5.39 MJ at the lowest V value, emphasizing how routing decisions can substantially impact energy consumption for specific task types.
Even with reduced emphasis on efficiency, \algname maintains competitive or superior performance compared to alternative methods like RouteLLM and RouterDC while consuming less energy on average. 
These findings underscore the flexibility of our approach in accommodating different deployment scenarios where either performance or energy efficiency might be prioritized, while consistently outperforming baseline single-model approaches for the same levels of request satisfaction.

\subsection{The relationship between \texorpdfstring{$\alpha$}{alpha} and $V$}
Our experimental evaluation across multiple reasoning benchmarks (ARC Challenge, ARC Easy, BoolQ, LogiQA, PiQA, SciQ, and SocialIQA) exhibits the exact relationship between $\alpha$ and $V$ that we show in our theoretical analysis. 
The results demonstrate that lower $V$ values ($V=0.0001$) consistently achieve higher request satisfaction rates while incurring greater computational costs, exhibiting a slower convergence to stability but ultimately reaching higher performance plateaus that can satisfy more demanding Service Level Agreement (SLA) thresholds. 
Conversely, higher $V$ values ($V=0.01$) prioritize cost efficiency, resulting in significantly lower average costs, faster initial convergence, but ultimately lower satisfaction plateaus that are very close to $\alpha$. Medium V values ($V=0.001$) strike a compelling balance, offering reasonable satisfaction rates with moderate computational investment. 
Notably, our Winogrande analysis illuminates the explicit relationship between SLA satisfaction timing and both $\alpha$ thresholds and $V$ values, with higher $\alpha$ requirements (0.65, 0.7, and 0.75) correspondingly satisfied at later request points (steps 740, 803, and 994, respectively). When compared against baseline methods, our approach approaches provides similar satisfaction levels like RouterDC while maintaining substantially lower computational costs, demonstrating superior efficiency in the satisfaction-cost frontier. 
These findings underpin that $V$ provides an intuitive and flexible mechanism for system operators to deliberately navigate performance-cost trade-offs according to application-specific requirements, enabling precise calibration between resource efficiency and quality of service in large-scale LLM deployment environments.

\subsection{Predictor Training Evaluation}
\begin{figure}
    \centering
    \includegraphics[width=\linewidth, angle=0]{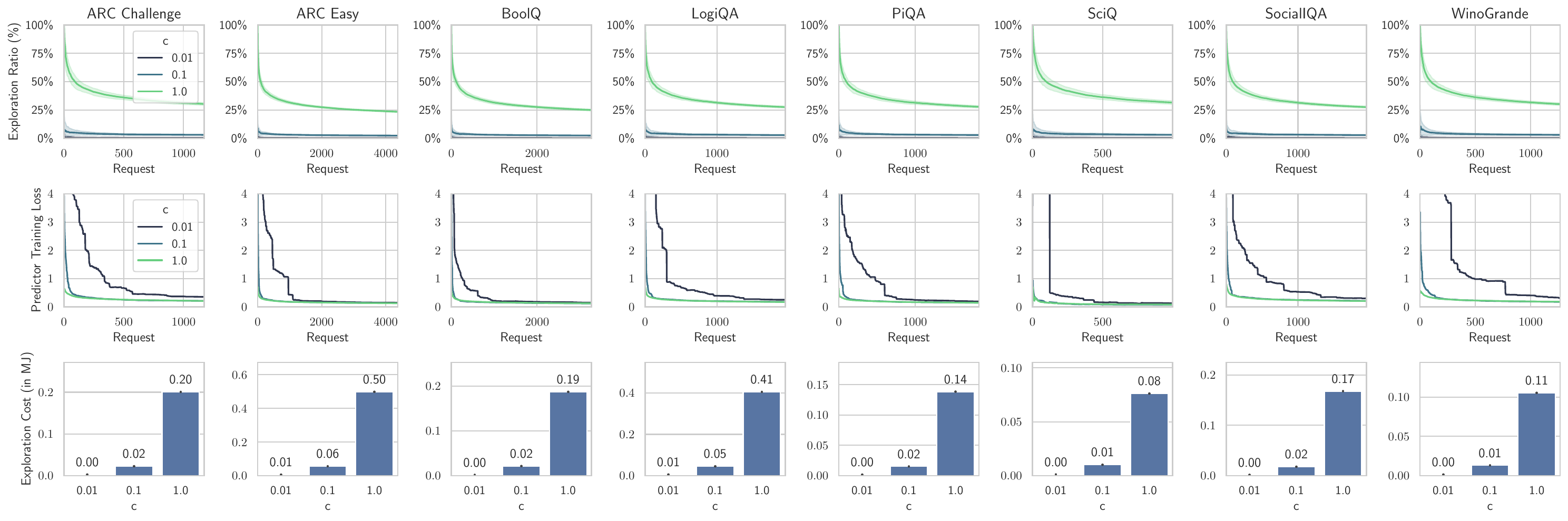}
    \caption{Full overview of predictor training cost across all benchmarks used in our paper.}
    \label{fig:addl_c_results_complete_appendix}
\end{figure}

Our analysis of exploration-exploitation dynamics across eight reasoning benchmarks reveals critical insights for efficient predictive modeling (\Cref{fig:addl_c_results_complete_appendix}). 
We observe that the exploration parameter (c) exhibits predictable effects across benchmarks, with higher values (c = 1.0) maintaining robust exploration but at approximately ten-fold increased energy costs compared to conservative settings (c = 0.01). 
Notably, task complexity correlates with resource requirements, as evidenced by the significantly higher exploration costs. 
The predictor training loss patterns indicate that higher exploration parameters facilitate faster convergence and lower overall loss values, suggesting more robust optimization, though with diminishing returns relative to energy expenditure. 
These findings highlight the importance of context-aware parameter selection in balancing performance gains against computational costs, particularly relevant as AI systems scale and energy efficiency becomes increasingly critical. 
Generally, we find that choosing $c=0.1$ provides a strong basis for \algname across benchmarks.

\begin{table}[!ht]
\caption{Additional results for our main results with a smaller value for $V = 0.0001$, which reduces the priority for cost efficiency.}
\label{tab:results_part12}
\resizebox{\textwidth}{!}{
    \centering
    \begin{tabular}{lrrrrrrrrr}
    \cmidrule(lr){1-1}\cmidrule(lr){2-4}\cmidrule(lr){5-7}\cmidrule(lr){8-10}
    Benchmark & \multicolumn{3}{c}{ARC Challenge ($\alpha = 50\%$)} & \multicolumn{3}{c}{ARC Easy ($\alpha = 75\%$)} & \multicolumn{3}{c}{BoolQ  ($\alpha = 80\%$)} \\
    Method & \thead{Operating \\ Cost (in MJ)} & \thead{Request \\ Satisfaction (in \%)} & \thead{Model Call Ratio (in \%) \\ (L70B/L8B/L1B)} & \thead{Operating \\ Cost (in MJ)} & \thead{Request \\ Satisfaction (in \%)} & \thead{Model Call Ratio (in \%) \\ (L70B/L8B/L1B)} & \thead{Operating \\ Cost (in MJ)} & \thead{Request \\ Satisfaction (in \%)} & \thead{Model Call Ratio (in \%) \\ (L70B/L8B/L1B)} \\
    \cmidrule(lr){1-1}\cmidrule(lr){2-4}\cmidrule(lr){5-7}\cmidrule(lr){8-10}
    Llama 1B & 0.09$\scriptscriptstyle\pm0.00$ & \textcolor{red}{37.88$\scriptscriptstyle\pm5.39$} & 0\% / 0\% / 100\% & 0.20$\scriptscriptstyle\pm0.00$ & \textcolor{red}{62.76$\scriptscriptstyle\pm5.37$} & 0\% / 0\% / 100\% & 0.14$\scriptscriptstyle\pm0.00$ & \textcolor{red}{69.17$\scriptscriptstyle\pm5.13$} & 0\% / 0\% / 100\% \\
    Llama 8B & \underline{0.46$\scriptscriptstyle\pm0.00$} & \textcolor{darkgreen}{54.44$\scriptscriptstyle\pm5.54$} & 0\% / 100\% / 0\% & \underline{0.97$\scriptscriptstyle\pm0.00$} & \textcolor{darkgreen}{79.72$\scriptscriptstyle\pm4.47$} & 0\% / 100\% / 0\% & \underline{0.43$\scriptscriptstyle\pm0.00$} & \textcolor{darkgreen}{84.16$\scriptscriptstyle\pm4.06$} & 0\% / 100\% / 0\% \\
    Llama 70B & 2.35$\scriptscriptstyle\pm0.01$ & \textcolor{darkgreen}{60.84$\scriptscriptstyle\pm5.43$} & 100\% / 0\% / 0\% & 5.05$\scriptscriptstyle\pm0.01$ & \textcolor{darkgreen}{83.12$\scriptscriptstyle\pm4.16$} & 100\% / 0\% / 0\% & 3.40$\scriptscriptstyle\pm0.00$ & \textcolor{darkgreen}{88.78$\scriptscriptstyle\pm3.51$} & 100\% / 0\% / 0\% \\
    Educated Guessing & \textbf{1.00$\scriptscriptstyle\pm0.09$} & \textcolor{darkgreen}{51.65$\scriptscriptstyle\pm2.98$} & 35\% / 31\% / 34\% & \textbf{2.00$\scriptscriptstyle\pm0.08$} & \textcolor{red}{74.00$\scriptscriptstyle\pm4.39$} & 31\% / 32\% / 36\% & \textbf{1.31$\scriptscriptstyle\pm0.04$} & \textcolor{darkgreen}{80.47$\scriptscriptstyle\pm1.08$} & 33\% / 34\% / 33\% \\
    RouteLLM & 1.24$\scriptscriptstyle\pm0.10$ & \textcolor{darkgreen}{51.17$\scriptscriptstyle\pm2.93$} & 50\% / 0\% / 50\% & 4.05$\scriptscriptstyle\pm0.01$ & \textcolor{darkgreen}{82.54$\scriptscriptstyle\pm2.12$} & 100\% / 0\% / 0\% & 2.96$\scriptscriptstyle\pm0.04$ & \textcolor{darkgreen}{86.83$\scriptscriptstyle\pm1.27$} & 87\% / 0\% / 13\% \\
    RouterDC & 2.09$\scriptscriptstyle\pm0.06$ & \textcolor{darkgreen}{60.94$\scriptscriptstyle\pm2.92$} & 88\% / 12\% / 0\% & 3.61$\scriptscriptstyle\pm0.06$ & \textcolor{darkgreen}{82.30$\scriptscriptstyle\pm2.60$} & 85\% / 15\% / 0\% & 2.14$\scriptscriptstyle\pm0.05$ & \textcolor{darkgreen}{87.06$\scriptscriptstyle\pm2.70$} & 58\% / 42\% / 0\% \\
    \textbf{MESS+ (ours)} & 1.51$\scriptscriptstyle\pm0.09$ & \textcolor{darkgreen}{54.58$\scriptscriptstyle\pm3.15$} & 70\% / 9\% / 21\% & 4.87$\scriptscriptstyle\pm0.09$ & \textcolor{darkgreen}{78.20$\scriptscriptstyle\pm1.70$} & 54\% / 28\% / 19\% & 1.38$\scriptscriptstyle\pm0.05$ & \textcolor{darkgreen}{81.12$\scriptscriptstyle\pm4.99$} & 38\% / 30\% / 32\% \\
    \cmidrule(lr){1-1}\cmidrule(lr){2-4}\cmidrule(lr){5-7}\cmidrule(lr){8-10}
    \end{tabular}
    }

    \resizebox{\textwidth}{!}{
    \centering
    \begin{tabular}{lrrrrrrrrr}
    \cmidrule(lr){1-1}\cmidrule(lr){2-4}\cmidrule(lr){5-7}\cmidrule(lr){8-10}
    Benchmark & \multicolumn{3}{c}{LogiQA ($\alpha = 40\%$)} & \multicolumn{3}{c}{PiQA ($\alpha = 78\%$)} & \multicolumn{3}{c}{SciQ ($\alpha = 96\%$)} \\
    Method & \thead{Operating \\ Cost (in MJ)} & \thead{Request \\ Satisfaction (in \%)} & \thead{Model Call Ratio (in \%) \\ (L70B/L8B/L1B)} & \thead{Operating \\ Cost (in MJ)} & \thead{Request \\ Satisfaction (in \%)} & \thead{Model Call Ratio (in \%) \\ (L70B/L8B/L1B)} & \thead{Operating \\ Cost (in MJ)} & \thead{Request \\ Satisfaction (in \%)} & \thead{Model Call Ratio (in \%) \\ (L70B/L8B/L1B)} \\
    \cmidrule(lr){1-1}\cmidrule(lr){2-4}\cmidrule(lr){5-7}\cmidrule(lr){8-10}
    Llama 1B & 0.17$\scriptscriptstyle\pm0.00$ & \textcolor{red}{27.19$\scriptscriptstyle\pm4.94$} & 0\% / 0\% / 100\% & 0.07$\scriptscriptstyle\pm0.00$ & \textcolor{red}{74.05$\scriptscriptstyle\pm4.87$} & 0\% / 0\% / 100\% & 0.10$\scriptscriptstyle\pm0.00$ & \textcolor{red}{93.80$\scriptscriptstyle\pm2.68$} & 0\% / 0\% / 100\% \\
    Llama 8B & 0.81$\scriptscriptstyle\pm0.00$ & \textcolor{red}{29.03$\scriptscriptstyle\pm5.04$} & 0\% / 100\% / 0\% & \underline{0.36$\scriptscriptstyle\pm0.00$} & \textcolor{darkgreen}{79.33$\scriptscriptstyle\pm4.50$} & 0\% / 100\% / 0\% & \underline{0.44$\scriptscriptstyle\pm0.00$} & \textcolor{darkgreen}{97.00$\scriptscriptstyle\pm1.90$} & 0\% / 100\% / 0\% \\
    Llama 70B & \underline{4.11$\scriptscriptstyle\pm0.02$} & \textcolor{darkgreen}{49.31$\scriptscriptstyle\pm5.56$} & 100\% / 0\% / 0\% & 1.84$\scriptscriptstyle\pm0.01$ & \textcolor{darkgreen}{82.70$\scriptscriptstyle\pm4.20$} & 100\% / 0\% / 0\% & 2.23$\scriptscriptstyle\pm0.02$ & \textcolor{darkgreen}{97.10$\scriptscriptstyle\pm1.87$} & 100\% / 0\% / 0\% \\
    Educated Guessing & 2.51$\scriptscriptstyle\pm0.09$ & \textcolor{red}{39.88$\scriptscriptstyle\pm4.57$} & 56\% / 21\% / 22\% & \textbf{0.76$\scriptscriptstyle\pm0.04$} & \textcolor{darkgreen}{78.89$\scriptscriptstyle\pm1.52$} & 34\% / 32\% / 34\% & 0.92$\scriptscriptstyle\pm0.09$ & \textcolor{darkgreen}{96.51$\scriptscriptstyle\pm1.49$} & 31\% / 36\% / 32\% \\
    RouteLLM & 1.33$\scriptscriptstyle\pm0.04$ & \textcolor{darkgreen}{47.71$\scriptscriptstyle\pm3.38$} & 98\% / 0\% / 2\% & 1.25$\scriptscriptstyle\pm0.05$ & \textcolor{darkgreen}{78.35$\scriptscriptstyle\pm1.42$} & 66\% / 0\% / 34\% & 2.16$\scriptscriptstyle\pm0.04$ & \textcolor{darkgreen}{97.76$\scriptscriptstyle\pm0.73$} & 95\% / 0\% / 5\% \\
    RouterDC & \textbf{1.09$\scriptscriptstyle\pm0.08$} & \textcolor{darkgreen}{47.13$\scriptscriptstyle\pm3.08$} & 70\% / 29\% / 2\% & 1.85$\scriptscriptstyle\pm0.01$ & \textcolor{darkgreen}{82.34$\scriptscriptstyle\pm1.33$} & 100\% / 0\% / 0\% & 1.90$\scriptscriptstyle\pm0.07$ & \textcolor{darkgreen}{97.95$\scriptscriptstyle\pm0.81$} & 82\% / 18\% / 0\% \\
    \textbf{MESS+ (ours)} & 2.97$\scriptscriptstyle\pm0.09$ & \textcolor{darkgreen}{43.89$\scriptscriptstyle\pm4.52$} & 73\% / 3\% / 24\% & 0.84$\scriptscriptstyle\pm0.04$ & \textcolor{darkgreen}{79.23$\scriptscriptstyle\pm2.84$} & 45\% / 35\% / 21\% & \textbf{0.41$\scriptscriptstyle\pm0.05$} & \textcolor{darkgreen}{96.12$\scriptscriptstyle\pm2.33$} & 22\% / 33\% / 45\% \\
    \cmidrule(lr){1-1}\cmidrule(lr){2-4}\cmidrule(lr){5-7}\cmidrule(lr){8-10}
    \end{tabular}
    }

    \resizebox{\textwidth}{!}{
    \centering
    \begin{tabular}{lrrrrrrrrr}
    \cmidrule(lr){1-1}\cmidrule(lr){2-4}\cmidrule(lr){5-7}\cmidrule(lr){8-10}
    Category & \multicolumn{3}{c}{SocialIQA ($\alpha = 44\%$)} & \multicolumn{3}{c}{Winogrande ($\alpha = 70\%$)} & \multicolumn{3}{c}{\textit{Avg. across all Benchmarks ($\alpha = 66\%$)}} \\
    Subcategory & \thead{Operating \\ Cost (in MJ)} & \thead{Request \\ Satisfaction (in \%)} & \thead{Model Call Ratio (in \%) \\ (L70B/L8B/L1B)} & \thead{Operating \\ Cost (in MJ)} & \thead{Request \\ Satisfaction (in \%)} & \thead{Model Call Ratio (in \%) \\ (L70B/L8B/L1B)} & \thead{Operating \\ Cost (in MJ)} & \thead{Request \\ Satisfaction (in \%)} & \thead{Model Call Ratio (in \%) \\ (L70B/L8B/L1B)} \\
    \cmidrule(lr){1-1}\cmidrule(lr){2-4}\cmidrule(lr){5-7}\cmidrule(lr){8-10}
    Llama 1B & 0.13$\scriptscriptstyle\pm0.00$ & \textcolor{red}{41.71$\scriptscriptstyle\pm5.48$} & 0\% / 0\% / 100\% & 0.06$\scriptscriptstyle\pm0.00$ & \textcolor{red}{59.67$\scriptscriptstyle\pm5.45$} & 0\% / 0\% / 100\% & 0.12$\scriptscriptstyle\pm0.00$ & \textcolor{red}{58.28$\scriptscriptstyle\pm4.92$} & 0\% / 0\% / 100\% \\
    Llama 8B & \underline{0.59$\scriptscriptstyle\pm0.00$} & \textcolor{darkgreen}{48.31$\scriptscriptstyle\pm5.55$} & 0\% / 100\% / 0\% & \underline{0.25$\scriptscriptstyle\pm0.00$} & \textcolor{darkgreen}{73.64$\scriptscriptstyle\pm4.90$} & 0\% / 100\% / 0\% & \underline{0.54$\scriptscriptstyle\pm0.00$} & \textcolor{darkgreen}{68.20$\scriptscriptstyle\pm4.49$} & 0\% / 100\% / 0\% \\
    Llama 70B & 3.00$\scriptscriptstyle\pm0.00$ & \textcolor{darkgreen}{48.67$\scriptscriptstyle\pm5.56$} & 100\% / 0\% / 0\% & 1.29$\scriptscriptstyle\pm0.00$ & \textcolor{darkgreen}{79.08$\scriptscriptstyle\pm4.52$} & 100\% / 0\% / 0\% & 2.91$\scriptscriptstyle\pm0.01$ & \textcolor{darkgreen}{73.70$\scriptscriptstyle\pm4.35$} & 100\% / 0\% / 0\% \\
    Educated Guessing & \textbf{1.22$\scriptscriptstyle\pm0.06$} & \textcolor{darkgreen}{47.71$\scriptscriptstyle\pm2.50$} & 33\% / 32\% / 35\% & \textbf{0.54$\scriptscriptstyle\pm0.04$} & \textcolor{darkgreen}{70.67$\scriptscriptstyle\pm3.35$} & 35\% / 30\% / 35\% & \textbf{1.28$\scriptscriptstyle\pm0.07$} & \textcolor{darkgreen}{67.47$\scriptscriptstyle\pm2.73$} & 36\% / 31\% / 33\% \\
    RouteLLM & 2.02$\scriptscriptstyle\pm0.07$ & \textcolor{darkgreen}{44.32$\scriptscriptstyle\pm2.40$} & 65\% / 0\% / 35\% & 1.27$\scriptscriptstyle\pm0.02$ & \textcolor{darkgreen}{80.82$\scriptscriptstyle\pm2.53$} & 97\% / 0\% / 3\% & 2.04$\scriptscriptstyle\pm0.05$ & \textcolor{darkgreen}{71.19$\scriptscriptstyle\pm2.10$} & 82\% / 0\% / 18\% \\
    RouterDC & 2.89$\scriptscriptstyle\pm0.03$ & \textcolor{darkgreen}{46.76$\scriptscriptstyle\pm2.62$} & 95\% / 5\% / 0\% & 1.30$\scriptscriptstyle\pm0.00$ & \textcolor{darkgreen}{80.86$\scriptscriptstyle\pm2.49$} & 100\% / 0\% / 0\% & 2.11$\scriptscriptstyle\pm0.04$ & \textcolor{darkgreen}{73.17$\scriptscriptstyle\pm2.32$} & 85\% / 15\% / 0\% \\
    \textbf{MESS+ (ours)} & 1.46$\scriptscriptstyle\pm0.07$ & \textcolor{darkgreen}{46.21$\scriptscriptstyle\pm2.88$} & 43\% / 24\% / 33\% & 0.90$\scriptscriptstyle\pm0.04$ & \textcolor{darkgreen}{74.93$\scriptscriptstyle\pm2.52$} & 73\% / 8\% / 19\% & 1.79$\scriptscriptstyle\pm0.06$ & \textcolor{darkgreen}{69.16$\scriptscriptstyle\pm3.12$} & 52\% / 21\% / 27\% \\
    \cmidrule(lr){1-1}\cmidrule(lr){2-4}\cmidrule(lr){5-7}\cmidrule(lr){8-10}
    \end{tabular}
    }
\end{table}
\begin{table}[!ht]
\caption{Additional results for our main results with a smaller value for $V = 0.00001$, which reduces the priority for cost efficiency even further.}
\label{tab:results_part13}
\resizebox{\textwidth}{!}{
    \centering
    \begin{tabular}{lrrrrrrrrr}
        \cmidrule(lr){1-1}\cmidrule(lr){2-4}\cmidrule(lr){5-7}\cmidrule(lr){8-10}
        Category & \multicolumn{3}{c}{ARC Challenge ($\alpha = 50\%$)} & \multicolumn{3}{c}{ARC Easy ($\alpha = 75\%$)} & \multicolumn{3}{c}{BoolQ  ($\alpha = 80\%$)} \\
        Subcategory & \thead{Operating \\ Cost (in MJ)} & \thead{Request \\ Satisfaction (in \%)} & \thead{Model Call Ratio (in \%) \\ (L70B/L8B/L1B)} & \thead{Operating \\ Cost (in MJ)} & \thead{Request \\ Satisfaction (in \%)} & \thead{Model Call Ratio (in \%) \\ (L70B/L8B/L1B)} & \thead{Operating \\ Cost (in MJ)} & \thead{Request \\ Satisfaction (in \%)} & \thead{Model Call Ratio (in \%) \\ (L70B/L8B/L1B)} \\
        \cmidrule(lr){1-1}\cmidrule(lr){2-4}\cmidrule(lr){5-7}\cmidrule(lr){8-10}
        Llama 1B & 0.09$\scriptscriptstyle\pm0.00$ & \textcolor{red}{37.88$\scriptscriptstyle\pm5.39$} & 0\% / 0\% / 100\% & 0.20$\scriptscriptstyle\pm0.00$ & \textcolor{red}{62.76$\scriptscriptstyle\pm5.37$} & 0\% / 0\% / 100\% & 0.14$\scriptscriptstyle\pm0.00$ & \textcolor{red}{69.17$\scriptscriptstyle\pm5.13$} & 0\% / 0\% / 100\% \\
        Llama 8B & \underline{0.46$\scriptscriptstyle\pm0.00$} & \textcolor{darkgreen}{54.44$\scriptscriptstyle\pm5.54$} & 0\% / 100\% / 0\% & \underline{0.97$\scriptscriptstyle\pm0.00$} & \textcolor{darkgreen}{79.72$\scriptscriptstyle\pm4.47$} & 0\% / 100\% / 0\% & \underline{0.43$\scriptscriptstyle\pm0.00$} & \textcolor{darkgreen}{84.16$\scriptscriptstyle\pm4.06$} & 0\% / 100\% / 0\% \\
        Llama 70B & 2.35$\scriptscriptstyle\pm0.01$ & \textcolor{darkgreen}{60.84$\scriptscriptstyle\pm5.43$} & 100\% / 0\% / 0\% & 5.05$\scriptscriptstyle\pm0.01$ & \textcolor{darkgreen}{83.12$\scriptscriptstyle\pm4.16$} & 100\% / 0\% / 0\% & 3.40$\scriptscriptstyle\pm0.00$ & \textcolor{darkgreen}{88.78$\scriptscriptstyle\pm3.51$} & 100\% / 0\% / 0\% \\
        Educated Guessing & \textbf{1.00$\scriptscriptstyle\pm0.09$} & \textcolor{darkgreen}{51.65$\scriptscriptstyle\pm2.98$} & 35\% / 31\% / 34\% & \textbf{2.00$\scriptscriptstyle\pm0.08$} & \textcolor{red}{74.00$\scriptscriptstyle\pm4.39$} & 31\% / 32\% / 36\% & \textbf{1.31$\scriptscriptstyle\pm0.04$} & \textcolor{darkgreen}{80.47$\scriptscriptstyle\pm1.08$} & 33\% / 34\% / 33\% \\
        RouteLLM & 1.24$\scriptscriptstyle\pm0.10$ & \textcolor{darkgreen}{51.17$\scriptscriptstyle\pm2.93$} & 50\% / 0\% / 50\% & 4.05$\scriptscriptstyle\pm0.01$ & \textcolor{darkgreen}{82.54$\scriptscriptstyle\pm2.12$} & 100\% / 0\% / 0\% & 2.96$\scriptscriptstyle\pm0.04$ & \textcolor{darkgreen}{86.83$\scriptscriptstyle\pm1.27$} & 87\% / 0\% / 13\% \\
        RouterDC & 2.09$\scriptscriptstyle\pm0.06$ & \textcolor{darkgreen}{60.94$\scriptscriptstyle\pm2.92$} & 88\% / 12\% / 0\% & 3.61$\scriptscriptstyle\pm0.06$ & \textcolor{darkgreen}{82.30$\scriptscriptstyle\pm2.60$} & 85\% / 15\% / 0\% & 2.14$\scriptscriptstyle\pm0.05$ & \textcolor{darkgreen}{87.06$\scriptscriptstyle\pm2.70$} & 58\% / 42\% / 0\% \\
        \textbf{MESS+ (ours)} & 1.61$\scriptscriptstyle\pm0.09$ & \textcolor{darkgreen}{54.31$\scriptscriptstyle\pm2.87$} & 68\% / 11\% / 21\% & 5.39$\scriptscriptstyle\pm0.09$ & \textcolor{darkgreen}{78.28$\scriptscriptstyle\pm1.63$} & 65\% / 16\% / 19\% & 1.38$\scriptscriptstyle\pm0.05$ & \textcolor{darkgreen}{82.25$\scriptscriptstyle\pm3.11$} & 40\% / 33\% / 27\% \\
        \cmidrule(lr){1-1}\cmidrule(lr){2-4}\cmidrule(lr){5-7}\cmidrule(lr){8-10}
        \end{tabular}
    }

    \resizebox{\textwidth}{!}{
    \centering
    \begin{tabular}{lrrrrrrrrr}
    \cmidrule(lr){1-1}\cmidrule(lr){2-4}\cmidrule(lr){5-7}\cmidrule(lr){8-10}
    Category & \multicolumn{3}{c}{LogiQA ($\alpha = 40\%$)} & \multicolumn{3}{c}{PiQA ($\alpha = 78\%$)} & \multicolumn{3}{c}{SciQ ($\alpha = 96\%$)} \\
    Subcategory & \thead{Operating \\ Cost (in MJ)} & \thead{Request \\ Satisfaction (in \%)} & \thead{Model Call Ratio (in \%) \\ (L70B/L8B/L1B)} & \thead{Operating \\ Cost (in MJ)} & \thead{Request \\ Satisfaction (in \%)} & \thead{Model Call Ratio (in \%) \\ (L70B/L8B/L1B)} & \thead{Operating \\ Cost (in MJ)} & \thead{Request (in \%) \\ Satisfaction} & \thead{Model Call Ratio (in \%) \\ (L70B/L8B/L1B)} \\
    \cmidrule(lr){1-1}\cmidrule(lr){2-4}\cmidrule(lr){5-7}\cmidrule(lr){8-10}
    Llama 1B & 0.17$\scriptscriptstyle\pm0.00$ & \textcolor{red}{27.19$\scriptscriptstyle\pm4.94$} & 0\% / 0\% / 100\% & 0.07$\scriptscriptstyle\pm0.00$ & \textcolor{red}{74.05$\scriptscriptstyle\pm4.87$} & 0\% / 0\% / 100\% & 0.10$\scriptscriptstyle\pm0.00$ & \textcolor{red}{93.80$\scriptscriptstyle\pm2.68$} & 0\% / 0\% / 100\% \\
    Llama 8B & 0.81$\scriptscriptstyle\pm0.00$ & \textcolor{red}{29.03$\scriptscriptstyle\pm5.04$} & 0\% / 100\% / 0\% & \underline{0.36$\scriptscriptstyle\pm0.00$} & \textcolor{darkgreen}{79.33$\scriptscriptstyle\pm4.50$} & 0\% / 100\% / 0\% & \underline{0.44$\scriptscriptstyle\pm0.00$} & \textcolor{darkgreen}{97.00$\scriptscriptstyle\pm1.90$} & 0\% / 100\% / 0\% \\
    Llama 70B & \underline{4.11$\scriptscriptstyle\pm0.02$} & \textcolor{darkgreen}{49.31$\scriptscriptstyle\pm5.56$} & 100\% / 0\% / 0\% & 1.84$\scriptscriptstyle\pm0.01$ & \textcolor{darkgreen}{82.70$\scriptscriptstyle\pm4.20$} & 100\% / 0\% / 0\% & 2.23$\scriptscriptstyle\pm0.02$ & \textcolor{darkgreen}{97.10$\scriptscriptstyle\pm1.87$} & 100\% / 0\% / 0\% \\
    Educated Guessing & 2.51$\scriptscriptstyle\pm0.09$ & \textcolor{red}{39.88$\scriptscriptstyle\pm4.57$} & 56\% / 21\% / 22\% & \textbf{0.76$\scriptscriptstyle\pm0.04$} & \textcolor{darkgreen}{78.89$\scriptscriptstyle\pm1.52$} & 34\% / 32\% / 34\% & 0.92$\scriptscriptstyle\pm0.09$ & \textcolor{darkgreen}{96.51$\scriptscriptstyle\pm1.49$} & 31\% / 36\% / 32\% \\
    RouteLLM & 1.33$\scriptscriptstyle\pm0.04$ & \textcolor{darkgreen}{47.71$\scriptscriptstyle\pm3.38$} & 98\% / 0\% / 2\% & 1.25$\scriptscriptstyle\pm0.05$ & \textcolor{darkgreen}{78.35$\scriptscriptstyle\pm1.42$} & 66\% / 0\% / 34\% & 2.16$\scriptscriptstyle\pm0.04$ & \textcolor{darkgreen}{97.76$\scriptscriptstyle\pm0.73$} & 95\% / 0\% / 5\% \\
    RouterDC & \textbf{1.09$\scriptscriptstyle\pm0.08$} & \textcolor{darkgreen}{47.13$\scriptscriptstyle\pm3.08$} & 70\% / 29\% / 2\% & 1.85$\scriptscriptstyle\pm0.01$ & \textcolor{darkgreen}{82.34$\scriptscriptstyle\pm1.33$} & 100\% / 0\% / 0\% & 1.90$\scriptscriptstyle\pm0.07$ & \textcolor{darkgreen}{97.95$\scriptscriptstyle\pm0.81$} & 82\% / 18\% / 0\% \\
    \textbf{MESS+ (ours)} & 2.86$\scriptscriptstyle\pm0.09$ & \textcolor{darkgreen}{43.89$\scriptscriptstyle\pm4.55$} & 72\% / 2\% / 26\% & 1.01$\scriptscriptstyle\pm0.04$ & \textcolor{darkgreen}{79.94$\scriptscriptstyle\pm2.62$} & 51\% / 38\% / 12\% & \textbf{0.56$\scriptscriptstyle\pm0.07$} & \textcolor{darkgreen}{96.17$\scriptscriptstyle\pm2.12$} & 20\% / 37\% / 43\% \\
    \cmidrule(lr){1-1}\cmidrule(lr){2-4}\cmidrule(lr){5-7}\cmidrule(lr){8-10}
    \end{tabular}
    }

    \resizebox{\textwidth}{!}{
    \centering
    \begin{tabular}{lrrrrrrrrr}
\cmidrule(lr){1-1}\cmidrule(lr){2-4}\cmidrule(lr){5-7}\cmidrule(lr){8-10}
Category & \multicolumn{3}{c}{SocialIQA ($\alpha = 44\%$)} & \multicolumn{3}{c}{Winogrande ($\alpha = 70\%$)} & \multicolumn{3}{c}{\textit{Avg. across all Benchmarks ($\alpha = 66\%$)}} \\
Subcategory & \thead{Operating \\ Cost (in MJ)} & \thead{Request \\ Satisfaction (in \%)} & \thead{Model Call Ratio (in \%) \\ (L70B/L8B/L1B)} & \thead{Operating \\ Cost (in MJ)} & \thead{Request \\ Satisfaction (in \%)} & \thead{Model Call Ratio (in \%) \\ (L70B/L8B/L1B)} & \thead{Operating \\ Cost (in MJ)} & \thead{Request \\ Satisfaction (in \%)} & \thead{Model Call Ratio (in \%) \\ (L70B/L8B/L1B)} \\
\cmidrule(lr){1-1}\cmidrule(lr){2-4}\cmidrule(lr){5-7}\cmidrule(lr){8-10}
Llama 1B & 0.13$\scriptscriptstyle\pm0.00$ & \textcolor{red}{41.71$\scriptscriptstyle\pm5.48$} & 0\% / 0\% / 100\% & 0.06$\scriptscriptstyle\pm0.00$ & \textcolor{red}{59.67$\scriptscriptstyle\pm5.45$} & 0\% / 0\% / 100\% & 0.12$\scriptscriptstyle\pm0.00$ & \textcolor{red}{58.28$\scriptscriptstyle\pm4.92$} & 0\% / 0\% / 100\% \\
Llama 8B & \underline{0.59$\scriptscriptstyle\pm0.00$} & \textcolor{darkgreen}{48.31$\scriptscriptstyle\pm5.55$} & 0\% / 100\% / 0\% & \underline{0.25$\scriptscriptstyle\pm0.00$} & \textcolor{darkgreen}{73.64$\scriptscriptstyle\pm4.90$} & 0\% / 100\% / 0\% & \underline{0.54$\scriptscriptstyle\pm0.00$} & \textcolor{darkgreen}{68.20$\scriptscriptstyle\pm4.49$} & 0\% / 100\% / 0\% \\
Llama 70B & 3.00$\scriptscriptstyle\pm0.00$ & \textcolor{darkgreen}{48.67$\scriptscriptstyle\pm5.56$} & 100\% / 0\% / 0\% & 1.29$\scriptscriptstyle\pm0.00$ & \textcolor{darkgreen}{79.08$\scriptscriptstyle\pm4.52$} & 100\% / 0\% / 0\% & 2.91$\scriptscriptstyle\pm0.01$ & \textcolor{darkgreen}{73.70$\scriptscriptstyle\pm4.35$} & 100\% / 0\% / 0\% \\
Educated Guessing & \textbf{1.22$\scriptscriptstyle\pm0.06$} & \textcolor{darkgreen}{47.71$\scriptscriptstyle\pm2.50$} & 33\% / 32\% / 35\% & \textbf{0.54$\scriptscriptstyle\pm0.04$} & \textcolor{darkgreen}{70.67$\scriptscriptstyle\pm3.35$} & 35\% / 30\% / 35\% & \textbf{1.28$\scriptscriptstyle\pm0.07$} & \textcolor{darkgreen}{67.47$\scriptscriptstyle\pm2.73$} & 36\% / 31\% / 33\% \\
RouteLLM & 2.02$\scriptscriptstyle\pm0.07$ & \textcolor{darkgreen}{44.32$\scriptscriptstyle\pm2.40$} & 65\% / 0\% / 35\% & 1.27$\scriptscriptstyle\pm0.02$ & \textcolor{darkgreen}{80.82$\scriptscriptstyle\pm2.53$} & 97\% / 0\% / 3\% & 2.04$\scriptscriptstyle\pm0.05$ & \textcolor{darkgreen}{71.19$\scriptscriptstyle\pm2.10$} & 82\% / 0\% / 18\% \\
RouterDC & 2.89$\scriptscriptstyle\pm0.03$ & \textcolor{darkgreen}{46.76$\scriptscriptstyle\pm2.62$} & 95\% / 5\% / 0\% & 1.30$\scriptscriptstyle\pm0.00$ & \textcolor{darkgreen}{80.86$\scriptscriptstyle\pm2.49$} & 100\% / 0\% / 0\% & 2.11$\scriptscriptstyle\pm0.04$ & \textcolor{darkgreen}{73.17$\scriptscriptstyle\pm2.32$} & 85\% / 15\% / 0\% \\
\textbf{MESS+ (ours)} & 1.34$\scriptscriptstyle\pm0.06$ & \textcolor{darkgreen}{46.75$\scriptscriptstyle\pm2.64$} & 45\% / 35\% / 20\% & 0.87$\scriptscriptstyle\pm0.04$ & \textcolor{darkgreen}{74.69$\scriptscriptstyle\pm2.55$} & 71\% / 10\% / 18\% & 1.88$\scriptscriptstyle\pm0.07$ & \textcolor{darkgreen}{69.41$\scriptscriptstyle\pm2.76$} & 54\% / 23\% / 23\% \\
\cmidrule(lr){1-1}\cmidrule(lr){2-4}\cmidrule(lr){5-7}\cmidrule(lr){8-10}
\end{tabular}
    }
\end{table}

\begin{figure}
    \centering
    \begin{subfigure}{\textwidth}
        \includegraphics[width=\textwidth]{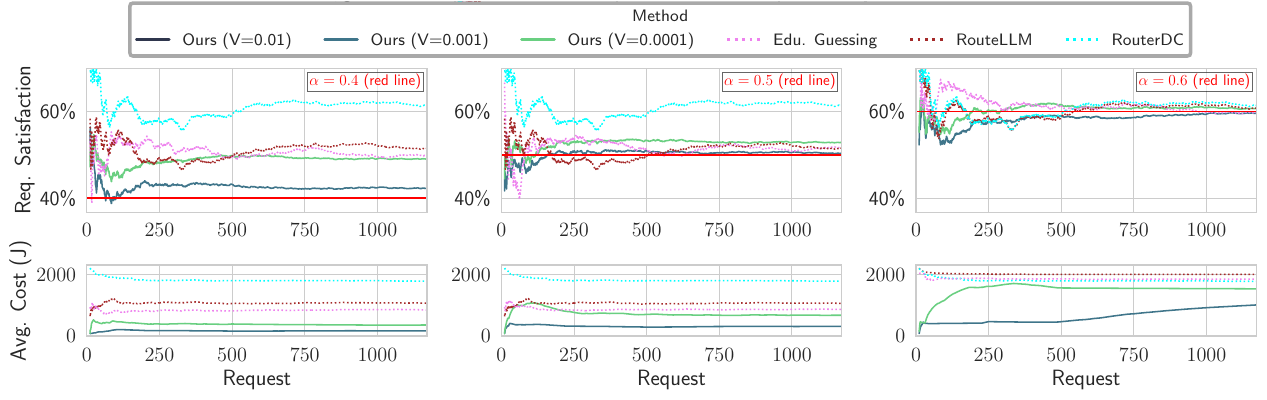}
        \caption{ARC Challenge}
    \end{subfigure}
    \hfill
    \begin{subfigure}{\textwidth}
        \includegraphics[width=\textwidth]{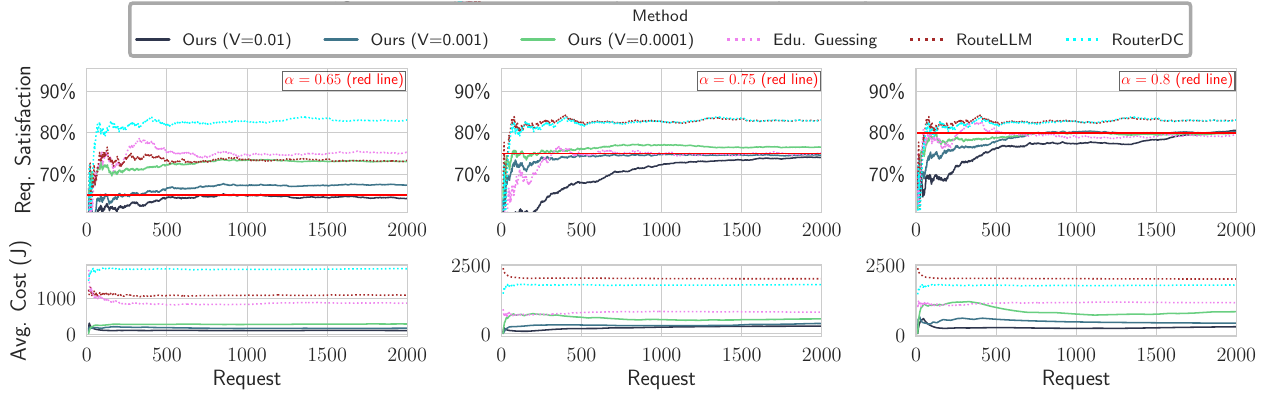}
        \caption{ARC Easy}
    \end{subfigure}
    \hfill
    \begin{subfigure}{\textwidth}
        \includegraphics[width=\textwidth]{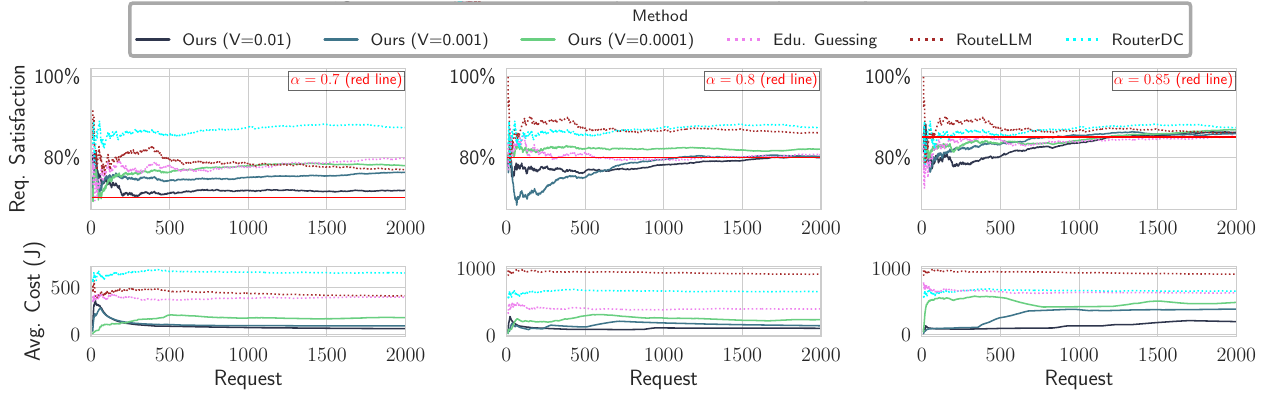}
        \caption{BoolQ}
    \end{subfigure}
    \hfill
    \begin{subfigure}{\textwidth}
        \includegraphics[width=\textwidth]{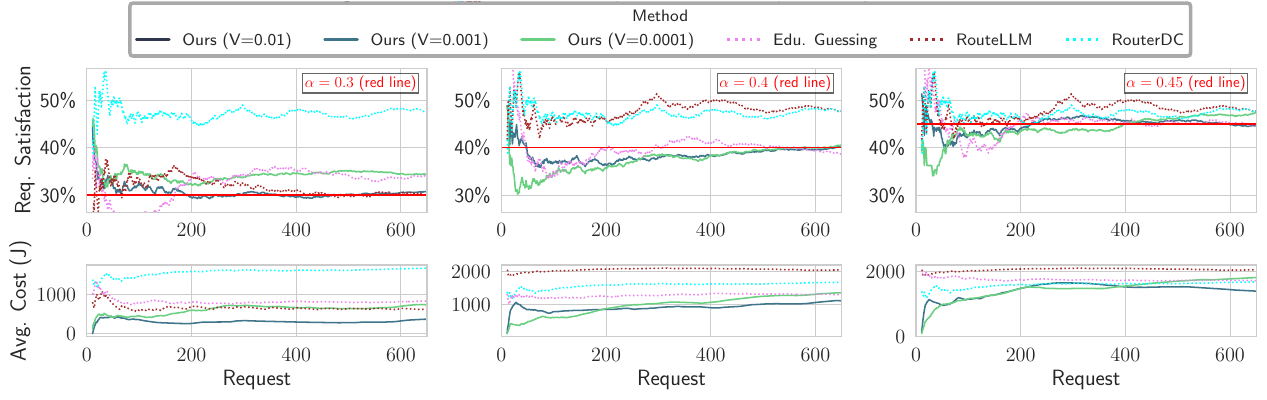}
        \caption{LogiQA}
    \end{subfigure}

    \caption{The dynamics between $\alpha$ and $V$ manifest further across all benchmarks in addition to Winogrande in the main paper. Part 1.}
    \label{fig:alpha_v_interplay_appendix1}
\end{figure}

\begin{figure}
    \centering
    \begin{subfigure}{\textwidth}
        \includegraphics[width=\textwidth]{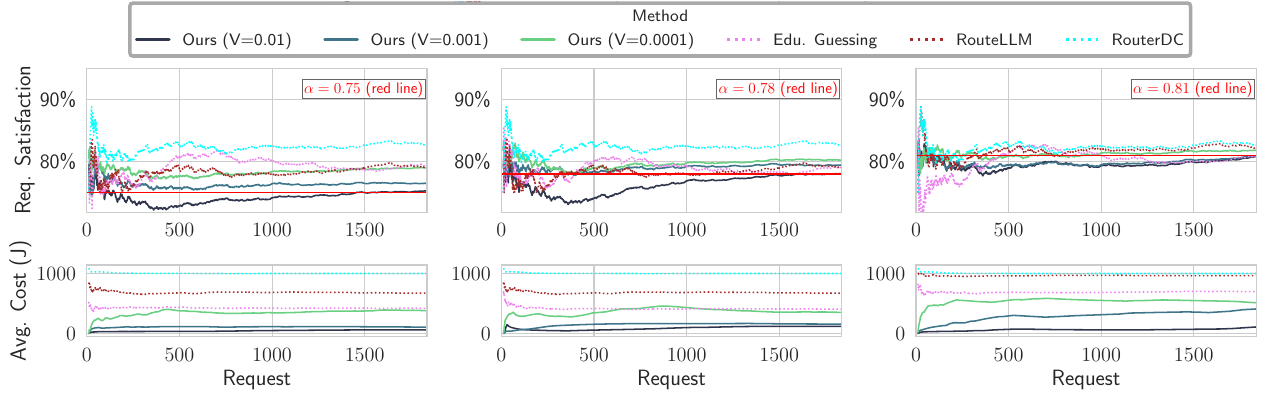}
        \caption{PiQA}
    \end{subfigure}
    \hfill
    \begin{subfigure}{\textwidth}
        \includegraphics[width=\textwidth]{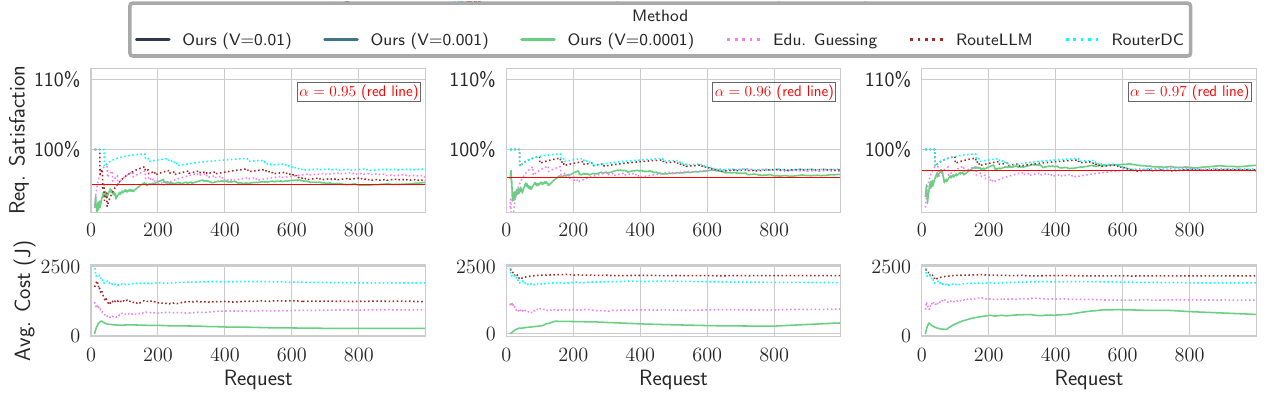}
        \caption{SciQ}
    \end{subfigure}
    \hfill
    \begin{subfigure}{\textwidth}
        \includegraphics[width=\textwidth]{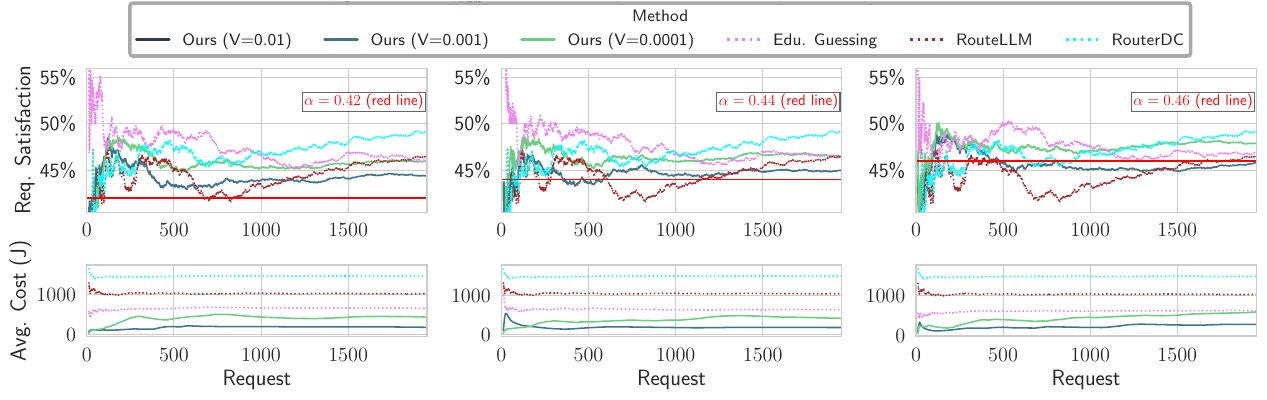}
        \caption{SocialIQA}
    \end{subfigure}

    \caption{The dynamics between $\alpha$ and $V$ manifest further across all benchmarks in addition to Winogrande in the main paper. Part 2.}
    \label{fig:alpha_v_interplay_appendix2}
\end{figure}




\subsection{Additional Experiments on Larger Model Zoos}
To demonstrate the performance of \algname even in larger zoos, we conduct additional experiments on a zoo containing four models: Qwen 2.5 32B (Q32B), Qwen 2 7B (Q7B), Qwen 2 1.5B (Q1.5B), and Qwen 2 0.5B (Q0.5B) with the checkpoints, \texttt{unsloth/Qwen2.5-32B-Instruct-bnb-4bit}, \texttt{unsloth/Qwen2-7B-bnb-4bit}, \texttt{Qwen/Qwen2-1.5B-Instruct}, and \texttt{Qwen/Qwen2-0.5B-Instruct}, respectively. 
All checkpoints are readily available on the HuggingFace Hub. We leave all other hyperparameters unchanged and follow the setup we use in our Llama3 model-only zoo.
Aside from the Llama 3-only and Qwen 2-only model zoos, we also provide results for a mixed model zoo and an evaluation of varied operating cost characteristics. 

\subsubsection{Supplementary Results in Support of the Main Findings with Four Qwen 2 Models}

\begin{table}
\caption{Main results with the Qwen 2 model family. Specifically, we use Qwen 2.5 32B, Qwen 2 7B, Qwen 2 1.5B, and Qwen 2 0.5B. \algname also outperforms our Educated Guessing baseline in larger model zoos. $V = 0.0001$.}
\label{tab:results_part1}
    \resizebox{\textwidth}{!}{
        \begin{tabular}{lccccccccc}
\cmidrule(lr){1-1}\cmidrule(lr){2-4}\cmidrule(lr){5-7}\cmidrule(lr){8-10}
Category & \multicolumn{3}{c}{ARC Challenge ($\alpha = 55\%$)} & \multicolumn{3}{c}{ARC Easy ($\alpha = 77\%$)} & \multicolumn{3}{c}{BoolQ  ($\alpha = 80\%$)} \\
Subcategory & \thead{Operating \\ Cost} & \thead{Request. \\ Satisfaction} & \thead{Model Call Ratio \\ (Q32B/Q7B/Q1.5B/Q0.5B)} & \thead{Operating \\ Cost} & \thead{Request. \\ Satisfaction} & \thead{Model Call Ratio \\ (Q32B/Q7B/Q1.5B/Q0.5B)} & \thead{Operating \\ Cost} & \thead{Request. \\ Satisfaction} & \thead{Model Call Ratio \\ (Q32B/Q7B/Q1.5B/Q0.5B)} \\
\cmidrule(lr){1-1}\cmidrule(lr){2-4}\cmidrule(lr){5-7}\cmidrule(lr){8-10}
Qwen2 0.5B & 0.10$\scriptscriptstyle\pm0.00$ & \textcolor{red}{30.03$\scriptscriptstyle\pm45.86$} & 0\% / 0\% / 0\% / 100\% & 0.21$\scriptscriptstyle\pm0.00$ & \textcolor{red}{54.88$\scriptscriptstyle\pm49.77$} & 0\% / 0\% / 0\% / 100\% & 0.11$\scriptscriptstyle\pm0.00$ & \textcolor{red}{63.09$\scriptscriptstyle\pm48.26$} & 0\% / 0\% / 0\% / 100\% \\
Qwen2 1.5B & 0.14$\scriptscriptstyle\pm0.00$ & \textcolor{red}{40.10$\scriptscriptstyle\pm5.45$} & 0\% / 0\% / 100\% / 0\% & 0.28$\scriptscriptstyle\pm0.00$ & \textcolor{red}{66.62$\scriptscriptstyle\pm5.24$} & 0\% / 0\% / 100\% / 0\% & 0.12$\scriptscriptstyle\pm0.00$ & \textcolor{red}{76.27$\scriptscriptstyle\pm4.73$} & 0\% / 0\% / 100\% / 0\% \\
Qwen2 7B & 0.31$\scriptscriptstyle\pm0.00$ & \textcolor{red}{50.94$\scriptscriptstyle\pm5.56$} & 0\% / 100\% / 0\% / 0\% & 0.70$\scriptscriptstyle\pm0.00$ & \textcolor{red}{75.42$\scriptscriptstyle\pm4.78$} & 0\% / 100\% / 0\% / 0\% & \underline{0.25$\scriptscriptstyle\pm0.00$} & \textcolor{darkgreen}{84.13$\scriptscriptstyle\pm4.06$} & 0\% / 100\% / 0\% / 0\% \\
Qwen2.5 32B & \underline{1.33$\scriptscriptstyle\pm0.00$} & \textcolor{darkgreen}{58.28$\scriptscriptstyle\pm5.48$} & 100\% / 0\% / 0\% / 0\% & \underline{2.73$\scriptscriptstyle\pm0.00$} & \textcolor{darkgreen}{78.20$\scriptscriptstyle\pm4.59$} & 100\% / 0\% / 0\% / 0\% & 1.63$\scriptscriptstyle\pm0.00$ & \textcolor{darkgreen}{89.60$\scriptscriptstyle\pm3.39$} & 100\% / 0\% / 0\% / 0\% \\
Educated Guessing & 1.26$\scriptscriptstyle\pm0.00$ & \textcolor{darkgreen}{56.76$\scriptscriptstyle\pm3.66$} & 82\% / 15\% / 2\% / 1\% & 2.22$\scriptscriptstyle\pm0.00$ & \textcolor{darkgreen}{77.46$\scriptscriptstyle\pm1.27$} & 76\% / 20\% / 2\% / 2\% & 0.81$\scriptscriptstyle\pm0.00$ & \textcolor{darkgreen}{82.40$\scriptscriptstyle\pm1.15$} & 41\% / 34\% / 14\% / 11\% \\
RouteLLM & 1.33$\scriptscriptstyle\pm0.01$ & \textcolor{darkgreen}{58.16$\scriptscriptstyle\pm2.56$} & 100\% / 0\% / 0\% / 0\% & 2.73$\scriptscriptstyle\pm0.01$ & \textcolor{darkgreen}{77.60$\scriptscriptstyle\pm2.79$} & 99\% / 0\% / 0\% / 1\% & 1.43$\scriptscriptstyle\pm0.00$ & \textcolor{darkgreen}{87.31$\scriptscriptstyle\pm1.96$} & 87\% / 0\% / 0\% / 13\% \\
RouterDC & 1.26$\scriptscriptstyle\pm0.00$ & \textcolor{darkgreen}{58.65$\scriptscriptstyle\pm2.63$} & 89\% / 2\% / 9\% / 0\% & 2.03$\scriptscriptstyle\pm0.00$ & \textcolor{darkgreen}{77.44$\scriptscriptstyle\pm4.15$} & 70\% / 19\% / 3\% / 8\% & 1.45$\scriptscriptstyle\pm0.00$ & \textcolor{darkgreen}{89.41$\scriptscriptstyle\pm1.52$} & 87\% / 8\% / 4\% / 1\% \\
\textbf{MESS+ (ours)} & \textbf{1.18$\scriptscriptstyle\pm0.00$} & \textcolor{darkgreen}{56.21$\scriptscriptstyle\pm4.26$} & 83\% / 7\% / 2\% / 8\% & \textbf{1.69$\scriptscriptstyle\pm0.00$} & \textcolor{darkgreen}{77.06$\scriptscriptstyle\pm2.17$} & 53\% / 42\% / 2\% / 3\% & \textbf{0.78$\scriptscriptstyle\pm0.00$} & \textcolor{darkgreen}{82.48$\scriptscriptstyle\pm3.16$} & 40\% / 32\% / 11\% / 17\% \\
\cmidrule(lr){1-1}\cmidrule(lr){2-4}\cmidrule(lr){5-7}\cmidrule(lr){8-10}
\end{tabular}

    }
    
    \resizebox{\textwidth}{!}{
        \begin{tabular}{lccccccccc}
\cmidrule(lr){1-1}\cmidrule(lr){2-4}\cmidrule(lr){5-7}\cmidrule(lr){8-10}
Category & \multicolumn{3}{c}{LogiQA ($\alpha = 33\%$)} & \multicolumn{3}{c}{PiQA ($\alpha = 79\%$)} & \multicolumn{3}{c}{SciQ ($\alpha = 93\%$)} \\
Subcategory & \thead{Operating \\ Cost} & \thead{Request. \\ Satisfaction} & \thead{Model Call Ratio \\ (Q32B/Q7B/Q1.5B/Q0.5B)} & \thead{Operating \\ Cost} & \thead{Request. \\ Satisfaction} & \thead{Model Call Ratio \\ (Q32B/Q7B/Q1.5B/Q0.5B)} & \thead{Operating \\ Cost} & \thead{Request. \\ Satisfaction} & \thead{Model Call Ratio \\ (Q32B/Q7B/Q1.5B/Q0.5B)} \\
\cmidrule(lr){1-1}\cmidrule(lr){2-4}\cmidrule(lr){5-7}\cmidrule(lr){8-10}
Qwen2 0.5B & 0.06$\scriptscriptstyle\pm0.00$ & \textcolor{red}{25.35$\scriptscriptstyle\pm43.53$} & 0\% / 0\% / 0\% / 100\% & 0.09$\scriptscriptstyle\pm0.00$ & \textcolor{red}{69.15$\scriptscriptstyle\pm46.20$} & 0\% / 0\% / 0\% / 100\% & 0.09$\scriptscriptstyle\pm0.00$ & \textcolor{red}{91.20$\scriptscriptstyle\pm28.34$} & 0\% / 0\% / 0\% / 100\% \\
Qwen2 1.5B & 0.08$\scriptscriptstyle\pm0.00$ & \textcolor{red}{24.27$\scriptscriptstyle\pm4.77$} & 0\% / 0\% / 100\% / 0\% & 0.11$\scriptscriptstyle\pm0.00$ & \textcolor{red}{76.06$\scriptscriptstyle\pm4.74$} & 0\% / 0\% / 100\% / 0\% & \underline{0.12$\scriptscriptstyle\pm0.00$} & \textcolor{darkgreen}{94.40$\scriptscriptstyle\pm2.56$} & 0\% / 0\% / 100\% / 0\% \\
Qwen2 7B & 0.06$\scriptscriptstyle\pm0.00$ & \textcolor{red}{31.18$\scriptscriptstyle\pm5.15$} & 0\% / 100\% / 0\% / 0\% & \underline{0.25$\scriptscriptstyle\pm0.00$} & \textcolor{darkgreen}{79.49$\scriptscriptstyle\pm4.49$} & 0\% / 100\% / 0\% / 0\% & 0.34$\scriptscriptstyle\pm0.00$ & \textcolor{darkgreen}{95.50$\scriptscriptstyle\pm2.30$} & 0\% / 100\% / 0\% / 0\% \\
Qwen2.5 32B & \underline{0.80$\scriptscriptstyle\pm0.00$} & \textcolor{darkgreen}{40.86$\scriptscriptstyle\pm5.47$} & 100\% / 0\% / 0\% / 0\% & 1.06$\scriptscriptstyle\pm0.00$ & \textcolor{darkgreen}{80.41$\scriptscriptstyle\pm4.41$} & 100\% / 0\% / 0\% / 0\% & 1.19$\scriptscriptstyle\pm0.00$ & \textcolor{darkgreen}{96.70$\scriptscriptstyle\pm1.99$} & 100\% / 0\% / 0\% / 0\% \\
Educated Guessing & 0.39$\scriptscriptstyle\pm0.00$ & \textcolor{darkgreen}{33.08$\scriptscriptstyle\pm3.70$} & 50\% / 22\% / 18\% / 10\% & 0.71$\scriptscriptstyle\pm0.00$ & \textcolor{darkgreen}{79.02$\scriptscriptstyle\pm2.16$} & 55\% / 31\% / 6\% / 8\% & 0.44$\scriptscriptstyle\pm0.00$ & \textcolor{darkgreen}{94.35$\scriptscriptstyle\pm1.15$} & 25\% / 24\% / 25\% / 27\% \\
RouteLLM & 0.53$\scriptscriptstyle\pm0.00$ & \textcolor{darkgreen}{33.62$\scriptscriptstyle\pm3.75$} & 64\% / 0\% / 0\% / 36\% & 1.01$\scriptscriptstyle\pm0.00$ & \textcolor{darkgreen}{79.28$\scriptscriptstyle\pm1.58$} & 95\% / 0\% / 0\% / 5\% & 0.68$\scriptscriptstyle\pm0.00$ & \textcolor{darkgreen}{94.72$\scriptscriptstyle\pm1.34$} & 53\% / 0\% / 0\% / 47\% \\
RouterDC & 0.38$\scriptscriptstyle\pm0.00$ & \textcolor{darkgreen}{33.16$\scriptscriptstyle\pm3.20$} & 50\% / 31\% / 12\% / 7\% & 0.74$\scriptscriptstyle\pm0.00$ & \textcolor{darkgreen}{79.86$\scriptscriptstyle\pm1.67$} & 63\% / 17\% / 17\% / 3\% & 0.46$\scriptscriptstyle\pm0.00$ & \textcolor{darkgreen}{94.97$\scriptscriptstyle\pm0.79$} & 30\% / 16\% / 12\% / 42\% \\
\textbf{MESS+ (ours)} & \textbf{0.34$\scriptscriptstyle\pm0.00$} & \textcolor{darkgreen}{33.16$\scriptscriptstyle\pm5.62$} & 49\% / 33\% / 1\% / 17\% & \textbf{0.66$\scriptscriptstyle\pm0.00$} & \textcolor{darkgreen}{79.30$\scriptscriptstyle\pm2.54$} & 56\% / 25\% / 13\% / 6\% & \textbf{0.21$\scriptscriptstyle\pm0.00$} & \textcolor{darkgreen}{93.91$\scriptscriptstyle\pm1.75$} & 18\% / 3\% / 22\% / 57\% \\
\cmidrule(lr){1-1}\cmidrule(lr){2-4}\cmidrule(lr){5-7}\cmidrule(lr){8-10}
\end{tabular}
    }
    \resizebox{\textwidth}{!}{
        \begin{tabular}{lccccccccc}
\cmidrule(lr){1-1}\cmidrule(lr){2-4}\cmidrule(lr){5-7}\cmidrule(lr){8-10}
Category & \multicolumn{3}{c}{SocialIQA ($\alpha = 47\%$)} & \multicolumn{3}{c}{Winogrande ($\alpha = 71\%$)} & \multicolumn{3}{c}{Mean ($\alpha = 0.67$)} \\
Subcategory & \thead{Operating \\ Cost} & \thead{Request. \\ Satisfaction} & \thead{Model Call Ratio \\ (Q32B/Q7B/Q1.5B/Q0.5B)} & \thead{Operating \\ Cost} & \thead{Request. \\ Satisfaction} & \thead{Model Call Ratio \\ (Q32B/Q7B/Q1.5B/Q0.5B)} & \thead{Operating \\ Cost} & \thead{Request. \\ Satisfaction} & \thead{Model Call Ratio \\ (Q32B/Q7B/Q1.5B/Q0.5B)} \\
\cmidrule(lr){1-1}\cmidrule(lr){2-4}\cmidrule(lr){5-7}\cmidrule(lr){8-10}
Qwen2 0.5B & 0.26$\scriptscriptstyle\pm0.00$ & \textcolor{red}{43.35$\scriptscriptstyle\pm49.56$} & 0\% / 0\% / 0\% / 100\% & 0.06$\scriptscriptstyle\pm0.00$ & \textcolor{red}{55.88$\scriptscriptstyle\pm49.67$} & 0\% / 0\% / 0\% / 100\% & 0.12$\scriptscriptstyle\pm0.00$ & \textcolor{red}{54.12$\scriptscriptstyle\pm45.15$} & 0\% / 0\% / 0\% / 100\% \\
Qwen2 1.5B & 0.36$\scriptscriptstyle\pm0.00$ & \textcolor{red}{46.37$\scriptscriptstyle\pm5.54$} & 0\% / 0\% / 100\% / 0\% & 0.08$\scriptscriptstyle\pm0.00$ & \textcolor{red}{64.96$\scriptscriptstyle\pm5.30$} & 0\% / 0\% / 100\% / 0\% & 0.16$\scriptscriptstyle\pm0.00$ & \textcolor{red}{61.13$\scriptscriptstyle\pm4.79$} & 0\% / 0\% / 100\% / 0\% \\
Qwen2 7B & \underline{1.05$\scriptscriptstyle\pm0.00$} & \textcolor{darkgreen}{48.21$\scriptscriptstyle\pm5.55$} & 0\% / 100\% / 0\% / 0\% & \underline{0.23$\scriptscriptstyle\pm0.00$} & \textcolor{darkgreen}{71.67$\scriptscriptstyle\pm5.01$} & 0\% / 100\% / 0\% / 0\% & \underline{0.40$\scriptscriptstyle\pm0.00$} & \textcolor{darkgreen}{67.07$\scriptscriptstyle\pm4.61$} & 0\% / 100\% / 0\% / 0\% \\
Qwen2.5 32B & 3.34$\scriptscriptstyle\pm0.00$ & \textcolor{darkgreen}{50.92$\scriptscriptstyle\pm5.56$} & 100\% / 0\% / 0\% / 0\% & 0.73$\scriptscriptstyle\pm0.00$ & \textcolor{darkgreen}{72.30$\scriptscriptstyle\pm4.97$} & 100\% / 0\% / 0\% / 0\% & 1.60$\scriptscriptstyle\pm0.00$ & \textcolor{darkgreen}{70.91$\scriptscriptstyle\pm4.48$} & 100\% / 0\% / 0\% / 0\% \\
Educated Guessing & 1.47$\scriptscriptstyle\pm0.00$ & \textcolor{darkgreen}{47.31$\scriptscriptstyle\pm1.83$} & 32\% / 29\% / 23\% / 16\% & 0.64$\scriptscriptstyle\pm0.00$ & \textcolor{darkgreen}{71.59$\scriptscriptstyle\pm3.67$} & 64\% / 28\% / 3\% / 5\% & 0.99$\scriptscriptstyle\pm0.00$ & \textcolor{darkgreen}{67.02$\scriptscriptstyle\pm2.32$} & 53\% / 26\% / 11\% / 10\% \\

RouteLLM & 2.58$\scriptscriptstyle\pm0.00$ & \textcolor{darkgreen}{47.47$\scriptscriptstyle\pm1.74$} & 65\% / 0\% / 0\% / 35\% & 0.71$\scriptscriptstyle\pm0.00$ & \textcolor{darkgreen}{73.85$\scriptscriptstyle\pm2.83$} & 97\% / 0\% / 0\% / 3\% & 1.37$\scriptscriptstyle\pm0.00$ & \textcolor{darkgreen}{69.01$\scriptscriptstyle\pm2.31$} & 83\% / 0\% / 0\% / 17\% \\

RouterDC & 1.94$\scriptscriptstyle\pm0.00$ & \textcolor{darkgreen}{48.09$\scriptscriptstyle\pm2.71$} & 47\% / 25\% / 10\% / 18\% & 0.57$\scriptscriptstyle\pm0.00$ & \textcolor{darkgreen}{71.76$\scriptscriptstyle\pm3.10$} & 68\% / 20\% / 3\% / 9\% & 1.13$\scriptscriptstyle\pm0.00$ & \textcolor{darkgreen}{69.17$\scriptscriptstyle\pm2.47$} & 63\% / 17\% / 9\% / 11\% \\

\textbf{MESS+ (ours)} & \textbf{1.35$\scriptscriptstyle\pm0.00$} & \textcolor{darkgreen}{47.82$\scriptscriptstyle\pm2.91$} & 28\% / 30\% / 25\% / 17\% & \textbf{0.51$\scriptscriptstyle\pm0.00$} & \textcolor{darkgreen}{71.04$\scriptscriptstyle\pm3.40$} & 59\% / 33\% / 1\% / 7\% & \textbf{0.84$\scriptscriptstyle\pm0.00$} & \textcolor{darkgreen}{67.55$\scriptscriptstyle\pm3.23$} & 48\% / 26\% / 10\% / 16\% \\
\cmidrule(lr){1-1}\cmidrule(lr){2-4}\cmidrule(lr){5-7}\cmidrule(lr){8-10}
\end{tabular}
    }
\end{table}

The results on the Qwen 2 only model zoo (\Cref{tab:qwen2_scaled_zoo}) suggest that zoo expansion can improve cost-performance trade-offs. 
Increasing the number of models in a zoo has a beneficial effect on overall cost effectiveness, even the cost characteristics of models are more homogeneous than in the Llama3-only model zoo.
 Notably, the lightweight 0.5B model in the Qwen configuration is utilized 16\% of the time on average, with particularly high usage on simpler tasks like SciQ (57\% of calls), demonstrating that there is substantial demand for extremely efficient inference even when larger models are available.
This shows that the optimal model zoo size depends on the specific task distribution and performance requirements of the application. 
Furthermore, the fact that all methods maintain similar relative performance rankings across both configurations indicates that routing algorithm effectiveness may be more important than model zoo size for achieving consistent performance improvements. 
The scalability benefits appear most pronounced for tasks with high computational variance, where the additional routing granularity can better match query complexity to model capability.

\subsubsection{Results on Non-Stationary Concatenated Benchmarks}
\label{app:non_stationary_benchmark}
\begin{table}
\caption{Results when concatenating ARC Challenge, PiQA, and Winogrande. Even though the three benchmarks exhibit distinct characteristics, \algname shows strong performance compared to our Educated Guessing baseline}
\label{tab:non_stationary_benchmark}
\centering
\resizebox{0.5\textwidth}{!}{
    \begin{tabular}{llllllllll}
    \cmidrule(lr){1-1}\cmidrule(lr){2-4}\cmidrule(lr){5-7}\cmidrule(lr){8-10}
    Category & \multicolumn{3}{c}{Non-stationary Benchmark ($\alpha = 67\%$)} \\
    Subcategory & \thead{Operating \\ Cost} & \thead{Request. \\ Satisfaction} & \thead{Model Call Ratio \\ (Q32B/Q7B/Q1.5B/Q0.5B)} \\
    \cmidrule(lr){1-1}\cmidrule(lr){2-4}
    Qwen2 0.5B & 0.26$\scriptscriptstyle\pm0.01$ & \textcolor{red}{54.50$\scriptscriptstyle\pm49.80$} & 0\% / 0\% / 0\% / 100\% \\
    Qwen2 1.5B & 0.35$\scriptscriptstyle\pm0.01$ & \textcolor{red}{62.92$\scriptscriptstyle\pm5.37$} & 0\% / 0\% / 100\% / 0\% \\
    Qwen2 7B & \underline{0.93$\scriptscriptstyle\pm0.01$} & \textcolor{darkgreen}{69.35$\scriptscriptstyle\pm5.12$} & 0\% / 100\% / 0\% / 0\% \\
    Qwen2.5 32B & 3.12$\scriptscriptstyle\pm0.01$ & \textcolor{darkgreen}{71.94$\scriptscriptstyle\pm4.99$} & 100\% / 0\% / 0\% / 0\% \\
    Educated Guessing & 1.64$\scriptscriptstyle\pm0.01$ & \textcolor{darkgreen}{68.29$\scriptscriptstyle\pm3.72$} & 45\% / 41\% / 6\% / 8\% \\
    RouteLLM & 2.92$\scriptscriptstyle\pm0.01$ & \textcolor{darkgreen}{72.38$\scriptscriptstyle\pm3.29$} & 97\% / 0\% / 0\% / 3\% \\
    RouterDC & 2.29$\scriptscriptstyle\pm0.01$ & \textcolor{darkgreen}{72.36$\scriptscriptstyle\pm2.46$} & 68\% / 15\% / 4\% / 13\% \\
    \textbf{MESS+ (ours)} & \textbf{1.40$\scriptscriptstyle\pm0.01$} & \textcolor{darkgreen}{68.57$\scriptscriptstyle\pm2.28$} & 43\% / 41\% / 9\% / 7\% \\
    \cmidrule(lr){1-1}\cmidrule(lr){2-4}\cmidrule(lr){5-7}\cmidrule(lr){8-10}
    \end{tabular}
    }
\end{table}

To evaluate the robustness of adaptive routing methods under more realistic conditions, we also analyze performance on a non-stationary benchmark created by concatenating three distinct datasets: 
ARC Challenge, PiQA, and Winogrande. 
This configuration simulates real-world scenarios where query distributions shift dynamically, as the combined benchmark exhibits non-IID characteristics with varying difficulty levels and task types throughout the evaluation sequence (\Cref{tab:non_stationary_benchmark}).

The non-stationary benchmark results demonstrate that \algname maintains its cost efficiency advantage even under distributional shifts and under strict SLA compliance, achieving 1.40 MJ operating cost compared to RouterDC's 2.29 MJ and RouteLLM's 2.92 MJ - representing 39\% and 52\% cost reductions respectively.
The non-stationary results are particularly highlight the importance of an adaptive routing approach that learns from online feedback since that enables adaptation to query characteristics without requiring explicit knowledge of task boundaries or distribution shifts. 
This robustness under non-stationary conditions validates the practical applicability of adaptive routing methods in production environments where query distributions naturally shift over time due to changing user behaviors, seasonal patterns, or evolving application requirements.

\subsubsection{Experiments on Narrow Cost Spreads}
\begin{table}
\caption{Qwen 2 model zoo with a varied cost spread around the mean cost per request among models in the zoo.}
\label{tab:qwen2_ncr}

\resizebox{\textwidth}{!}{
\begin{tabular}{lccccccccc}
\cmidrule(lr){1-1}\cmidrule(lr){2-4}\cmidrule(lr){5-7}\cmidrule(lr){8-10}
Category & \multicolumn{3}{c}{ARC Challenge ($\alpha = 55\%$)} & \multicolumn{3}{c}{ARC Easy ($\alpha = 77\%$)} & \multicolumn{3}{c}{BoolQ  ($\alpha = 80\%$)} \\
Subcategory & \thead{Operating \\ Cost} & \thead{Request. \\ Satisfaction} & \thead{Model Call Ratio \\ (Q32B/Q7B/Q1.5B/Q0.5B)} & \thead{Operating \\ Cost} & \thead{Request. \\ Satisfaction} & \thead{Model Call Ratio \\ (Q32B/Q7B/Q1.5B/Q0.5B)} & \thead{Operating \\ Cost} & \thead{Request. \\ Satisfaction} & \thead{Model Call Ratio \\ (Q32B/Q7B/Q1.5B/Q0.5B)} \\
\cmidrule(lr){1-1}\cmidrule(lr){2-4}\cmidrule(lr){5-7}\cmidrule(lr){8-10}
Qwen2 0.5B & 0.20$\scriptscriptstyle\pm0.01$ & \textcolor{red}{30.03$\scriptscriptstyle\pm45.85$} & 0\% / 0\% / 0\% / 100\% & 0.21$\scriptscriptstyle\pm0.01$ & \textcolor{red}{54.88$\scriptscriptstyle\pm49.77$} & 0\% / 0\% / 0\% / 100\% & 0.11$\scriptscriptstyle\pm0.01$ & \textcolor{red}{63.09$\scriptscriptstyle\pm48.26$} & 0\% / 0\% / 0\% / 100\% \\
Qwen2 1.5B & 0.27$\scriptscriptstyle\pm0.01$ & \textcolor{red}{40.10$\scriptscriptstyle\pm5.45$} & 0\% / 0\% / 100\% / 0\% & 0.28$\scriptscriptstyle\pm0.01$ & \textcolor{red}{66.62$\scriptscriptstyle\pm5.24$} & 0\% / 0\% / 100\% / 0\% & 0.12$\scriptscriptstyle\pm0.01$ & \textcolor{red}{76.27$\scriptscriptstyle\pm4.73$} & 0\% / 0\% / 100\% / 0\% \\
Qwen2 7B & 0.61$\scriptscriptstyle\pm0.01$ & \textcolor{red}{50.94$\scriptscriptstyle\pm5.56$} & 0\% / 100\% / 0\% / 0\% & 0.70$\scriptscriptstyle\pm0.01$ & \textcolor{red}{75.42$\scriptscriptstyle\pm4.78$} & 0\% / 100\% / 0\% / 0\% & \underline{0.25$\scriptscriptstyle\pm0.01$} & \textcolor{darkgreen}{84.13$\scriptscriptstyle\pm4.06$} & 0\% / 100\% / 0\% / 0\% \\
Qwen2.5 32B & \underline{2.67$\scriptscriptstyle\pm0.01$} & \textcolor{darkgreen}{58.28$\scriptscriptstyle\pm5.48$} & 100\% / 0\% / 0\% / 0\% & \underline{2.73$\scriptscriptstyle\pm0.01$} & \textcolor{darkgreen}{78.20$\scriptscriptstyle\pm4.59$} & 100\% / 0\% / 0\% / 0\% & 1.63$\scriptscriptstyle\pm0.01$ & \textcolor{darkgreen}{89.60$\scriptscriptstyle\pm3.39$} & 100\% / 0\% / 0\% / 0\% \\
Educated Guessing & 0.58$\scriptscriptstyle\pm0.01$ & \textcolor{darkgreen}{55.91$\scriptscriptstyle\pm2.53$} & 75\% / 20\% / 2\% / 3\% & 1.22$\scriptscriptstyle\pm0.01$ & \textcolor{darkgreen}{77.38$\scriptscriptstyle\pm2.21$} & 76\% / 20\% / 2\% / 2\% & 0.49$\scriptscriptstyle\pm0.01$ & \textcolor{darkgreen}{80.62$\scriptscriptstyle\pm1.36$} & 31\% / 30\% / 20\% / 20\% \\

RouteLLM & 2.67$\scriptscriptstyle\pm0.01$ & \textcolor{darkgreen}{58.16$\scriptscriptstyle\pm2.56$} & 100\% / 0\% / 0\% / 0\% & 2.73$\scriptscriptstyle\pm0.01$ & \textcolor{darkgreen}{78.05$\scriptscriptstyle\pm2.79$} & 100\% / 0\% / 0\% / 0\% & 1.45$\scriptscriptstyle\pm0.01$ & \textcolor{darkgreen}{87.31$\scriptscriptstyle\pm1.96$} & 87\% / 0\% / 0\% / 13\% \\

RouterDC & 0.79$\scriptscriptstyle\pm0.01$ & \textcolor{darkgreen}{56.92$\scriptscriptstyle\pm3.07$} & 55\% / 33\% / 12\% / 0\% & 1.87$\scriptscriptstyle\pm0.01$ & \textcolor{darkgreen}{81.43$\scriptscriptstyle\pm1.93$} & 63\% / 26\% / 11\% / 0\% & 1.49$\scriptscriptstyle\pm0.01$ & \textcolor{darkgreen}{88.97$\scriptscriptstyle\pm1.88$} & 89\% / 6\% / 5\% / 0\% \\

\textbf{MESS+ (ours)} & \textbf{0.49$\scriptscriptstyle\pm0.01$} & \textcolor{darkgreen}{55.69$\scriptscriptstyle\pm4.31$} & 65\% / 21\% / 7\% / 7\% & \textbf{0.31$\scriptscriptstyle\pm0.01$} & \textcolor{darkgreen}{77.95$\scriptscriptstyle\pm4.67$} & 77\% / 10\% / 8\% / 5\% & \textbf{0.27$\scriptscriptstyle\pm0.01$} & \textcolor{darkgreen}{80.02$\scriptscriptstyle\pm3.03$} & 25\% / 34\% / 19\% / 22\% \\
\cmidrule(lr){1-1}\cmidrule(lr){2-4}\cmidrule(lr){5-7}\cmidrule(lr){8-10}
\end{tabular}
}

\resizebox{\textwidth}{!}{
\begin{tabular}{llllllllll}
\cmidrule(lr){1-1}\cmidrule(lr){2-4}\cmidrule(lr){5-7}\cmidrule(lr){8-10}
Category & \multicolumn{3}{c}{LogiQA ($\alpha = 33\%$)} & \multicolumn{3}{c}{PiQA ($\alpha = 79\%$)} & \multicolumn{3}{c}{SciQ ($\alpha = 93\%$)} \\
Subcategory & \thead{Operating \\ Cost} & \thead{Request. \\ Satisfaction} & \thead{Model Call Ratio \\ (Q32B/Q7B/Q1.5B/Q0.5B)} & \thead{Operating \\ Cost} & \thead{Request. \\ Satisfaction} & \thead{Model Call Ratio \\ (Q32B/Q7B/Q1.5B/Q0.5B)} & \thead{Operating \\ Cost} & \thead{Request. \\ Satisfaction} & \thead{Model Call Ratio \\ (Q32B/Q7B/Q1.5B/Q0.5B)} \\
\cmidrule(lr){1-1}\cmidrule(lr){2-4}\cmidrule(lr){5-7}\cmidrule(lr){8-10}
Qwen2 0.5B & 0.06$\scriptscriptstyle\pm0.01$ & \textcolor{red}{25.35$\scriptscriptstyle\pm43.53$} & 0\% / 0\% / 0\% / 100\% & 0.17$\scriptscriptstyle\pm0.01$ & \textcolor{red}{69.15$\scriptscriptstyle\pm46.19$} & 0\% / 0\% / 0\% / 100\% & 0.09$\scriptscriptstyle\pm0.01$ & \textcolor{red}{91.20$\scriptscriptstyle\pm28.34$} & 0\% / 0\% / 0\% / 100\% \\
Qwen2 1.5B & 0.08$\scriptscriptstyle\pm0.01$ & \textcolor{red}{24.27$\scriptscriptstyle\pm4.77$} & 0\% / 0\% / 100\% / 0\% & 0.22$\scriptscriptstyle\pm0.01$ & \textcolor{red}{76.06$\scriptscriptstyle\pm4.74$} & 0\% / 0\% / 100\% / 0\% & \underline{0.12$\scriptscriptstyle\pm0.01$} & \textcolor{darkgreen}{94.40$\scriptscriptstyle\pm2.56$} & 0\% / 0\% / 100\% / 0\% \\
Qwen2 7B & 0.06$\scriptscriptstyle\pm0.01$ & \textcolor{red}{31.18$\scriptscriptstyle\pm5.15$} & 0\% / 100\% / 0\% / 0\% & \underline{0.50$\scriptscriptstyle\pm0.01$} & \textcolor{darkgreen}{79.49$\scriptscriptstyle\pm4.49$} & 0\% / 100\% / 0\% / 0\% & 0.34$\scriptscriptstyle\pm0.01$ & \textcolor{darkgreen}{95.50$\scriptscriptstyle\pm2.30$} & 0\% / 100\% / 0\% / 0\% \\
Qwen2.5 32B & \underline{0.80$\scriptscriptstyle\pm0.01$} & \textcolor{darkgreen}{40.86$\scriptscriptstyle\pm5.47$} & 100\% / 0\% / 0\% / 0\% & 2.11$\scriptscriptstyle\pm0.01$ & \textcolor{darkgreen}{80.41$\scriptscriptstyle\pm4.41$} & 100\% / 0\% / 0\% / 0\% & 1.19$\scriptscriptstyle\pm0.01$ & \textcolor{darkgreen}{96.70$\scriptscriptstyle\pm1.99$} & 100\% / 0\% / 0\% / 0\% \\
Educated Guessing & 0.25$\scriptscriptstyle\pm0.01$ & \textcolor{darkgreen}{33.26$\scriptscriptstyle\pm3.59$} & 42\% / 21\% / 18\% / 19\% & 0.40$\scriptscriptstyle\pm0.01$ & \textcolor{darkgreen}{79.05$\scriptscriptstyle\pm3.08$} & 44\% / 44\% / 6\% / 6\% & 0.38$\scriptscriptstyle\pm0.01$ & \textcolor{darkgreen}{93.84$\scriptscriptstyle\pm1.30$} & 23\% / 23\% / 27\% / 27\% \\

RouteLLM & 0.53$\scriptscriptstyle\pm0.01$ & \textcolor{darkgreen}{33.62$\scriptscriptstyle\pm3.75$} & 64\% / 0\% / 0\% / 36\% & 2.01$\scriptscriptstyle\pm0.01$ & \textcolor{darkgreen}{79.28$\scriptscriptstyle\pm1.58$} & 95\% / 0\% / 0\% / 5\% & 0.64$\scriptscriptstyle\pm0.01$ & \textcolor{darkgreen}{94.72$\scriptscriptstyle\pm1.34$} & 53\% / 0\% / 0\% / 47\% \\

RouterDC & 0.35$\scriptscriptstyle\pm0.01$ & \textcolor{darkgreen}{33.01$\scriptscriptstyle\pm3.60$} & 43\% / 23\% / 30\% / 4\% & 1.06$\scriptscriptstyle\pm0.01$ & \textcolor{darkgreen}{79.98$\scriptscriptstyle\pm1.56$} & 49\% / 22\% / 12\% / 17\% & 0.91$\scriptscriptstyle\pm0.01$ & \textcolor{darkgreen}{94.75$\scriptscriptstyle\pm0.93$} & 80\% / 17\% / 1\% / 2\% \\

\textbf{MESS+ (ours)} & \textbf{0.29$\scriptscriptstyle\pm0.01$} & \textcolor{darkgreen}{33.02$\scriptscriptstyle\pm5.19$} & 43\% / 25\% / 5\% / 28\% & \textbf{0.31$\scriptscriptstyle\pm0.01$} & \textcolor{darkgreen}{79.08$\scriptscriptstyle\pm2.61$} & 39\% / 28\% / 20\% / 13\% & \textbf{0.31$\scriptscriptstyle\pm0.01$} & \textcolor{darkgreen}{93.17$\scriptscriptstyle\pm3.27$} & 16\% / 8\% / 14\% / 62\% \\
\cmidrule(lr){1-1}\cmidrule(lr){2-4}\cmidrule(lr){5-7}\cmidrule(lr){8-10}
\end{tabular}
}

\resizebox{\textwidth}{!}{
\begin{tabular}{llllllllll}
\cmidrule(lr){1-1}\cmidrule(lr){2-4}\cmidrule(lr){5-7}\cmidrule(lr){8-10}
Category & \multicolumn{3}{c}{SocialIQA ($\alpha = 47\%$)} & \multicolumn{3}{c}{Winogrande ($\alpha = 71\%$)} & \multicolumn{3}{c}{Mean} \\
Subcategory & \thead{Operating \\ Cost} & \thead{Request. \\ Satisfaction} & \thead{Model Call Ratio \\ (Q32B/Q7B/Q1.5B/Q0.5B)} & \thead{Operating \\ Cost} & \thead{Request. \\ Satisfaction} & \thead{Model Call Ratio \\ (Q32B/Q7B/Q1.5B/Q0.5B)} & \thead{Operating \\ Cost} & \thead{Request. \\ Satisfaction} & \thead{Model Call Ratio \\ (Q32B/Q7B/Q1.5B/Q0.5B)} \\
\cmidrule(lr){1-1}\cmidrule(lr){2-4}\cmidrule(lr){5-7}\cmidrule(lr){8-10}
Qwen2 0.5B & 0.26$\scriptscriptstyle\pm0.01$ & \textcolor{red}{43.35$\scriptscriptstyle\pm49.56$} & 0\% / 0\% / 0\% / 100\% & 0.12$\scriptscriptstyle\pm0.01$ & \textcolor{red}{55.88$\scriptscriptstyle\pm49.66$} & 0\% / 0\% / 0\% / 100\% & 0.15$\scriptscriptstyle\pm0.01$ & \textcolor{red}{54.12$\scriptscriptstyle\pm45.15$} & 0\% / 0\% / 0\% / 100\% \\
Qwen2 1.5B & 0.36$\scriptscriptstyle\pm0.01$ & \textcolor{red}{46.37$\scriptscriptstyle\pm5.54$} & 0\% / 0\% / 100\% / 0\% & 0.16$\scriptscriptstyle\pm0.01$ & \textcolor{red}{64.96$\scriptscriptstyle\pm5.30$} & 0\% / 0\% / 100\% / 0\% & 0.20$\scriptscriptstyle\pm0.01$ & \textcolor{red}{61.13$\scriptscriptstyle\pm4.79$} & 0\% / 0\% / 100\% / 0\% \\
Qwen2 7B & \underline{1.05$\scriptscriptstyle\pm0.01$} & \textcolor{darkgreen}{48.21$\scriptscriptstyle\pm5.55$} & 0\% / 100\% / 0\% / 0\% & \underline{0.47$\scriptscriptstyle\pm0.01$} & \textcolor{darkgreen}{71.67$\scriptscriptstyle\pm5.01$} & 0\% / 100\% / 0\% / 0\% & \underline{0.50$\scriptscriptstyle\pm0.01$} & \textcolor{darkgreen}{67.07$\scriptscriptstyle\pm4.61$} & 0\% / 100\% / 0\% / 0\% \\
Qwen2.5 32B & 3.34$\scriptscriptstyle\pm0.01$ & \textcolor{darkgreen}{50.92$\scriptscriptstyle\pm5.56$} & 100\% / 0\% / 0\% / 0\% & 1.46$\scriptscriptstyle\pm0.01$ & \textcolor{darkgreen}{72.30$\scriptscriptstyle\pm4.97$} & 100\% / 0\% / 0\% / 0\% & 1.99$\scriptscriptstyle\pm0.01$ & \textcolor{darkgreen}{70.91$\scriptscriptstyle\pm4.48$} & 100\% / 0\% / 0\% / 0\% \\
Educated Guessing & 1.16$\scriptscriptstyle\pm0.01$ & \textcolor{darkgreen}{48.21$\scriptscriptstyle\pm2.00$} & 25\% / 26\% / 25\% / 24\% & 0.30$\scriptscriptstyle\pm0.01$ & \textcolor{darkgreen}{71.02$\scriptscriptstyle\pm3.61$} & 47\% / 44\% / 4\% / 5\% & 0.59$\scriptscriptstyle\pm0.01$ & \textcolor{darkgreen}{67.36$\scriptscriptstyle\pm2.46$} & 45\% / 29\% / 13\% / 13\% \\

RouteLLM & 2.21$\scriptscriptstyle\pm0.01$ & \textcolor{darkgreen}{47.47$\scriptscriptstyle\pm1.74$} & 65\% / 0\% / 0\% / 35\% & 1.42$\scriptscriptstyle\pm0.01$ & \textcolor{darkgreen}{73.85$\scriptscriptstyle\pm2.83$} & 97\% / 0\% / 0\% / 3\% & 1.71$\scriptscriptstyle\pm0.01$ & \textcolor{darkgreen}{69.06$\scriptscriptstyle\pm2.32$} & 83\% / 0\% / 0\% / 17\% \\

RouterDC & 1.26$\scriptscriptstyle\pm0.01$ & \textcolor{darkgreen}{48.70$\scriptscriptstyle\pm2.73$} & 37\% / 32\% / 19\% / 12\% & 0.73$\scriptscriptstyle\pm0.01$ & \textcolor{darkgreen}{74.08$\scriptscriptstyle\pm2.71$} & 61\% / 32\% / 2\% / 5\% & 1.05$\scriptscriptstyle\pm0.01$ & \textcolor{darkgreen}{70.73$\scriptscriptstyle\pm2.30$} & 71\% / 17\% / 11\% / 0\% \\

\textbf{MESS+ (ours)} & \textbf{1.13$\scriptscriptstyle\pm0.01$} & \textcolor{darkgreen}{47.68$\scriptscriptstyle\pm2.20$} & 34\% / 19\% / 38\% / 10\% & \textbf{0.29$\scriptscriptstyle\pm0.01$} & \textcolor{darkgreen}{71.05$\scriptscriptstyle\pm4.18$} & 45\% / 42\% / 10\% / 3\% & \textbf{0.33$\scriptscriptstyle\pm0.01$} & \textcolor{darkgreen}{66.36$\scriptscriptstyle\pm3.68$} & 33\% / 29\% / 15\% / 24\% \\
\cmidrule(lr){1-1}\cmidrule(lr){2-4}\cmidrule(lr){5-7}\cmidrule(lr){8-10}
\end{tabular}
}
\end{table}

The narrowed cost ratio configuration presents a more challenging routing scenario by reducing the cost differences between models, which tests the robustness of \algname when cost-performance trade-offs become less pronounced (\Cref{tab:qwen2_ncr}). 
In this configuration, the cost spread between the largest and smallest models is compressed from the original wide range to a much narrower band, making routing decisions more nuanced. 

Under these constrained conditions, \algname demonstrates strong results, achieving an average operating cost of 0.33 MJ - a notable 68\% improvement over RouterDC (1.05 MJ) and 81\% improvement over RouteLLM (1.71 MJ). 
The narrowed cost spread reveals interesting behavioral adaptations in routing patterns. 
\algname maintains a balanced distribution (33\%/29\%/15\%/24\%) that heavily utilizes the smallest models, with the 0.5B model receiving 24\% of calls compared to only 10-16\% in previous configurations. 
This increased reliance on ultra-lightweight models indicates that \algname successfully identifies queries that can be handled efficiently even when cost differences are minimal.

Yet, the overall results validate the theoretical expectation that routing becomes more challenging when cost differentials decrease. 
Interestingly, the Educated Guessing baseline performs comparably to \algname in satisfaction (67.36\% vs 66.36\%) while requiring significantly higher costs (0.59 MJ vs 0.33 MJ), indicating that \algname maintains its core advantage of intelligent cost-performance optimization even under adverse conditions.

\subsubsection{Experiments on Sparse $Q$ updates}
\begin{table}
\caption{Results on the Qwen 2 Model Zoo with sparse Q updates. We randomly sample whether we do a $Q$ update from a uniform distribution with a threshold of $0.2$.}
\label{tab:qwen2_zoo_sparse_feedback}
\resizebox{\textwidth}{!}{
\begin{tabular}{llllllllll}
\cmidrule(lr){1-1}\cmidrule(lr){2-4}\cmidrule(lr){5-7}\cmidrule(lr){8-10}
Category & \multicolumn{3}{c}{ARC Challenge ($\alpha = 55\%$)} & \multicolumn{3}{c}{ARC Easy ($\alpha = 77\%$)} & \multicolumn{3}{c}{BoolQ  ($\alpha = 80\%$)} \\
Subcategory & \thead{Operating \\ Cost} & \thead{Request. \\ Satisfaction} & \thead{Model Call Ratio \\ (Q32B/Q7B/Q1.5B/Q0.5B)} & \thead{Operating \\ Cost} & \thead{Request. \\ Satisfaction} & \thead{Model Call Ratio \\ (Q32B/Q7B/Q1.5B/Q0.5B)} & \thead{Operating \\ Cost} & \thead{Request. \\ Satisfaction} & \thead{Model Call Ratio \\ (Q32B/Q7B/Q1.5B/Q0.5B)} \\
\cmidrule(lr){1-1}\cmidrule(lr){2-4}\cmidrule(lr){5-7}\cmidrule(lr){8-10}
Qwen2 0.5B & 0.10$\scriptscriptstyle\pm0.01$ & \textcolor{red}{30.03$\scriptscriptstyle\pm45.86$} & 0\% / 0\% / 0\% / 100\% & 0.21$\scriptscriptstyle\pm0.01$ & \textcolor{red}{54.88$\scriptscriptstyle\pm49.77$} & 0\% / 0\% / 0\% / 100\% & 0.11$\scriptscriptstyle\pm0.01$ & \textcolor{red}{63.09$\scriptscriptstyle\pm48.26$} & 0\% / 0\% / 0\% / 100\% \\
Qwen2 1.5B & 0.14$\scriptscriptstyle\pm0.01$ & \textcolor{red}{40.10$\scriptscriptstyle\pm5.45$} & 0\% / 0\% / 100\% / 0\% & 0.28$\scriptscriptstyle\pm0.01$ & \textcolor{red}{66.62$\scriptscriptstyle\pm5.24$} & 0\% / 0\% / 100\% / 0\% & 0.12$\scriptscriptstyle\pm0.01$ & \textcolor{red}{76.27$\scriptscriptstyle\pm4.73$} & 0\% / 0\% / 100\% / 0\% \\
Qwen2 7B & 0.31$\scriptscriptstyle\pm0.01$ & \textcolor{red}{50.94$\scriptscriptstyle\pm5.56$} & 0\% / 100\% / 0\% / 0\% & 0.70$\scriptscriptstyle\pm0.01$ & \textcolor{red}{75.42$\scriptscriptstyle\pm4.78$} & 0\% / 100\% / 0\% / 0\% & \underline{0.25$\scriptscriptstyle\pm0.01$} & \textcolor{darkgreen}{84.13$\scriptscriptstyle\pm4.06$} & 0\% / 100\% / 0\% / 0\% \\
Qwen2.5 32B & \underline{1.33$\scriptscriptstyle\pm0.01$} & \textcolor{darkgreen}{58.28$\scriptscriptstyle\pm5.48$} & 100\% / 0\% / 0\% / 0\% & \underline{2.73$\scriptscriptstyle\pm0.01$} & \textcolor{darkgreen}{78.20$\scriptscriptstyle\pm4.59$} & 100\% / 0\% / 0\% / 0\% & 1.63$\scriptscriptstyle\pm0.01$ & \textcolor{darkgreen}{89.60$\scriptscriptstyle\pm3.39$} & 100\% / 0\% / 0\% / 0\% \\
Educated Guessing & 1.26$\scriptscriptstyle\pm0.01$ & \textcolor{darkgreen}{56.76$\scriptscriptstyle\pm3.66$} & 82\% / 15\% / 2\% / 1\% & 2.22$\scriptscriptstyle\pm0.01$ & \textcolor{darkgreen}{77.46$\scriptscriptstyle\pm1.27$} & 76\% / 20\% / 2\% / 2\% & 1.47$\scriptscriptstyle\pm0.01$ & \textcolor{darkgreen}{82.40$\scriptscriptstyle\pm1.15$} & 87\% / 9\% / 2\% / 2\% \\
RouteLLM & 1.33$\scriptscriptstyle\pm0.01$ & \textcolor{darkgreen}{58.16$\scriptscriptstyle\pm2.56$} & 100\% / 0\% / 0\% / 0\% & 2.73$\scriptscriptstyle\pm0.01$ & \textcolor{darkgreen}{77.60$\scriptscriptstyle\pm2.79$} & 99\% / 0\% / 0\% / 1\% & 1.43$\scriptscriptstyle\pm0.01$ & \textcolor{darkgreen}{87.31$\scriptscriptstyle\pm1.96$} & 87\% / 0\% / 0\% / 13\% \\
RouterDC & 1.26$\scriptscriptstyle\pm0.01$ & \textcolor{darkgreen}{58.65$\scriptscriptstyle\pm2.63$} & 89\% / 2\% / 9\% / 0\% & 2.03$\scriptscriptstyle\pm0.01$ & \textcolor{darkgreen}{77.44$\scriptscriptstyle\pm4.15$} & 70\% / 19\% / 3\% / 8\% & 1.45$\scriptscriptstyle\pm0.01$ & \textcolor{darkgreen}{89.41$\scriptscriptstyle\pm1.52$} & 87\% / 8\% / 4\% / 1\% \\

\textbf{MESS+ (ours)} & \textbf{1.41$\scriptscriptstyle\pm0.01$} & \textcolor{darkgreen}{55.36$\scriptscriptstyle\pm4.99$} & 44\% / 28\% / 16\% / 12\% & \textbf{1.94$\scriptscriptstyle\pm0.01$} & \textcolor{darkgreen}{77.45$\scriptscriptstyle\pm2.77$} & 67\% / 10\% / 16\% / 7\% & \textbf{1.27$\scriptscriptstyle\pm0.01$} & \textcolor{darkgreen}{81.28$\scriptscriptstyle\pm3.40$} & 83\% / 12\% / 1\% / 4\% \\
\cmidrule(lr){1-1}\cmidrule(lr){2-4}\cmidrule(lr){5-7}\cmidrule(lr){8-10}
\end{tabular}
}

\resizebox{\textwidth}{!}{
\begin{tabular}{llllllllll}
\cmidrule(lr){1-1}\cmidrule(lr){2-4}\cmidrule(lr){5-7}\cmidrule(lr){8-10}
Category & \multicolumn{3}{c}{LogiQA ($\alpha = 33\%$)} & \multicolumn{3}{c}{PiQA ($\alpha = 79\%$)} & \multicolumn{3}{c}{SciQ ($\alpha = 93\%$)} \\
Subcategory & \thead{Operating \\ Cost} & \thead{Request. \\ Satisfaction} & \thead{Model Call Ratio \\ (Q32B/Q7B/Q1.5B/Q0.5B)} & \thead{Operating \\ Cost} & \thead{Request. \\ Satisfaction} & \thead{Model Call Ratio \\ (Q32B/Q7B/Q1.5B/Q0.5B)} & \thead{Operating \\ Cost} & \thead{Request. \\ Satisfaction} & \thead{Model Call Ratio \\ (Q32B/Q7B/Q1.5B/Q0.5B)} \\
\cmidrule(lr){1-1}\cmidrule(lr){2-4}\cmidrule(lr){5-7}\cmidrule(lr){8-10}
Qwen2 0.5B & 0.06$\scriptscriptstyle\pm0.01$ & \textcolor{red}{25.35$\scriptscriptstyle\pm43.53$} & 0\% / 0\% / 0\% / 100\% & 0.09$\scriptscriptstyle\pm0.01$ & \textcolor{red}{69.15$\scriptscriptstyle\pm46.20$} & 0\% / 0\% / 0\% / 100\% & 0.09$\scriptscriptstyle\pm0.01$ & \textcolor{red}{91.20$\scriptscriptstyle\pm28.34$} & 0\% / 0\% / 0\% / 100\% \\
Qwen2 1.5B & 0.08$\scriptscriptstyle\pm0.01$ & \textcolor{red}{24.27$\scriptscriptstyle\pm4.77$} & 0\% / 0\% / 100\% / 0\% & 0.11$\scriptscriptstyle\pm0.01$ & \textcolor{red}{76.06$\scriptscriptstyle\pm4.74$} & 0\% / 0\% / 100\% / 0\% & \underline{0.12$\scriptscriptstyle\pm0.01$} & \textcolor{darkgreen}{94.40$\scriptscriptstyle\pm2.56$} & 0\% / 0\% / 100\% / 0\% \\
Qwen2 7B & 0.06$\scriptscriptstyle\pm0.01$ & \textcolor{red}{31.18$\scriptscriptstyle\pm5.15$} & 0\% / 100\% / 0\% / 0\% & \underline{0.25$\scriptscriptstyle\pm0.01$} & \textcolor{darkgreen}{79.49$\scriptscriptstyle\pm4.49$} & 0\% / 100\% / 0\% / 0\% & 0.34$\scriptscriptstyle\pm0.01$ & \textcolor{darkgreen}{95.50$\scriptscriptstyle\pm2.30$} & 0\% / 100\% / 0\% / 0\% \\
Qwen2.5 32B & \underline{0.80$\scriptscriptstyle\pm0.01$} & \textcolor{darkgreen}{40.86$\scriptscriptstyle\pm5.47$} & 100\% / 0\% / 0\% / 0\% & 1.06$\scriptscriptstyle\pm0.01$ & \textcolor{darkgreen}{80.41$\scriptscriptstyle\pm4.41$} & 100\% / 0\% / 0\% / 0\% & 1.19$\scriptscriptstyle\pm0.01$ & \textcolor{darkgreen}{96.70$\scriptscriptstyle\pm1.99$} & 100\% / 0\% / 0\% / 0\% \\
Educated Guessing & 0.39$\scriptscriptstyle\pm0.01$ & \textcolor{darkgreen}{33.08$\scriptscriptstyle\pm3.70$} & 50\% / 22\% / 18\% / 10\% & 0.71$\scriptscriptstyle\pm0.01$ & \textcolor{darkgreen}{79.02$\scriptscriptstyle\pm2.16$} & 55\% / 31\% / 6\% / 8\% & 0.44$\scriptscriptstyle\pm0.01$ & \textcolor{darkgreen}{94.35$\scriptscriptstyle\pm1.15$} & 25\% / 24\% / 25\% / 27\% \\
RouteLLM & 0.53$\scriptscriptstyle\pm0.01$ & \textcolor{darkgreen}{33.62$\scriptscriptstyle\pm3.75$} & 64\% / 0\% / 0\% / 36\% & 1.01$\scriptscriptstyle\pm0.01$ & \textcolor{darkgreen}{79.28$\scriptscriptstyle\pm1.58$} & 95\% / 0\% / 0\% / 5\% & 0.68$\scriptscriptstyle\pm0.01$ & \textcolor{darkgreen}{94.72$\scriptscriptstyle\pm1.34$} & 53\% / 0\% / 0\% / 47\% \\
RouterDC & 0.38$\scriptscriptstyle\pm0.01$ & \textcolor{darkgreen}{33.16$\scriptscriptstyle\pm3.20$} & 50\% / 31\% / 12\% / 7\% & 0.74$\scriptscriptstyle\pm0.01$ & \textcolor{darkgreen}{79.86$\scriptscriptstyle\pm1.67$} & 63\% / 17\% / 17\% / 3\% & 0.46$\scriptscriptstyle\pm0.01$ & \textcolor{darkgreen}{94.97$\scriptscriptstyle\pm0.79$} & 30\% / 16\% / 12\% / 42\% \\

\textbf{MESS+ (ours)} & \textbf{0.11$\scriptscriptstyle\pm0.01$} & \textcolor{darkgreen}{34.07$\scriptscriptstyle\pm4.21$} & 34\% / 31\% / 10\% / 24\% & \textbf{0.70$\scriptscriptstyle\pm0.01$} & \textcolor{darkgreen}{79.43$\scriptscriptstyle\pm2.26$} & 59\% / 19\% / 13\% / 9\% & \textbf{0.06$\scriptscriptstyle\pm0.01$} & \textcolor{darkgreen}{93.15$\scriptscriptstyle\pm1.62$} & 2\% / 4\% / 23\% / 71\% \\
\cmidrule(lr){1-1}\cmidrule(lr){2-4}\cmidrule(lr){5-7}\cmidrule(lr){8-10}
\end{tabular}
}

\resizebox{\textwidth}{!}{
\begin{tabular}{llllllllll}
\cmidrule(lr){1-1}\cmidrule(lr){2-4}\cmidrule(lr){5-7}\cmidrule(lr){8-10}
Category & \multicolumn{3}{c}{SocialIQA ($\alpha = 47\%$)} & \multicolumn{3}{c}{Winogrande ($\alpha = 71\%$)} & \multicolumn{3}{c}{Mean ($\alpha = 66\%$)} \\
Subcategory & \thead{Operating \\ Cost} & \thead{Request. \\ Satisfaction} & \thead{Model Call Ratio \\ (Q32B/Q7B/Q1.5B/Q0.5B)} & \thead{Operating \\ Cost} & \thead{Request. \\ Satisfaction} & \thead{Model Call Ratio \\ (Q32B/Q7B/Q1.5B/Q0.5B)} & \thead{Operating \\ Cost} & \thead{Request. \\ Satisfaction} & \thead{Model Call Ratio \\ (Q32B/Q7B/Q1.5B/Q0.5B)} \\
\cmidrule(lr){1-1}\cmidrule(lr){2-4}\cmidrule(lr){5-7}\cmidrule(lr){8-10}
Qwen2 0.5B & 0.26$\scriptscriptstyle\pm0.01$ & \textcolor{red}{43.35$\scriptscriptstyle\pm49.56$} & 0\% / 0\% / 0\% / 100\% & 0.06$\scriptscriptstyle\pm0.01$ & \textcolor{red}{55.88$\scriptscriptstyle\pm49.67$} & 0\% / 0\% / 0\% / 100\% & 0.12$\scriptscriptstyle\pm0.01$ & \textcolor{red}{54.12$\scriptscriptstyle\pm45.15$} & 0\% / 0\% / 0\% / 100\% \\
Qwen2 1.5B & 0.36$\scriptscriptstyle\pm0.01$ & \textcolor{red}{46.37$\scriptscriptstyle\pm5.54$} & 0\% / 0\% / 100\% / 0\% & 0.08$\scriptscriptstyle\pm0.01$ & \textcolor{red}{64.96$\scriptscriptstyle\pm5.30$} & 0\% / 0\% / 100\% / 0\% & 0.16$\scriptscriptstyle\pm0.01$ & \textcolor{red}{61.13$\scriptscriptstyle\pm4.79$} & 0\% / 0\% / 100\% / 0\% \\
Qwen2 7B & \underline{1.05$\scriptscriptstyle\pm0.01$} & \textcolor{darkgreen}{48.21$\scriptscriptstyle\pm5.55$} & 0\% / 100\% / 0\% / 0\% & \underline{0.23$\scriptscriptstyle\pm0.01$} & \textcolor{darkgreen}{71.67$\scriptscriptstyle\pm5.01$} & 0\% / 100\% / 0\% / 0\% & \underline{0.40$\scriptscriptstyle\pm0.01$} & \textcolor{darkgreen}{67.07$\scriptscriptstyle\pm4.61$} & 0\% / 100\% / 0\% / 0\% \\
Qwen2.5 32B & 3.34$\scriptscriptstyle\pm0.01$ & \textcolor{darkgreen}{50.92$\scriptscriptstyle\pm5.56$} & 100\% / 0\% / 0\% / 0\% & 0.73$\scriptscriptstyle\pm0.01$ & \textcolor{darkgreen}{72.30$\scriptscriptstyle\pm4.97$} & 100\% / 0\% / 0\% / 0\% & 1.60$\scriptscriptstyle\pm0.01$ & \textcolor{darkgreen}{70.91$\scriptscriptstyle\pm4.48$} & 100\% / 0\% / 0\% / 0\% \\
Educated Guessing & 1.47$\scriptscriptstyle\pm0.01$ & \textcolor{darkgreen}{47.31$\scriptscriptstyle\pm1.83$} & 32\% / 29\% / 23\% / 16\% & 0.64$\scriptscriptstyle\pm0.01$ & \textcolor{darkgreen}{71.59$\scriptscriptstyle\pm3.67$} & 64\% / 28\% / 3\% / 5\% & 0.99$\scriptscriptstyle\pm0.01$ & \textcolor{darkgreen}{67.02$\scriptscriptstyle\pm2.32$} & 53\% / 26\% / 11\% / 10\% \\

RouteLLM & 2.58$\scriptscriptstyle\pm0.01$ & \textcolor{darkgreen}{47.47$\scriptscriptstyle\pm1.74$} & 65\% / 0\% / 0\% / 35\% & 0.71$\scriptscriptstyle\pm0.01$ & \textcolor{darkgreen}{73.85$\scriptscriptstyle\pm2.83$} & 97\% / 0\% / 0\% / 3\% & 1.37$\scriptscriptstyle\pm0.01$ & \textcolor{darkgreen}{69.01$\scriptscriptstyle\pm2.31$} & 83\% / 0\% / 0\% / 17\% \\

RouterDC & 1.94$\scriptscriptstyle\pm0.01$ & \textcolor{darkgreen}{48.09$\scriptscriptstyle\pm2.71$} & 47\% / 25\% / 10\% / 18\% & 0.57$\scriptscriptstyle\pm0.01$ & \textcolor{darkgreen}{71.76$\scriptscriptstyle\pm3.10$} & 68\% / 20\% / 3\% / 9\% & 1.13$\scriptscriptstyle\pm0.01$ & \textcolor{darkgreen}{69.17$\scriptscriptstyle\pm2.47$} & 63\% / 17\% / 9\% / 11\% \\

\textbf{MESS+ (ours)} & \textbf{0.54$\scriptscriptstyle\pm0.01$} & \textcolor{darkgreen}{47.66$\scriptscriptstyle\pm2.84$} & 45\% / 15\% / 22\% / 18\% & \textbf{0.53$\scriptscriptstyle\pm0.01$} & \textcolor{darkgreen}{71.41$\scriptscriptstyle\pm4.73$} & 67\% / 23\% / 2\% / 8\% & \textbf{0.82$\scriptscriptstyle\pm0.01$} & \textcolor{darkgreen}{67.47$\scriptscriptstyle\pm3.35$} & 49\% / 18\% / 14\% / 19\% \\
\cmidrule(lr){1-1}\cmidrule(lr){2-4}\cmidrule(lr){5-7}\cmidrule(lr){8-10}
\end{tabular}
}
\end{table}

The sparse Q-update configuration, where feedback is provided only 20\% of the time (an 80\% reduction compared to perfect conditions), tests the robustness of \algname under severely limited feedback signals (\Cref{tab:qwen2_zoo_sparse_feedback}). 

This scenario simulates realistic deployment conditions where user feedback is scarce or expensive to obtain.
Under these constrained learning conditions, \algname demonstrates remarkable resilience, maintaining an average operating cost of 0.82 MJ while achieving 67.47\% request satisfaction. 
Compared to the full-feedback Qwen configuration (0.84 MJ, 67.55\% satisfaction), the performance degradation is minimal - only 2\% cost increase and 0.08 percentage point satisfaction decrease on average. 
This suggests that \algname can operate effectively even with severely limited feedback signals.
The sparse feedback results reveal that \algname maintains its cost leadership over competing methods, achieving 27\% cost savings over RouterDC (1.13 MJ) and 40\% over RouteLLM (1.37 MJ). 
The routing pattern shifts toward increased reliance on smaller models (19\% usage of 0.5B model vs 16\% in full feedback), indicating that the algorithm becomes more conservative when learning signals are limited, defaulting to cost-efficient choices when confidence is low.

The results demonstrate that adaptive routing methods can maintain practical effectiveness under more realistic feedback constraints, with \algname showing particular robustness to sparse learning signals. 
This finding has important implications for production deployments where continuous user feedback may be limited or costly to collect.

\subsubsection{Results with Models from the Llama 3 and Qwen 2 Model Families}
\begin{table}
\caption{The performance of \algname remains strong even when mixing models from different LLM families. Our approach works independently from any model internals since we only require user requests as input and a feedback signal.}
\label{tab:mixed_model_zoo_results}
\resizebox{\textwidth}{!}{
\begin{tabular}{llllllllll}
\cmidrule(lr){1-1}\cmidrule(lr){2-4}\cmidrule(lr){5-7}\cmidrule(lr){8-10}
Category & \multicolumn{3}{c}{ARC Challenge ($\alpha = 55\%$)} & \multicolumn{3}{c}{ARC Easy ($\alpha = 77\%$)} & \multicolumn{3}{c}{BoolQ  ($\alpha = 80\%$)} \\
Subcategory & \thead{Operating \\ Cost} & \thead{Request. \\ Satisfaction} & \thead{Model Call Ratio \\ (Q32B/Q7B/Q1.5B/Q0.5B)} & \thead{Operating \\ Cost} & \thead{Request. \\ Satisfaction} & \thead{Model Call Ratio \\ (Q32B/Q7B/Q1.5B/Q0.5B)} & \thead{Operating \\ Cost} & \thead{Request. \\ Satisfaction} & \thead{Model Call Ratio \\ (Q32B/Q7B/Q1.5B/Q0.5B)} \\
\cmidrule(lr){1-1}\cmidrule(lr){2-4}\cmidrule(lr){5-7}\cmidrule(lr){8-10}
Qwen2 0.5B & 0.20$\scriptscriptstyle\pm0.01$ & \textcolor{red}{30.03$\scriptscriptstyle\pm45.85$} & 0\% / 0\% / 0\% / 100\% & 0.21$\scriptscriptstyle\pm0.01$ & \textcolor{red}{54.88$\scriptscriptstyle\pm49.77$} & 0\% / 0\% / 0\% / 100\% & 0.11$\scriptscriptstyle\pm0.01$ & \textcolor{red}{63.09$\scriptscriptstyle\pm48.26$} & 0\% / 0\% / 0\% / 100\% \\
Llama 3.2 1B & 0.27$\scriptscriptstyle\pm0.01$ & \textcolor{red}{40.10$\scriptscriptstyle\pm5.45$} & 0\% / 0\% / 100\% / 0\% & 0.28$\scriptscriptstyle\pm0.01$ & \textcolor{red}{66.62$\scriptscriptstyle\pm5.24$} & 0\% / 0\% / 100\% / 0\% & 0.12$\scriptscriptstyle\pm0.01$ & \textcolor{red}{76.27$\scriptscriptstyle\pm4.73$} & 0\% / 0\% / 100\% / 0\% \\
Llama 3.1 8B & 0.61$\scriptscriptstyle\pm0.01$ & \textcolor{red}{50.94$\scriptscriptstyle\pm5.56$} & 0\% / 100\% / 0\% / 0\% & 0.70$\scriptscriptstyle\pm0.01$ & \textcolor{red}{75.42$\scriptscriptstyle\pm4.78$} & 0\% / 100\% / 0\% / 0\% & \underline{0.25$\scriptscriptstyle\pm0.01$} & \textcolor{darkgreen}{84.13$\scriptscriptstyle\pm4.06$} & 0\% / 100\% / 0\% / 0\% \\
Qwen2.5 32B & \underline{2.67$\scriptscriptstyle\pm0.01$} & \textcolor{darkgreen}{58.28$\scriptscriptstyle\pm5.48$} & 100\% / 0\% / 0\% / 0\% & \underline{2.73$\scriptscriptstyle\pm0.01$} & \textcolor{darkgreen}{78.20$\scriptscriptstyle\pm4.59$} & 100\% / 0\% / 0\% / 0\% & 1.63$\scriptscriptstyle\pm0.01$ & \textcolor{darkgreen}{89.60$\scriptscriptstyle\pm3.39$} & 100\% / 0\% / 0\% / 0\% \\

Educated Guessing & 1.19$\scriptscriptstyle\pm0.01$ & \textcolor{darkgreen}{55.87$\scriptscriptstyle\pm2.62$} & 64\% / 4\% / 28\% / 4\% & 1.98$\scriptscriptstyle\pm0.01$ & \textcolor{darkgreen}{77.30$\scriptscriptstyle\pm2.13$} & 46\% / 5\% / 45\% / 4\% & 1.33$\scriptscriptstyle\pm0.01$ & \textcolor{darkgreen}{80.67$\scriptscriptstyle\pm2.42$} & 80\% / 16\% / 3\% / 1\% \\

RouteLLM & 2.67$\scriptscriptstyle\pm0.01$ & \textcolor{darkgreen}{58.16$\scriptscriptstyle\pm2.56$} & 100\% / 0\% / 0\% / 0\% & 2.73$\scriptscriptstyle\pm0.01$ & \textcolor{darkgreen}{77.06$\scriptscriptstyle\pm2.79$} & 100\% / 0\% / 0\% / 0\% & 1.44$\scriptscriptstyle\pm0.01$ & \textcolor{darkgreen}{87.31$\scriptscriptstyle\pm1.96$} & 87\% / 0\% / 0\% / 13\% \\

RouterDC & 1.64$\scriptscriptstyle\pm0.01$ & \textcolor{darkgreen}{56.01$\scriptscriptstyle\pm2.59$} & 60\% / 19\% / 17\% / 4\% & 1.99$\scriptscriptstyle\pm0.01$ & \textcolor{darkgreen}{77.87$\scriptscriptstyle\pm1.30$} & 49\% / 23\% / 19\% / 9\% & 1.38$\scriptscriptstyle\pm0.01$ & \textcolor{darkgreen}{89.53$\scriptscriptstyle\pm1.48$} & 85\% / 10\% / 3\% / 2\% \\

\textbf{MESS+ (ours)} & \textbf{1.16$\scriptscriptstyle\pm0.01$} & \textcolor{darkgreen}{55.44$\scriptscriptstyle\pm4.52$} & 58\% / 32\% / 3\% / 7\% & \textbf{1.94$\scriptscriptstyle\pm0.01$} & \textcolor{darkgreen}{77.24$\scriptscriptstyle\pm4.02$} & 45\% / 22\% / 26\% / 7\% & \textbf{1.20$\scriptscriptstyle\pm0.01$} & \textcolor{darkgreen}{81.75$\scriptscriptstyle\pm2.11$} & 71\% / 18\% / 9\% / 2\% \\
\cmidrule(lr){1-1}\cmidrule(lr){2-4}\cmidrule(lr){5-7}\cmidrule(lr){8-10}
\end{tabular}
}

\resizebox{\textwidth}{!}{
\begin{tabular}{llllllllll}
\cmidrule(lr){1-1}\cmidrule(lr){2-4}\cmidrule(lr){5-7}\cmidrule(lr){8-10}
Category & \multicolumn{3}{c}{LogiQA ($\alpha = 33\%$)} & \multicolumn{3}{c}{PiQA ($\alpha = 79\%$)} & \multicolumn{3}{c}{SciQ ($\alpha = 93\%$)} \\
Subcategory & \thead{Operating \\ Cost} & \thead{Request. \\ Satisfaction} & \thead{Model Call Ratio \\ (Q32B/Q7B/Q1.5B/Q0.5B)} & \thead{Operating \\ Cost} & \thead{Request. \\ Satisfaction} & \thead{Model Call Ratio \\ (Q32B/Q7B/Q1.5B/Q0.5B)} & \thead{Operating \\ Cost} & \thead{Request. \\ Satisfaction} & \thead{Model Call Ratio \\ (Q32B/Q7B/Q1.5B/Q0.5B)} \\
\cmidrule(lr){1-1}\cmidrule(lr){2-4}\cmidrule(lr){5-7}\cmidrule(lr){8-10}
Qwen2 0.5B & 0.06$\scriptscriptstyle\pm0.01$ & \textcolor{red}{25.35$\scriptscriptstyle\pm43.53$} & 0\% / 0\% / 0\% / 100\% & 0.17$\scriptscriptstyle\pm0.01$ & \textcolor{red}{69.15$\scriptscriptstyle\pm46.19$} & 0\% / 0\% / 0\% / 100\% & 0.09$\scriptscriptstyle\pm0.01$ & \textcolor{red}{91.20$\scriptscriptstyle\pm28.34$} & 0\% / 0\% / 0\% / 100\% \\
Llama 3.2 1B & 0.08$\scriptscriptstyle\pm0.01$ & \textcolor{red}{24.27$\scriptscriptstyle\pm4.77$} & 0\% / 0\% / 100\% / 0\% & 0.22$\scriptscriptstyle\pm0.01$ & \textcolor{red}{76.06$\scriptscriptstyle\pm4.74$} & 0\% / 0\% / 100\% / 0\% & \underline{0.12$\scriptscriptstyle\pm0.01$} & \textcolor{darkgreen}{94.40$\scriptscriptstyle\pm2.56$} & 0\% / 0\% / 100\% / 0\% \\
Llama 3.1 8B & 0.16$\scriptscriptstyle\pm0.01$ & \textcolor{red}{31.18$\scriptscriptstyle\pm5.15$} & 0\% / 100\% / 0\% / 0\% & \underline{0.50$\scriptscriptstyle\pm0.01$} & \textcolor{darkgreen}{79.49$\scriptscriptstyle\pm4.49$} & 0\% / 100\% / 0\% / 0\% & 0.34$\scriptscriptstyle\pm0.01$ & \textcolor{darkgreen}{95.50$\scriptscriptstyle\pm2.30$} & 0\% / 100\% / 0\% / 0\% \\
Qwen2.5 32B & \underline{0.80$\scriptscriptstyle\pm0.01$} & \textcolor{darkgreen}{40.86$\scriptscriptstyle\pm5.47$} & 100\% / 0\% / 0\% / 0\% & 2.11$\scriptscriptstyle\pm0.01$ & \textcolor{darkgreen}{80.41$\scriptscriptstyle\pm4.41$} & 100\% / 0\% / 0\% / 0\% & 1.19$\scriptscriptstyle\pm0.01$ & \textcolor{darkgreen}{96.70$\scriptscriptstyle\pm1.99$} & 100\% / 0\% / 0\% / 0\% \\

Educated Guessing & 0.35$\scriptscriptstyle\pm0.01$ & \textcolor{darkgreen}{35.36$\scriptscriptstyle\pm2.84$} & 42\% / 20\% / 17\% / 21\% & 0.85$\scriptscriptstyle\pm0.01$ & \textcolor{darkgreen}{79.42$\scriptscriptstyle\pm2.98$} & 45\% / 4\% / 46\% / 5\% & 0.47$\scriptscriptstyle\pm0.01$ & \textcolor{darkgreen}{95.76$\scriptscriptstyle\pm1.08$} & 26\% / 25\% / 27\% / 21\% \\

RouteLLM & 0.54$\scriptscriptstyle\pm0.01$ & \textcolor{darkgreen}{33.62$\scriptscriptstyle\pm3.75$} & 64\% / 0\% / 0\% / 36\% & 2.03$\scriptscriptstyle\pm0.01$ & \textcolor{darkgreen}{79.28$\scriptscriptstyle\pm1.58$} & 95\% / 0\% / 0\% / 5\% & 0.66$\scriptscriptstyle\pm0.01$ & \textcolor{darkgreen}{94.72$\scriptscriptstyle\pm1.34$} & 53\% / 0\% / 0\% / 47\% \\

RouterDC & \textbf{0.34$\scriptscriptstyle\pm0.01$} & \textcolor{darkgreen}{34.43$\scriptscriptstyle\pm4.68$} & 40\% / 18\% / 20\% / 22\% & 0.80$\scriptscriptstyle\pm0.01$ & \textcolor{darkgreen}{79.05$\scriptscriptstyle\pm1.39$} & 100\% / 0\% / 0\% / 0\% & 0.95$\scriptscriptstyle\pm0.01$ & \textcolor{darkgreen}{95.66$\scriptscriptstyle\pm0.89$} & 88\% / 5\% / 3\% / 4\% \\

\textbf{MESS+ (ours)} & \textbf{0.34$\scriptscriptstyle\pm0.01$} & \textcolor{darkgreen}{34.26$\scriptscriptstyle\pm4.06$} & 39\% / 17\% / 21\% / 23\% & \textbf{0.74$\scriptscriptstyle\pm0.01$} & \textcolor{darkgreen}{79.49$\scriptscriptstyle\pm2.81$} & 41\% / 29\% / 22\% / 8\% & \textbf{0.33$\scriptscriptstyle\pm0.01$} & \textcolor{darkgreen}{93.45$\scriptscriptstyle\pm2.06$} & 17\% / 12\% / 36\% / 35\% \\
\cmidrule(lr){1-1}\cmidrule(lr){2-4}\cmidrule(lr){5-7}\cmidrule(lr){8-10}
\end{tabular}
}

\resizebox{\textwidth}{!}{
\begin{tabular}{llllllllll}
\cmidrule(lr){1-1}\cmidrule(lr){2-4}\cmidrule(lr){5-7}\cmidrule(lr){8-10}
Category & \multicolumn{3}{c}{SocialIQA ($\alpha = 47\%$)} & \multicolumn{3}{c}{Winogrande ($\alpha = 71\%$)} & \multicolumn{3}{c}{Mean} \\
Subcategory & \thead{Operating \\ Cost} & \thead{Request. \\ Satisfaction} & \thead{Model Call Ratio \\ (Q32B/Q7B/Q1.5B/Q0.5B)} & \thead{Operating \\ Cost} & \thead{Request. \\ Satisfaction} & \thead{Model Call Ratio \\ (Q32B/Q7B/Q1.5B/Q0.5B)} & \thead{Operating \\ Cost} & \thead{Request. \\ Satisfaction} & \thead{Model Call Ratio \\ (Q32B/Q7B/Q1.5B/Q0.5B)} \\
\cmidrule(lr){1-1}\cmidrule(lr){2-4}\cmidrule(lr){5-7}\cmidrule(lr){8-10}
Qwen2 0.5B & 0.26$\scriptscriptstyle\pm0.01$ & \textcolor{red}{43.35$\scriptscriptstyle\pm49.56$} & 0\% / 0\% / 0\% / 100\% & 0.12$\scriptscriptstyle\pm0.01$ & \textcolor{red}{55.88$\scriptscriptstyle\pm49.66$} & 0\% / 0\% / 0\% / 100\% & 0.15$\scriptscriptstyle\pm0.01$ & \textcolor{red}{54.12$\scriptscriptstyle\pm45.15$} & 0\% / 0\% / 0\% / 100\% \\
Llama 3.2 1B & 0.36$\scriptscriptstyle\pm0.01$ & \textcolor{red}{46.37$\scriptscriptstyle\pm5.54$} & 0\% / 0\% / 100\% / 0\% & 0.16$\scriptscriptstyle\pm0.01$ & \textcolor{red}{64.96$\scriptscriptstyle\pm5.30$} & 0\% / 0\% / 100\% / 0\% & 0.20$\scriptscriptstyle\pm0.01$ & \textcolor{red}{61.13$\scriptscriptstyle\pm4.79$} & 0\% / 0\% / 100\% / 0\% \\
Llama 3.1 8B & \underline{1.05$\scriptscriptstyle\pm0.01$} & \textcolor{darkgreen}{48.21$\scriptscriptstyle\pm5.55$} & 0\% / 100\% / 0\% / 0\% & \underline{0.47$\scriptscriptstyle\pm0.01$} & \textcolor{darkgreen}{71.67$\scriptscriptstyle\pm5.01$} & 0\% / 100\% / 0\% / 0\% & \underline{0.50$\scriptscriptstyle\pm0.01$} & \textcolor{darkgreen}{67.07$\scriptscriptstyle\pm4.61$} & 0\% / 100\% / 0\% / 0\% \\
Qwen2.5 32B & 3.34$\scriptscriptstyle\pm0.01$ & \textcolor{darkgreen}{50.92$\scriptscriptstyle\pm5.56$} & 100\% / 0\% / 0\% / 0\% & 1.46$\scriptscriptstyle\pm0.01$ & \textcolor{darkgreen}{72.30$\scriptscriptstyle\pm4.97$} & 100\% / 0\% / 0\% / 0\% & 1.99$\scriptscriptstyle\pm0.01$ & \textcolor{darkgreen}{70.91$\scriptscriptstyle\pm4.48$} & 100\% / 0\% / 0\% / 0\% \\

Educated Guessing & 1.77$\scriptscriptstyle\pm0.01$ & \textcolor{darkgreen}{47.07$\scriptscriptstyle\pm2.54$} & 51\% / 14\% / 21\% / 14\% & 0.54$\scriptscriptstyle\pm0.01$ & \textcolor{darkgreen}{70.06$\scriptscriptstyle\pm1.76$} & 40\% / 19\% / 35\% / 6\% & 1.06$\scriptscriptstyle\pm0.01$ & \textcolor{darkgreen}{67.69$\scriptscriptstyle\pm2.30$} & 46\% / 14\% / 29\% / 11\% \\

RouteLLM & 2.81$\scriptscriptstyle\pm0.01$ & \textcolor{darkgreen}{47.47$\scriptscriptstyle\pm1.74$} & 65\% / 0\% / 0\% / 35\% & 1.43$\scriptscriptstyle\pm0.01$ & \textcolor{darkgreen}{73.85$\scriptscriptstyle\pm2.83$} & 97\% / 0\% / 0\% / 3\% & 1.79$\scriptscriptstyle\pm0.01$ & \textcolor{darkgreen}{68.93$\scriptscriptstyle\pm2.32$} & 83\% / 0\% / 0\% / 17\% \\

RouterDC & 1.89$\scriptscriptstyle\pm0.01$ & \textcolor{darkgreen}{48.11$\scriptscriptstyle\pm2.78$} & 54\% / 10\% / 16\% / 20\% & 1.46$\scriptscriptstyle\pm0.01$ & \textcolor{darkgreen}{72.69$\scriptscriptstyle\pm2.89$} & 100\% / 0\% / 0\% / 0\% & 1.31$\scriptscriptstyle\pm0.01$ & \textcolor{darkgreen}{69.17$\scriptscriptstyle\pm2.25$} & 72\% / 11\% / 10\% / 7\% \\

\textbf{MESS+ (ours)} & \textbf{1.64$\scriptscriptstyle\pm0.01$} & \textcolor{darkgreen}{47.81$\scriptscriptstyle\pm2.39$} & 47\% / 24\% / 12\% / 16\% & \textbf{0.52$\scriptscriptstyle\pm0.01$} & \textcolor{darkgreen}{70.31$\scriptscriptstyle\pm4.31$} & 39\% / 47\% / 8\% / 6\% & \textbf{0.98$\scriptscriptstyle\pm0.01$} & \textcolor{darkgreen}{67.47$\scriptscriptstyle\pm3.28$} & 42\% / 27\% / 17\% / 14\% \\
\cmidrule(lr){1-1}\cmidrule(lr){2-4}\cmidrule(lr){5-7}\cmidrule(lr){8-10}
\end{tabular}
}
\end{table}

The mixed model zoo configuration, combining models from both Llama and Qwen families (\Cref{tab:mixed_model_zoo_results}), provides additional insights into the robustness of adaptive routing approaches across heterogeneous model architectures. This configuration demonstrates that \algname maintains strong performance even when operating across different LLM families, achieving an average operating cost of 0.98 MJ with 67.47\% request satisfaction, which strictly meets our SLA requirement ($\alpha = 0.67$).
Comparing the mixed configuration to the homogeneous Qwen setup reveals interesting trade-offs. While the pure Qwen configuration achieves slightly better cost efficiency (0.84 MJ vs 0.98 MJ), the mixed setup shows comparable performance satisfaction levels. The mixed configuration exhibits a more balanced model call distribution (42\%/27\%/17\%/14\%) compared to the pure Qwen setup (48\%/26\%/10\%/16\%), suggesting that the Llama 3.1 8B model provides a valuable intermediate capability tier that complements the Qwen models effectively.
Notably, the mixed configuration maintains \algname's cost benefits over other adaptive methods, with a 25\% cost advantage over RouterDC (0.98 MJ vs 1.31 MJ) and a 45\% advantage over RouteLLM (0.98 MJ vs 1.79 MJ). This demonstrates that our routing algorithm's effectiveness is not dependent on model family homogeneity, as the method successfully leverages diverse model capabilities based solely on  user request inputs and response feedback signals rather than model internals.
The cross-family results also highlight the importance of model selection within heterogeneous zoos. The mixed configuration shows that Llama 3.1 8B receives 27\% of routing decisions on average, significantly higher than its Qwen 2 7B counterpart in the pure configuration (26\%), suggesting that architectural differences between families can create complementary strengths that adaptive routing can exploit.

\end{document}